%% file: arxiv.tex
\pdfoutput=1
\documentclass[journal]{IEEEtran}

\usepackage{times}
\usepackage{cuted}
\usepackage{caption}
\usepackage[numbers]{natbib}
\usepackage{multicol}
\usepackage[bookmarks=true]{hyperref}
\usepackage[dvipsnames]{xcolor}

\newcommand{\todo}[1]{}
\newcommand{\jz}[1]{}
\newcommand{\zac}[1]{}
\newcommand{\simon}[1]{}

\usepackage{graphicx}
\input{preamble}
\arxivtrue
\renewcommand{\revised}[1]{#1}
\hypersetup{pdftitle={Sumo: Dynamic and Generalizable Whole-Body Loco-Manipulation},
            pdfauthor={John Z. Zhang, Maks Sorokin, Jan Bruedigam, Brandon Hung, Stephen Phillips, Dmitry Yershov, Farzad Niroui, Tong Zhao, Leonor Fermoselle, Xinghao Zhu, Chao Cao, Duy Ta, Tao Pang, Jiuguang Wang, Preston Culbertson, Zachary Manchester, Simon Le Cleac'h}}

\begin{document}

\title{Sumo: Dynamic and Generalizable Whole-Body Loco-Manipulation}

\author{\IEEEauthorblockN{%
John~Z.~Zhang$^{1,\,2}$,
Maks~Sorokin$^{2}$*,
Jan~Br\"udigam$^{2}$*,
Brandon~Hung$^{2}$*,
Stephen~Phillips$^{2}$,
Dmitry~Yershov$^{2}$,\\
Farzad~Niroui$^{2}$,
Tong~Zhao$^{2}$,
Leonor~Fermoselle$^{2}$,
Xinghao~Zhu$^{2}$,
Chao~Cao$^{2}$,
Duy~Ta$^{2}$,\\
Tao~Pang$^{2}$,
Jiuguang~Wang$^{2}$,
Preston~Culbertson$^{2,\,3}$,
Zachary~Manchester$^{1}$,
and~Simon~Le~Cl\'eac'h$^{2}$}\\[2pt]
\IEEEauthorblockA{$^{1}$MIT, $^{2}$RAI Institute, $^{3}$Cornell University, *Equal Contribution.\\
This work was done in part during an internship at the RAI Institute.\\
Corresponding Email: jzhang3@mit.edu.}}

\maketitle

\begin{strip}
\vspace*{-4em}
\input{figures/tasks_figure.tex}
\vspace{1.5em}
\end{strip}

\input{sections/abstract}

\IEEEpeerreviewmaketitle

\input{sections/introduction}
\input{sections/related_works}
\input{sections/methods}
\input{sections/experiments}
\input{sections/conclusions}
\input{sections/acknowledgement}

\bibliographystyle{plainnat}
\bibliography{references}
\input{sections/appendix}

\end{document}

%% file: preamble.tex
\usepackage{amsmath}
\usepackage{amsfonts}
\usepackage{amsthm}
\usepackage{amssymb}
\usepackage{mathrsfs}  
\usepackage{bm}
\usepackage[ruled,vlined]{algorithm2e}
\usepackage{dsfont} 
\usepackage{accents} 
\usepackage{tikz-cd} 
\usepackage{adjustbox} 
\usepackage{mathtools} 
\usepackage{bookmark}
\let\labelindent\relax
\usepackage{enumitem}

\usepackage{listings}
\usepackage{xcolor}
\usepackage{caption}
\usepackage{float}

\definecolor{codebg}{rgb}{0.95,0.95,0.95}
\definecolor{keywordcolor}{rgb}{0.0,0.0,0.6}   
\definecolor{typecolor}{rgb}{0.55,0.0,0.55}    
\definecolor{stringcolor}{rgb}{0.6,0.1,0.1}    
\definecolor{commentcolor}{rgb}{0.0,0.5,0.0}   
\definecolor{funcnamecolor}{rgb}{0.0,0.3,0.7}  

\definecolor{sumopurple}{RGB}{174,78,255}
\definecolor{rlnavy}{RGB}{24,20,37}
\definecolor{mpcgreen}{RGB}{71,223,123}

\usepackage{siunitx}

\usepackage{graphicx}
\usepackage{subfig}
\usepackage{pgfplots}
\usepgfplotslibrary{fillbetween}
\pgfplotsset{compat=1.16}

\usepackage{balance}

\newcommand{\comment}[1]{} 

\newcommand{\revised}[1]{{\color{blue}#1}}

\newif\ifarxiv
\arxivfalse

\usepackage{pifont}

\usepackage{lipsum}
\usepackage{booktabs}

%% file: figures/tasks_figure.tex

\centering
\renewcommand{\arraystretch}{0}  

\begin{tabular}{@{}c@{}c@{}c@{}c@{}}
    \includegraphics[width=\dimexpr0.25\textwidth-0.75pt\relax]{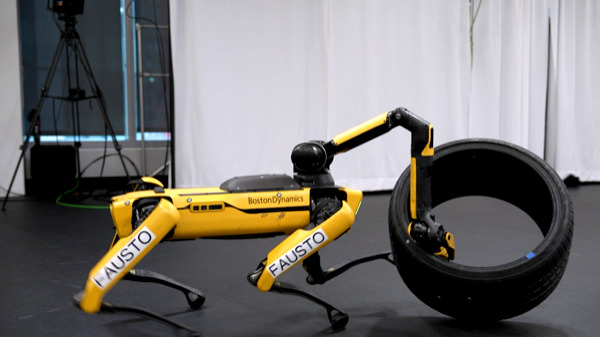}\hspace{1pt} &
    \includegraphics[width=\dimexpr0.25\textwidth-0.75pt\relax]{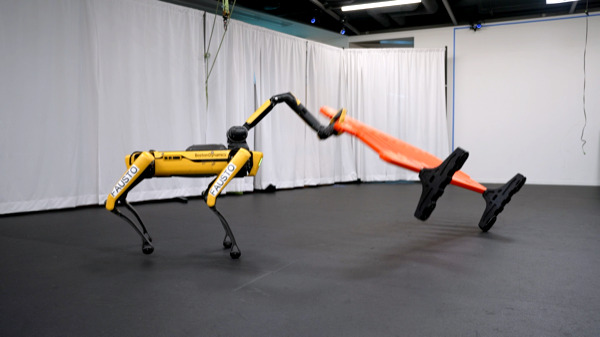}\hspace{1pt} &
    \includegraphics[width=\dimexpr0.25\textwidth-0.75pt\relax]{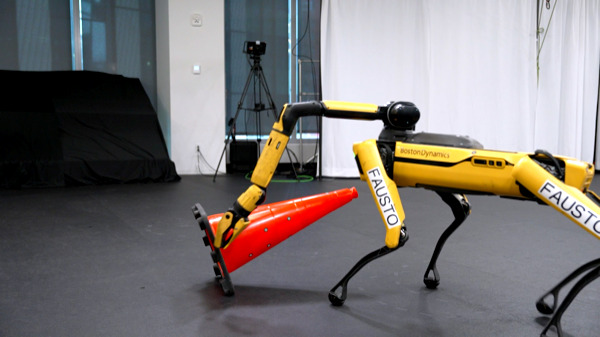}\hspace{1pt} &
    \includegraphics[width=\dimexpr0.25\textwidth-0.75pt\relax]{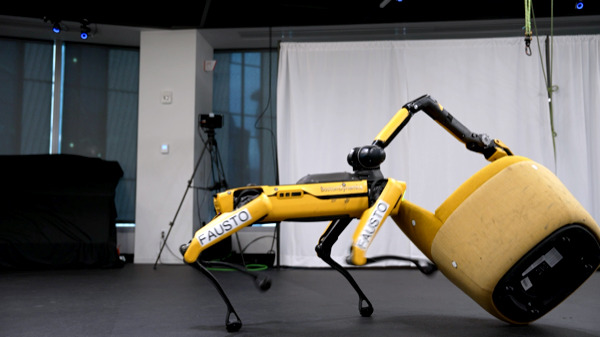} \\
    \noalign{\vskip 1pt}  

    \includegraphics[width=\dimexpr0.25\textwidth-0.75pt\relax]{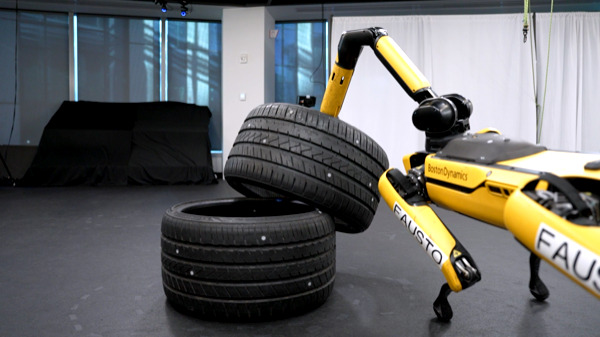}\hspace{1pt} &
    \includegraphics[width=\dimexpr0.25\textwidth-0.75pt\relax]{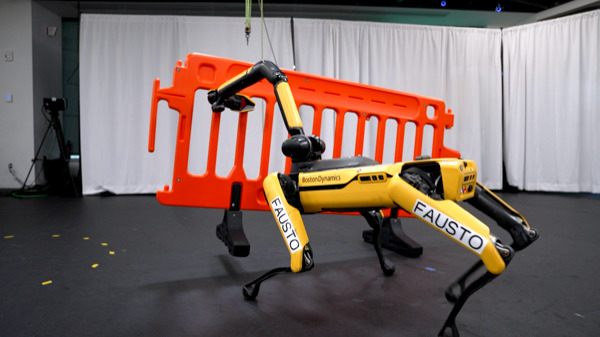}\hspace{1pt} &
    \includegraphics[width=\dimexpr0.25\textwidth-0.75pt\relax]{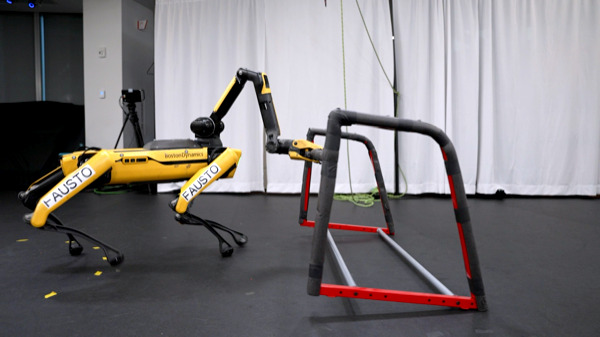}\hspace{1pt} &
    \includegraphics[width=\dimexpr0.25\textwidth-0.75pt\relax]{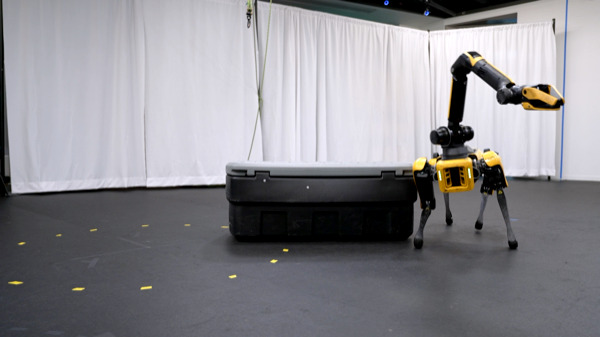} \\
    \noalign{\vskip 1pt}  

    \includegraphics[width=\dimexpr0.25\textwidth-0.75pt\relax]{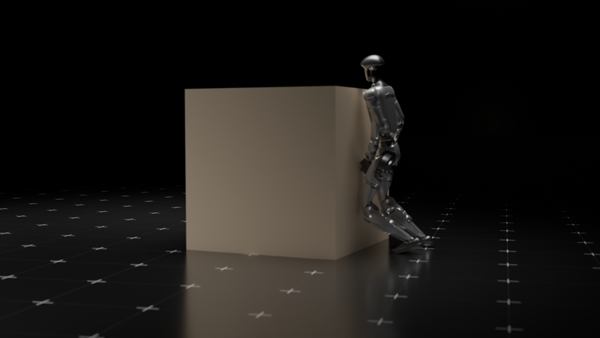}\hspace{1pt} &
    \includegraphics[width=\dimexpr0.25\textwidth-0.75pt\relax]{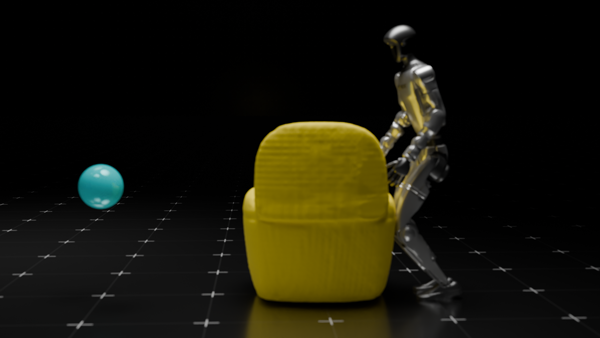}\hspace{1pt} &
    \includegraphics[width=\dimexpr0.25\textwidth-0.75pt\relax]{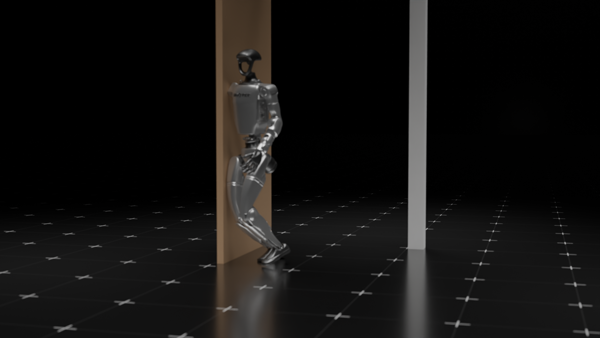}\hspace{1pt} &
    \includegraphics[width=\dimexpr0.25\textwidth-0.75pt\relax]{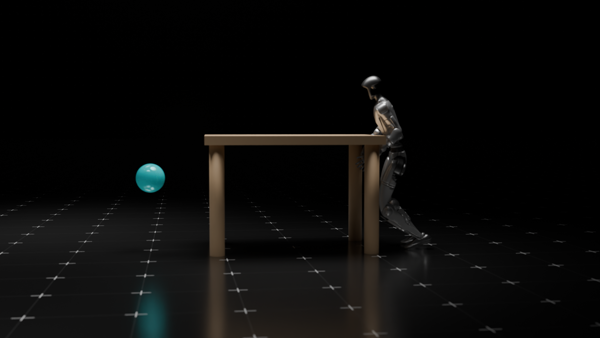} \\
\end{tabular}

\captionof{figure}{First row from left to right: Spot robot uprights a tire, crowd-control barrier, traffic cone, and chair. Second row from left to right: Spot robot stacks a tire on top of another tire, drags a crowd-control barrier, a tire rack, and large box to a target position. Third row from left to right: simulated G1 humanoid robot pushes a box, chair, door, and a dining table.}
\vspace{-10pt}
\label{fig:teaser_robot_tasks}

%% file: sections/abstract.tex
\begin{abstract}
This paper presents a sim-to-real approach that enables legged robots to dynamically manipulate large and heavy objects with whole-body dexterity. Our key insight is that by performing test-time steering of a pre-trained whole-body control policy with a sample-based planner, we can enable these robots to solve a variety of dynamic loco-manipulation tasks. Interestingly, we find our method generalizes to a diverse set of objects and tasks with no additional tuning or training, and can be further enhanced by flexibly adjusting the cost function at test time. We demonstrate the capabilities of our approach through a variety of challenging loco-manipulation tasks on a Spot quadruped robot in the real world, including uprighting a tire heavier than the robot's nominal lifting capacity and dragging a crowd-control barrier larger and taller than the robot itself. Additionally, we show that the same approach can be generalized to humanoid loco-manipulation tasks, such as opening a door and pushing a table, in simulation.
\ifarxiv Project code and videos are available at \href{https://sumo.rai-inst.com/}{https://sumo.rai-inst.com/}.\fi
\end{abstract}

%% file: sections/introduction.tex
\section{Introduction}\label{sec:intro}
Humans and animals physically interact with the world in creative ways: Their ability to gracefully make and break contact with their environment enables them to traverse treacherous terrain and move objects several times larger than themselves. Achieving similar levels of athletic intelligence for robotic systems has been a long-standing challenge. Over the past decade, developments in numerical optimization~\cite{cleach2024, grandia2023perceptive, bledt2018cheetah} and machine learning~\cite{hwangbo2019learning, cheng2024parkour} have made significant progress towards building robots with agility and dexterity~\cite{suh2025dexterous, qi2025simple}. The next research frontier lies in enabling robots to manipulate objects during locomotion, or so-called loco-manipulation. So far, research on loco-manipulation has largely focused on learning from human demonstrations through teleoperation or video imitation, which are usually limited to quasi static table-top settings and fails to leverage the passive dynamics of the object or the robot. Algorithms and systems that enable a robot to autonomously coordinate its entire body to dynamically move large and heavy objects during loco-manipulation have, so far, been elusive.

This paper studies loco-manipulation problems in which the manipulated object is large, heavy, or geometrically complex enough that success requires dynamic coordination of the robot's whole body. Recent works have shown impressive in-the-wild locomotion agility enabled by reinforcement learning (RL). However, adapting RL to each new manipulation task often requires substantial reward engineering, retraining, and compute, and can be difficult to generalize beyond the training distribution. On the other hand, sample-based model-predictive control (MPC) has proven to be a simple and training-free method for contact-rich manipulation~\cite{howell2022mjpc, alvarez2025real, li2025_judo, li2024drop}. Yet sample-based MPC struggles to find good solutions for high-degree-of-freedom robots or dynamically unstable tasks, which frequently occur in quadruped and humanoid loco-manipulation.

Our approach combines the strengths of both. We start from a pre-trained generalist whole-body control policy learned with RL and use real-time sample-based MPC to steer it for contact-rich manipulation of previously unseen objects. This hierarchy addresses key shortcomings of both end-to-end RL and sample-based MPC while preserving their main advantages. RL provides a robust whole-body controller when the training distribution can be modeled with sufficient randomization~\cite{2025relic}. In turn, the low-level policy reduces the effective action space for online planning and stabilizes the dynamics, mitigating the divergence issues associated with single-shooting rollouts of unstable systems.

Our framework, which we call Sumo, enables the high-level sample-based controller to steer the policy effectively for real-world manipulation. Our experiments reveal two interesting properties of this approach. First, hierarchical structure simplifies loco-manipulation. Compared to end-to-end MPC, planning in the command space of a pre-trained whole-body policy reduces the effective search space and stabilizes the robot's contact-rich dynamics. Compared to end-to-end RL, the same hierarchy achieves similar or better success with much simpler task objectives and without task-specific retraining. Second, planning enables generalization. Here, we define generalization as reusing the same controller on new objects and new task objectives by changing only the planner's object model or cost function at test time, without retraining. Keeping high-level decision making online allows the same framework to make these adaptations directly at deployment. \revised{Finally, our approach is also ammenable to desiging complex behaviors via reward shaping more quickly by avoiding offline policy training.} The result is a robust yet flexible framework that enables quadruped and humanoid robots to use their entire body to manipulate objects with complex geometries and with sizes and weights comparable to the robots themselves, Fig. \ref{fig:teaser_robot_tasks}.

Our contributions include:
\begin{itemize}
    \item A hierarchical framework that combines a pre-trained generalist whole-body control policy and test-time planning for dynamic whole-body loco-manipulation.
    \item A set of experimental comparisons showing that hierarchical structure simplifies loco-manipulation by reducing search difficulty and task-specification complexity relative to end-to-end MPC and end-to-end RL baselines.
    \item A set of experimental comparisons showing that test-time planning enables adaptation to new objects and new task objectives without retraining by changing the object model or cost function at deployment.
    \item A set of eight real-world demonstrations on the Spot quadruped robot and four demonstrations on the G1 humanoid robot in simulation covering diverse and challenging loco-manipulation scenarios.
    \item An open-source benchmark and dataset focusing on loco-manipulation tasks with objects that require whole-body coordination due to their size and weight relative to the robots' physical limits.
\end{itemize}

The remainder of this paper reviews related work (Sec. \ref{sec: related works}), details the Sumo framework (Sec. \ref{sec: methods}), analyzes it against baselines in controlled simulation (Sec. \ref{sec: analysis}), demonstrates it on a real-world Spot quadruped and a simulated G1 humanoid (Sec. \ref{sec: experiments}), and concludes with limitations and future directions (Sec. \ref{sec: conclusions}).

\begin{figure}
    \centering
    \includegraphics[width=\linewidth]{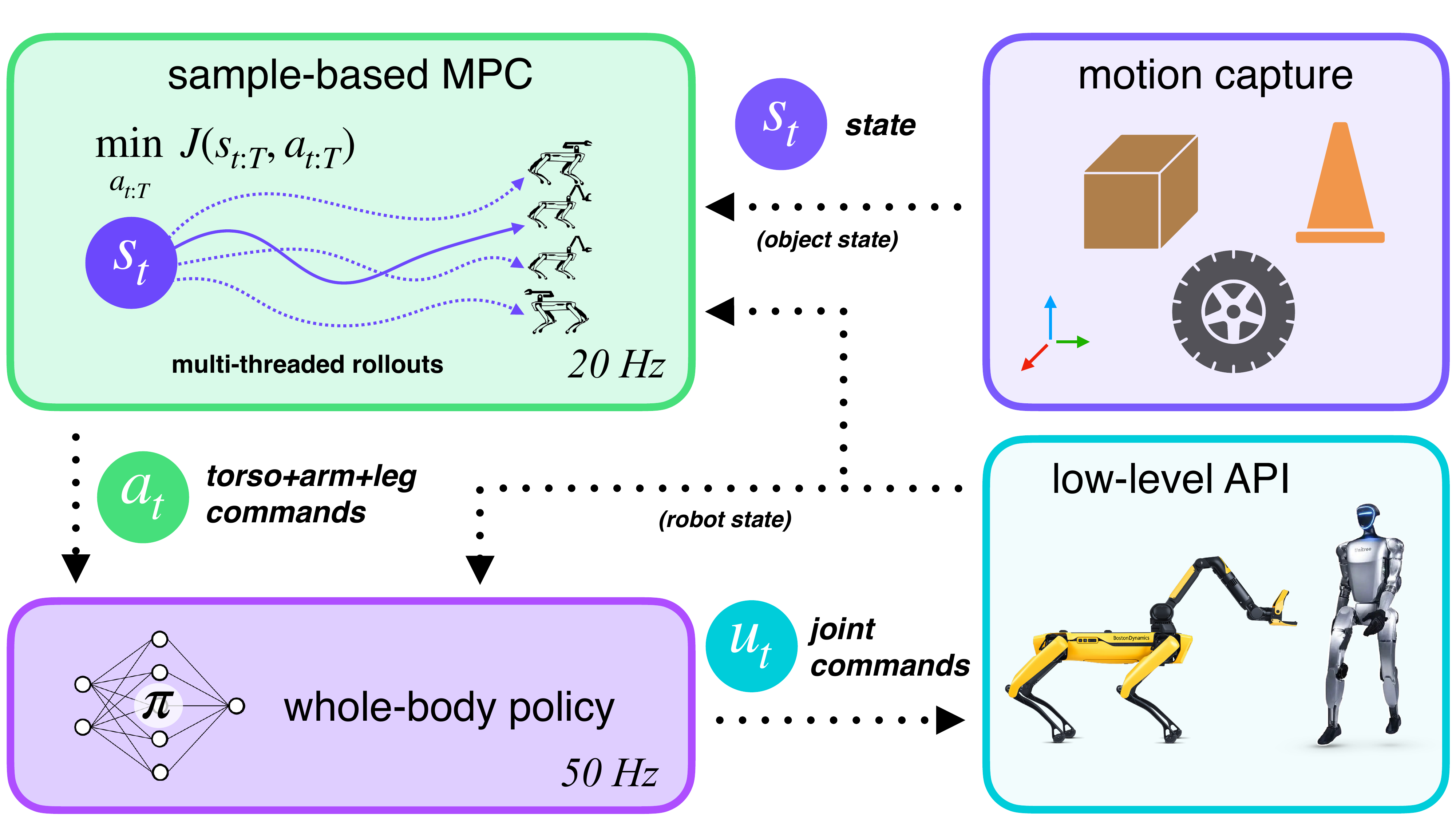}
    \caption{System overview: Our method takes a hierarchical approach that combines a pre-trained whole-body control (WBC) policy (purple) with high-level sample-based MPC (green). The low-level whole-body control policy takes in the current state and desired torso, arm, and leg commands and outputs the joint-level commands for the quadruped or humanoid robot at $50$Hz. The high-level sample-based MPC aims to minimize a task-specific cost function by taking in the current state estimate and solving for the desired torso, arm, and leg commands for the low-level policy at $20$Hz.
    }
    \vspace{-15pt}
    \label{fig:system_overview}
\end{figure}

%% file: sections/related_works.tex

\section{Background and Related Works}\label{sec: related works}
This section reviews relevant literature in RL, sample-based MPC, and their recent applications to loco-manipulation.

\subsection{Reinforcement Learning for Whole-Body Control}
Deep reinforcement learning aims to learn neural-network policies via self-play trial and error. In robotics, the advancements in large-scale parallel simulation platforms~\cite{mittal2023orbit, mujoco_warp2025}  on accelerated hardware (such as GPUs) have significantly reduced the cost of collecting data in simulation. As a result, on-policy RL algorithms, such as PPO~\cite{schulman2017ppo}, have become a staple in the sim-to-real policy optimization paradigm. Additionally, RL naturally incorporates robust policy optimization via domain randomization techniques~\cite{tobin2017domain}, reducing the so-called sim-to-real gap.

Despite its promise, this RL paradigm faces several significant drawbacks in practice: First, RL typically requires dense reward design to find efficient policies with finite compute~\cite{ng1999policy, andrychowicz2017hindsight}. Second, the final policy is often sensitive to reward design, and finding a suitable reward term itself is often a non-trivial and manual search problem. Because of these practical shortcomings, researchers typically have to go through a time-consuming loop of reward engineering, policy training, and hardware testing, before going back to reward engineering for each task. While the recent emergence of LLMs~\cite{yu2023language, ma2023eureka} can automate this process to some degree, reliably finding suitable rewards that lead to successful sim-to-real transfer still requires (expert) human-in-the-loop intervention, especially when policy optimization is time-consuming.

The sim-to-real RL paradigm has been particularly successful for legged robot locomotion~\cite{hwangbo2019learning, cheng2024parkour}, where rewards have become relatively standardized and one can carefully design curricula for difficult terrain variations. However, in open-world manipulation, robots are faced with reasoning over a diverse set of scenarios including objects that vary in size, weight, geometry, friction, etc. This test-time diversity makes randomizing all possible distributions challenging at training time for sim-to-real RL. As a result, successful real-world manipulation, so far, has come from learning from demonstrations~\cite{chi2024diffusionpolicy}.

In this work, we take advantage of RL's strength---learning robust locomotion policies over complex terrains offline---and leave the complexity of generalizable contact-rich decision making to online search via sample-based MPC.

\subsection{Sample-Based MPC}
Sample-based MPC is a family of zeroth-order trajectory-optimization methods that aims to solve for an optimal control sequence over a (typically short) prediction horizon.
Sample-based MPC has been derived from many perspectives, including path integrals~\cite{williams2016}, information theory~\cite{williams2018}, importance sampling~\cite{zhang2014crossentropy}, and diffusion~\cite{pan2024modelbaseddiffusiontrajectoryoptimization}. Additionally, sample-based MPC methods are closely related to other derivative-free optimization algorithms such as CMA-ES~\cite{hansen2016cma}.
Notably, similar to zeroth-order RL, which typically requires lots of compute \emph{offline}, sample-based MPC can be viewed as an \emph{online} policy-gradient method~\cite{qiu2025zerothorder} that is entirely training-free, making it extremely flexible at test time.

Because of its parallel nature and ease of implementation, this family of trajectory optimization algorithms has grown in popularity in recent years, along with the rise of modern parallel computing hardware. The derivative-free nature of sample-based MPC is amenable to contact-rich dynamics and learned black-box deep neural network dynamics models, where computing derivatives is expensive or unreliable~\cite{tracy2025trajectory, zhang2025wholebodymodelpredictivecontrollegged}, which often occurs in contact-rich control. In contrast, classical gradient-based methods such as the iterative linear-quadratic regulator (iLQR)~\cite{jacobsonMayne1970DDP, li2004iterative} and direct-collocation~\cite{kelly2017introduction} typically struggle in these scenarios without additional care to modify the derivatives~\cite{cleach2024}.

Online sample-based MPC has two main pitfalls: the curse of dimensionality and unstable dynamical systems. As an online search algorithm, sample-based MPC suffers from the curse of dimensionality that stems from high-dimensional input spaces associated with highly articulated robots and longer prediction horizons. Recent works have proposed to reduce the temporal dimensionality by sampling over low-order spline control points~\cite{howell2022mjpc, li2025_judo, alvarez2025real}, making sample-based MPC tractable in real-time for robots like dexterous hands, quadrupeds, and humanoids. Furthermore, sample-based MPC generally relies on single-shooting dynamics rollouts, leading to divergence on open-loop unstable systems, which commonly occur in legged robotics.

In our framework, by leveraging a low-level WBC policy, the sample-based MPC enjoys the benefit of sampling in a reduced action space on a stabilized dynamical system, drastically improving sample efficiency during online search.
\input{figures/dynamics_comparison.tex}

\subsection{Loco-Manipulation}
Building robots that manipulate objects during locomotion, or so-called ``loco-manipulation", has become an increasingly important yet challenging research direction as it inherits the fundamental challenges of both manipulation and locomotion for legged robots. Prior works have deployed both model-based methods~\cite{alvarez2025real} and RL methods~\cite{sferrazza2024humanoidbench} to achieve joint locomotion and manipulation of objects. Notably, several works have investigated using legs as manipulators~\cite{2025relic, cheng2023legmanip, arm2024pedipulate, sriganesh2025action, cheng2025rambo}.

For humanoid loco-manipulation, retargeting human motions has become a popular method to bootstrap downstream learning and optimization~\cite{yang2025omniretarget, zhang_slomo_2023}. However, these teleoperation and retargeting approaches are much less applicable to non-anthropomorphic robots, like Spot. \cite{sleiman2023versatile} explores building a planning and control framework for legged robot loco-manipulation, but the control module requires fixed contact modes during planning and is limited to quasi-static behaviors.

Different from existing literature, we focus on manipulating objects that are larger than the robot itself or heavier than the maximum lifting capacity of the robot, thus requiring the robot to efficiently coordinate its entire body to solve these tasks. Our method naturally incorporates RL and online MPC while retaining detailed robot dynamics and collision geometry, and without requiring fixed contact modes during planning~\cite{cheng2025rambo,sleiman2023versatile}.

\subsection{\revised{Hierarchical Control Methods}}
Many recent works aim to leverage the robustness of RL policies together with the structure of model-based control~\cite{dtc, cheng2025rambo, cajun2023, jeon2025residualmpc}. This broad family includes RL-augmented whole-body controllers, optimization-guided tracking policies, and residual policies around MPC controllers. These methods have demonstrated strong tracking, locomotion, and force-control capabilities by using model-based components to provide structure, reference motions, or low-level constraints while learned policies improve robustness. Sumo is closely related to this line of work, but studies a complementary setting: autonomous, goal-conditioned loco-manipulation in which no teleoperator, expert motion library, or task-specific model-based planner provides a reference motion. In this setting, the online optimizer must reason over object interactions and manipulation objectives directly.

More broadly, a common design pattern in hybrid RL-MPC for contact-rich robotics is a high-level RL, low-level MPC hierarchy~\cite{dtc, cheng2025rambo, cajun2023, jeon2025residualmpc}. Sumo instead explores the alternative high-level MPC, low-level RL paradigm, motivated by the recent emergence of powerful generalist tracking policies~\cite{2025relic, li2026omnitrackgeneralmotiontracking, luo2025sonicsupersizingmotiontracking} that typically interface with task-space teleoperation or follow motion references. By using MPC to steer such a policy online, we can synthesize dynamic behaviors from task objectives at deployment rather than depending on teleoperated or planner-generated references. \revised{Both the planner and the tracking policy are otherwise standard; beyond demonstrating this pairing, we analyze why the decomposition is effective, showing that the low-level policy shrinks the search space and stabilizes rollouts while keeping high-level optimization online retains a test-time flexibility that offline-trained high-level policies sacrifice.} \revised{Closest to our setting, Kim et al.~\cite{kim2025highspeed} combine a sampling-based planner with a low-level RL policy for high-speed quadruped locomotion and navigation on complex, discrete terrain. Sumo shares this structure but studies dynamic, contact-rich \emph{loco-manipulation}, where the planner must reason through the pretrained policy about object dynamics and manipulation objectives rather than terrain traversal.} Planning over pretrained robot policies has also been explored in tabletop manipulation~\cite{vgps, jain2025smoothseanevermade} where the base policy is trained via imitation learning.

\revised{Finally, an alternative high-level policy is a \emph{learned} decision maker such as a vision-language-action (VLA) model~\cite{kim2024openvla, black2024pi0}. VLAs offer semantic, open-world generalization from large-scale demonstrations but currently operate at low rates on quasi-static, gripper-centric tasks, and new embodiments or dynamic skills require substantial data collection; Sumo instead requires an object model and a cost function, and in return provides high-rate replanning for dynamic whole-body contact. The two are complementary: a VLA could supply goals, object models, or cost functions that Sumo executes through dynamic replanning (Sec.~\ref{sec: conclusions}).}

%% file: figures/dynamics_comparison.tex
\begin{figure}
\centering
\begin{tikzpicture}
    \node[anchor=south west,inner sep=0] (img) at (0, 0)
        {\includegraphics[width=0.99\linewidth]{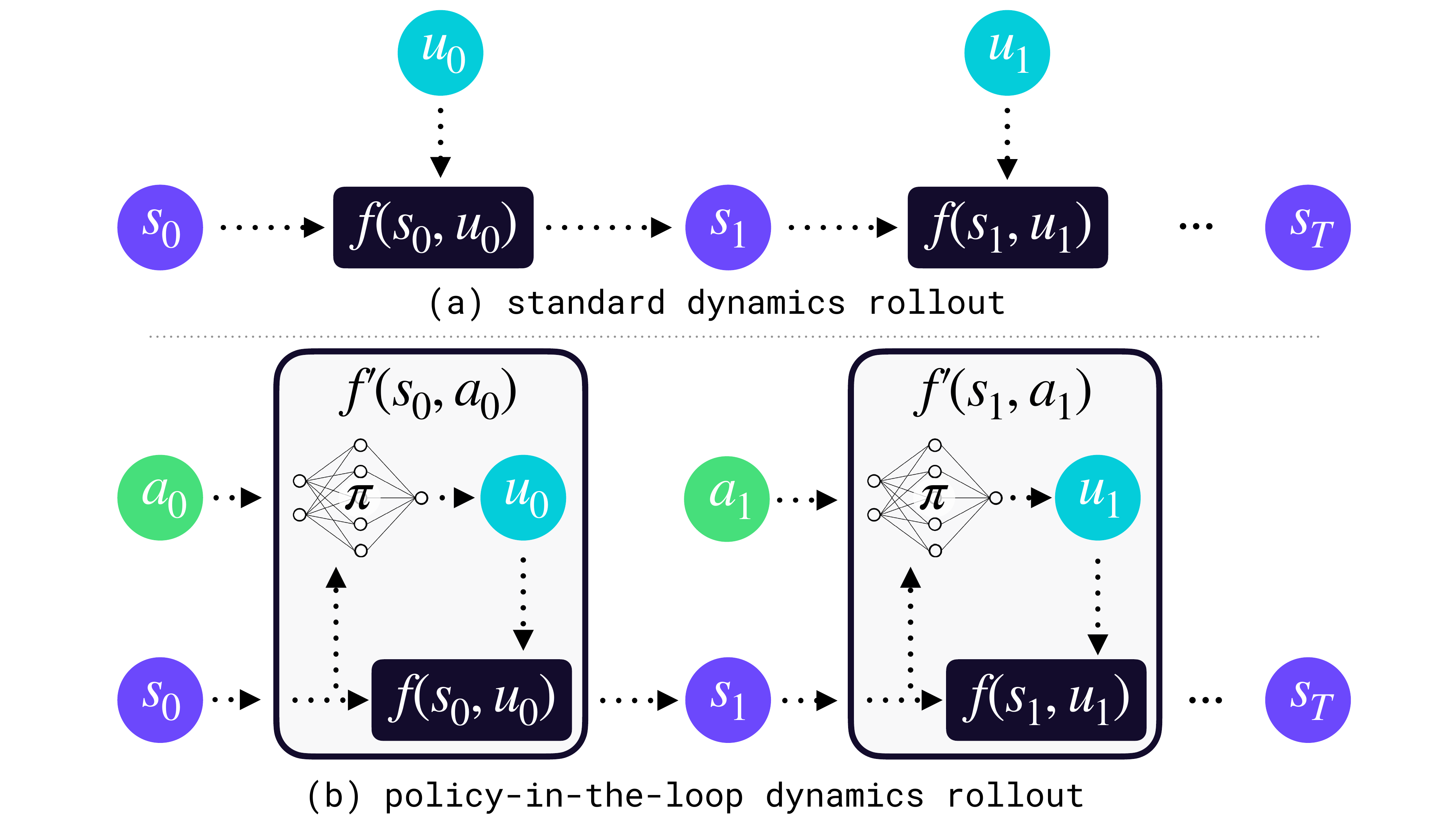}};

\end{tikzpicture}
\caption{Illustrations comparing (a) standard dynamics rollouts $f$ where the actions $u$ are the joint-level controls for the multi-body dynamics model and (b) our \revised{policy-in-the-loop} dynamics rollouts \revised{$f'$}, where the actions $a$ are inputs to the low-level locomotion policy \revised{$\pi$}.
}
\vspace{-20pt}
\label{fig:dynamics_comparison}
\end{figure}

%% file: sections/methods.tex

\section{Sumo} \label{sec: methods}
In this section, we provide a detailed description of our hierarchical control framework, which we call Sumo, that uses a high-level sample-based controller to steer a pre-trained whole-body control policy to perform whole-body loco-manipulation with quadruped and humanoid robots.
\begin{table}[t]
\centering
\caption{Timing comparison of MuJoCo simulation of $32$ parallel rollouts over $1.5$ seconds with and without the low-level policy on an Intel Core i7-12700K CPU.}
\label{tab:timing_comparison}
\begin{tabular}{lcc}
\toprule
Method & Rollout Time (ms) \\
\midrule
MuJoCo with low-level policy & $43.45 \pm 1.88$ ms  \\
MuJoCo only & $21.73 \pm 1.86$ ms \\
\bottomrule
\end{tabular}
\vspace{-20pt}
\end{table}
\subsection{Overview}
Our system, detailed in Figure \ref{fig:system_overview}, uses a pre-trained steerable whole-body control policy that takes in the current state and desired torso, arm, and optionally leg commands of the robot and outputs the low-level joint commands for the quadruped or humanoid robot at $50$Hz. At the high-level, we use a sample-based MPC policy (e.g., MPPI, CEM, etc.) that takes in the current state estimate and updates the desired torso and arm commands for the low-level locomotion policy at $20$Hz. Together, our framework enables complex loco-manipulation skills that can only be achieved through whole-body contact-rich coordination, such as uprighting and stacking a fallen tire heavier than the robot's lifting capacity (Fig. \ref{fig:spot_tasks_freeze_frame} (a) (e)), and uprighting and dragging a crowd control barrier larger than the robot itself (Fig. \ref{fig:spot_tasks_freeze_frame} (b) (f)).

\subsection{Pre-Trained Whole-Body Control Policy}
For the low-level policy, we use an existing whole-body control policy trained via RL that takes high-level commands and outputs the joint-level commands for the entire robot. Methods for training humanoid and quadruped policies have been studied extensively in the literature and are widely available~\cite{hwangbo2019learning, cheng2024parkour}. We do not aim to improve whole-body control capabilities in this paper, and instead focus on their downstream applications for loco-manipulation. Specifically, we use the Relic policy~\cite{2025relic} which is designed for multi-limb loco-manipulation on the Spot quadruped robot. In particular, the Relic policy enables stable gaits with only three legs and allows the robot to use the fourth leg as a manipulator in addition to its arm and torso. This setup allows the high-level sampling-based controller to automatically reason about using multiple limbs or a combination of limbs and torso to manipulate an otherwise challenging object.

For the G1 humanoid robot, we use the standard velocity-tracking policy from MJLab~\cite{mjlab2025} that takes in the desired torso velocity commands. While this policy is not explicitly trained for loco-manipulation and does not take in the desired arm commands, we find that simply overriding the arm commands with target commands from the high-level sampling-based MPC is also effective.

While we believe these are natural choices for the Spot and G1 robots, one can similarly choose other policies over arbitrary input spaces from which the sample-based controller can sample; we return to interface design considerations below.
\begin{algorithm}[t]
\color{blue}
\caption{Sumo Planning Iteration}
\label{alg:sumo_cem}
\KwIn{state estimate $\mathbf{s}_0$; previous plan spline $\mathcal{S}$; per-node noise scales $\boldsymbol{\sigma}$; task cost $J$}
$\bar{\mathbf{A}} \leftarrow \mathcal{S}(t)$ \tcp*[l]{warm start: time-shift previous plan}
\For(\tcp*[h]{in parallel}){each rollout $i$}{
    $\mathbf{A}^i \leftarrow \operatorname{clip}\!\left(\bar{\mathbf{A}} + \boldsymbol{\sigma} \odot \boldsymbol{\epsilon}^i\right)$, \quad $\boldsymbol{\epsilon}^i \sim \mathcal{N}(\mathbf{0}, \mathbf{I})$\;
    $\mathbf{a}^i(t) \leftarrow$ interpolate control points $\mathbf{A}^i$\;
    $\mathbf{s}^i_{k+1} = f^\prime\!\left(\mathbf{s}^i_k, \mathbf{a}^i(t_k)\right),\ \mathbf{s}^i_0 = \mathbf{s}_0$ \tcp*[l]{policy-in-the-loop rollout}
    $J^i \leftarrow \sum_k J\!\left(\mathbf{s}^i_k, \mathbf{a}^i(t_k)\right)$\;
}
retain the unperturbed plan $\bar{\mathbf{A}}$ as one rollout\;
$(\bar{\mathbf{A}}, \boldsymbol{\sigma}) \leftarrow \textsc{Update}\big(\{\mathbf{A}^i, J^i\}\big)$ \tcp*[l]{optimizer-specific distribution update}
$\mathcal{S} \leftarrow$ spline through the updated control points $\bar{\mathbf{A}}$; execute until the next iteration\;
\end{algorithm}
\subsection{Policy-in-the-Loop Parallel Rollouts}\label{sec: rollouts}
\revised{Let $\mathbf{s}$ denote the state of the full system, stacking the robot's floating-base pose, joint positions, and their velocities together with the pose and velocity of the manipulated object, and let $\mathbf{u}$ denote the joint-level control applied to the robot. A standard multi-body simulator advances this state as
\begin{equation}
    \mathbf{s}_{k+1} = f(\mathbf{s}_k, \mathbf{u}_k).
\label{eq:dynamics}
\end{equation}
Previous sample-based MPC searches directly over sequences of $\mathbf{u}$~\cite{alvarez2025real, li2025_judo}, Fig.~\ref{fig:dynamics_comparison} (a). Our low-level policy $\pi$ instead maps the state and a command $\mathbf{c}$ to joint-level controls, $\mathbf{u} = \pi(\mathbf{s}, \mathbf{c})$, while the high-level planner samples actions $\mathbf{a}$ that are mapped to policy commands, $\mathbf{c} = g(\mathbf{a})$, as detailed in Sec.~\ref{sec: high-level mpc}. A key component of our system is therefore a physics rollout that composes the policy with the simulator, giving the policy-in-the-loop dynamics
\begin{equation}
    \mathbf{s}_{k+1} = f'(\mathbf{s}_k, \mathbf{a}_k) \triangleq f\big(\mathbf{s}_k, \, \pi(\mathbf{s}_k, g(\mathbf{a}_k))\big),
\label{eq:policy_in_the_loop}
\end{equation}
so that the planner searches over sequences of $\mathbf{a}$ rather than $\mathbf{u}$, Fig.~\ref{fig:dynamics_comparison} (b). In practice, $\pi$ is evaluated at its $50$Hz control rate and the resulting $\mathbf{u}$ is held constant across the intervening simulator steps.}
\input{figures/generalization_comparison.tex}
This design choice is critical for online sample efficiency for several reasons: First, this allows the sample-based MPC to sample in a lower-dimensional action space, mitigating the curse of dimensionality commonly seen for search algorithms. Second, with the low-level locomotion policy as a stabilizing controller, we avoid the well-known divergence issues associated with single-shooting methods applied to unstable dynamics that commonly occur with legged robots. Finally, because the whole-body control policy handles the locomotion part of the problem, this setup allows us to design much simpler cost function that only needs to encode manipulation objectives.

We implement the policy-in-the-loop parallel simulation by augmenting the CPU-based MuJoCo physics engine~\cite{todorov2012mujoco} in C++ using thread pool.
In Table \ref{tab:timing_comparison} we report the rollout times of $32$ parallel rollouts over $1.5$ seconds with and without the low-level policy on an Intel Core i7-12700K CPU. Indeed, the low-level policy adds overhead to the physics simulation, but the total times are faster than the $20$Hz or $50$ms update rate of the sample-based MPC. \revised{A natural question is whether the rollouts should run on a CPU or GPU. In general, CPU parallelization offers low latency at lower throughput, whereas GPU batching offers high throughput at the cost of higher latency. In real-time control we prefer low latency, with faster replanning often preferred over higher-reward solutions computed at slower rates~\cite{alvarez2025real}. GPU simulators~\cite{mujoco_warp2025} remain an interesting direction for scaling test-time search when hundreds to thousands of samples are necessary. The policy-inference overhead in Table~\ref{tab:timing_comparison} remains modest because RL amortizes optimization at training time, which is what makes policy-in-the-loop rollouts practical compared to embedding a QP-based policy in every rollout.}

\subsection{High-Level Sample-Based MPC}\label{sec: high-level mpc}
Naturally, the high-level sample-based MPC can sample over any action space supported as inputs to the low-level policy. In practice, we choose to sample actions over a compact, task-relevant control vector, then rely on the low-level policy to map this to a full whole-body command. Here, we describe the details for the Spot robot with the Relic policy, but similar principles apply to other robots and low-level policies. \revised{Concretely, we specify the action space of $\mathbf{a}$ and the map $g$ from planner actions to policy commands $\mathbf{c}$ introduced in Sec.~\ref{sec: rollouts}.}

For the Spot robot, the full $\mathbf{c}_{\mathrm{policy}} \in \mathbb{R}^{25}$ includes the $\mathrm{SE}(2)$ velocities for the base $\mathbf{c}_{\mathrm{base}} \in \mathbb{R}^3$, the arm joint-angle targets $\mathbf{c}_\mathrm{{arm}} \in \mathbb{R}^6$, gripper position $\mathbf{c}_\mathrm{{gripper}} \in \mathbb{R}^1$, leg joint angle targets $\mathbf{c}_\mathrm{{leg}} \in \mathbb{R}^3$ for all four legs, and torso pitch, roll, and height targets $\mathbf{\mathbf{c}_\mathrm{torso}} \in \mathbb{R}^3$. By default, we only sample over the base velocities $\mathbf{a}_{\mathrm{base}} \in \mathbb{R}^3$ and arm joint angle targets $\mathbf{a}_{\mathrm{arm}} \in \mathbb{R}^6$ for the arm, excluding the gripper. To pad remaining policy commands, we set leg targets to zero (the low-level policy automatically outputs leg joint commands to track desired base velocities) and use default values for the torso pitch, roll, and height targets and set the gripper to a closed position. This base setup is sufficient for many loco-manipulation tasks but can be further enhanced by reasoning over legs, torso, and gripper commands at test time for tasks that require multi-limb coordination or gripper dexterity.

To reason over the torso roll, pitch, and height commands, we can simply add those dimensions $\mathbf{a}_{\text{torso}} \in \mathbb{R}^3$ to the action space. To enable the robot to use front legs for manipulation, we can additionally sample over $\mathbf{a}_{\text{leg}} \in \mathbb{R}^7 = [\mathbf{s}_{\text{leg}}, \mathbf{c}_{\text{leg}}]$ where $\mathbf{s}_{\text{leg}} \in [-1, 1]$ is a selection variable and $\mathbf{c}_{\text{leg}} \in \mathbb{R}^6$ is the leg joint angle targets, then use the selection variable to mask the leg joint angle targets:
\begin{equation}
    \mathbf{c}_{\text{leg}}^{\text{masked}} = \begin{cases}
    [\mathbf{c}_{\text{leg}} [0{:}3], 0, 0, 0] & \text{if } s_{\text{leg}} < -0.5 \text{ (FL)} \\
    [0, 0, 0, \mathbf{c}_{\text{leg}}[3{:}6]] & \text{if } s_{\text{leg}} > 0.5 \text{ (FR)} \\
    [0, 0, 0, 0, 0, 0] & \text{else} \text{ (no legs)}
    \end{cases}
\label{eq:leg_masking}
\end{equation}

Note that it is also possible to use rear legs for manipulation, but our implementation chooses to only focus on front legs as they cover most practical loco-manipulation tasks. Similarly, we can control the gripper via a binary action variable $\mathbf{a}_{\text{gripper}} \in [-1, 1]$ that maps to open and close positions:
\begin{equation}
    \mathbf{c}_{\text{gripper}} = \begin{cases}
    \mathbf{c}_{\text{gripper close}} & \text{if } \mathbf{a}_{\text{gripper}} > 0  \\
    \mathbf{c}_{\text{gripper open}} & \text{if } \mathbf{a}_{\text{gripper}} \leq 0
    \end{cases}
\label{eq:gripper_mapping}
\end{equation}
This design allows us to flexibly choose task-specific action spaces to be any combination of $\mathbf{a} = [\mathbf{a}_{\text{base}}, \mathbf{a}_{\text{arm}}, \mathbf{a}_{\text{torso}}, \mathbf{a}_{\text{leg}}, \mathbf{a}_{\text{gripper}}]$ for the sample-based MPC. Additionally, the selection variables allow the controller to reason over different manipulation modalities at test time. Note that our sample-based MPC naturally reasons over both continuous and discrete decision variables which would be challenging for traditional gradient-based optimal control algorithms.

For the G1 humanoid robot, we similarly sample over the $\mathrm{SE}(2)$ velocities for the torso and the arm joint angle targets for the arm. While it is straight forward to enable more inter-limb coordination by using a more steerable underlying low-level policy~\cite{liao2025beyondmimic}, we leave this investigation as a future direction.

\revised{In our experience, the low-level policy interface properties that matter most for planning are 1) a low-dimensional command space, 2) consistent closed-loop command effects (so sampled perturbations produce predictable behavior variations), and 3) stability under rapidly changing commands. A policy conditioned on per-limb end-effector targets is an appealing alternative: it would give the planner direct authority over contact locations and further shrink the search space, at the cost of being harder to train with high tracking fidelity near contact, where forces rather than positions dominate. As more steerable whole-body trackers emerge~\cite{liao2025beyondmimic, li2026omnitrackgeneralmotiontracking}, we expect the planner-over-policy recipe to benefit directly.}

\revised{\subsection{Planner Implementation Details}\label{sec: implementation details}
Algorithm~\ref{alg:sumo_cem} summarizes one planning iteration. Our sampling distribution is a Gaussian with a \emph{diagonal} covariance: noise is sampled independently per spline control point and action dimension, with no cross coupling. Actions are normalized---sampled in normalized units, clipped to the low-level policy's command limits, and affinely mapped to physical commands. The discrete variables ($\mathbf{s}_{\text{leg}}$, $\mathbf{a}_{\text{gripper}}$) are treated as continuous values and only thresholded via \eqref{eq:leg_masking} and \eqref{eq:gripper_mapping}. The planner is warm-started by time-shifting the previous plan to seed the distribution mean, and the unperturbed shifted plan is retained as one of the rollouts. Finally, commands are linearly interpolated between control points by default with options for zeroth-order or cubic interpolations.
The sampling noise follows the linear schedule of Sec.~\ref{sec: analysis}: the beginning of the plan is executed before the next replan and has been refined by previous iterations, so it benefits from low noise, while the tail of the horizon is least optimized and benefits from wider exploration.

The distribution update in Algorithm~\ref{alg:sumo_cem} produces a new mean $\bar{\mathbf{A}}$ and noise scale $\boldsymbol{\sigma}$ from the sampled control sequences and their costs. The per-node noise schedule (Sec.~\ref{sec: analysis}) shapes $\boldsymbol{\sigma}$ across the horizon at sampling time. Note that sample-based planners (CEM, MPPI, CMA-ES, etc.) are related zeroth-order policy-improvement updates and have been analyzed under a common framework in recent work~\cite{yi2024covompc, qiu2025zerothorder}, so the specific rule is not central to our framework. We adopt CEM for its simplicity, and our open-source implementation exposes other planners as drop-in alternatives.}

%% file: figures/generalization_comparison.tex
\begin{figure}[!t]
    \centering
    \begin{minipage}[b]{1.0\linewidth}
        \centering
        \textbf{(a) Move Object Tasks}
        \begin{tikzpicture}
            \begin{axis}[
                ybar,
                bar width=0.3cm,
                width=1.0\linewidth,
                height=4cm,
                ylabel={Success Rate},
                symbolic x coords={Box, Chair, Cone, Tire, Tire Rack},
                xtick=data,
                xticklabels={},
                ymin=-0.05, ymax=1.05,
                legend style={at={(0.5,-0.2)}, anchor=north, legend columns=-1},
                ymajorgrids=true,
                grid style=dashed,
                area legend,
            ]

            \addplot+[
                fill={rgb,255:red,244; green,176; blue,34}, 
                fill opacity=0.75,
                draw={rgb,255:red,244; green,176; blue,34}
            ]
            coordinates {(Box,0.85) (Chair,0.95) (Cone,1.0) (Tire,1.0) (Tire Rack,1.0)};
            \addplot+[
                fill={rgb,255:red,174; green,78; blue,255}, 
                fill opacity=0.75,
                draw={rgb,255:red,174; green,78; blue,255},
            ]
            coordinates {(Box,1.0) (Chair,0.4) (Cone,0.62) (Tire,0.2) (Tire Rack,0.1)};
            \addplot+[
                fill={rgb,255:red,19; green,12; blue,44}, 
                fill opacity=0.65,
                draw={rgb,255:red,19; green,12; blue,44},
                draw opacity=0.70,
            ]
            coordinates {(Box,0.96) (Chair,0.56) (Cone,0.0) (Tire,0.0) (Tire Rack,0.02)};
            \end{axis}
        \end{tikzpicture}
        \label{fig:generalization_push}
    \end{minipage}

    \vspace{-5pt}

    \begin{minipage}[b]{1.0\linewidth}
        \centering
        \textbf{(b) Upright Object Tasks}
        \begin{tikzpicture}
            \begin{axis}[
                ybar,
                bar width=0.3cm,
                width=1.0\linewidth,
                height=4cm,
                ylabel={Success Rate},
                symbolic x coords={Box, Chair, Cone, Tire, Tire Rack},
                xtick=data,
                ymin=-0.05, ymax=1.05,
                legend style={at={(0.5,-0.25)}, anchor=north, legend columns=-1},
                ymajorgrids=true,
                grid style=dashed,
                area legend,
            ]
            \addplot+[
                fill={rgb,255:red,244; green,176; blue,34}, 
                fill opacity=0.75,
                draw={rgb,255:red,244; green,176; blue,34}
            ]
            coordinates {(Box,1.0) (Chair, 0.95) (Cone, 1.0) (Tire,0.95) (Tire Rack,0.95)};
            \addplot+[
                fill={rgb,255:red,174; green,78; blue,255}, 
                fill opacity=0.75,
                draw={rgb,255:red,174; green,78; blue,255},
            ]
            coordinates {(Box,0.04) (Chair,0.14) (Cone,0.04) (Tire,0.16) (Tire Rack,0.02)};
            \addplot+[
                fill={rgb,255:red,19; green,12; blue,44}, 
                fill opacity=0.65,
                draw={rgb,255:red,19; green,12; blue,44},
                draw opacity=0.70,
            ]
            coordinates {(Box,0.0) (Chair,0.18) (Cone, 0.14) (Tire,0.02) (Tire Rack,0.16)};
            \legend{Sumo (Ours), E2E RL, Hierarchical RL}
            \end{axis}
        \end{tikzpicture}
        \label{fig:generalization_upright}
    \end{minipage}

    \caption{Top: comparison of Sumo (yellow, ours), E2E RL (purple), and hierarchical RL (navy, HRL) on pushing five different objects to a goal. Sumo generalizes to new objects by replacing the object model at test time, whereas E2E RL and HRL policies trained only on box pushing fail on the other objects. Bottom: Sumo generalizes to uprighting objectives by changing the planner cost at test time, whereas the same E2E RL and HRL policies fail without additional training.}
    \vspace{-20pt}
    \label{fig:generalization}
\end{figure}

%% file: sections/experiments.tex

\input{figures/hierarchy_comparison.tex}
\section{Experimental Analysis}\label{sec: analysis}
In this section, we use simulation experiments in MuJoCo~\cite{todorov2012mujoco} to analyze three advantages of Sumo: hierarchical structure simplifies loco-manipulation, test-time search enables generalization, and Sumo achieves higher sample efficiency than a hierarchical RL policy.

\subsection{Task Definition and Experiment Setup}
In this section, we consider two types of loco-manipulation tasks on the Spot quadruped robot: moving an object to a goal position (Move) and reorienting an object to the upright orientation (Upright). In both cases, we consider five objects covering a range of size, weight, and geometry representative of real-world scenarios: a $1.5$ kg box, a chair, a car tire, a tire rack, and a traffic cone.

For Move tasks, we consider success if the object is within $0.1$ meters of the goal and velocity is less than $0.05$ m/s within $30$ seconds. For Upright tasks, we consider task success if the object is within $0.1$ radians of the upright orientation and angular velocity is less than $0.05$ rad/s within $30$ seconds.

To simplify the analysis, we restrict the Sumo high-level planner to only control the torso $\mathrm{SE}(2)$ velocities and joint angle targets for the arm in this section and leave exploring more expressive policy behaviors to Sec. \ref{sec: experiments}. Additionally, we design a minimal cost function for Move tasks:
\begin{align}
    & J_{\text{Move}} =  \nonumber \\
    & w_{\text{goal}} \cdot \| \mathbf{p}_{\text{obj}} - \mathbf{p}_{\text{goal}} \|_2 + w_{\text{gripper}} \cdot \| \mathbf{p}_{\text{gripper}} - \mathbf{p}_{\text{obj}} \|_2 \nonumber \\
    & + w_{\text{vel}} \cdot \| \mathbf{v}_{\text{obj}} \|_2
\label{eq:move_cost_function}
\end{align}
where $\mathbf{p}_{\text{obj}}$, $\mathbf{p}_{\text{goal}}$, and $\mathbf{p}_{\text{gripper}}$ are the positions of the object, goal, and gripper in world frame, respectively, $\mathbf{v}_{\text{obj}}$ is the linear velocity of the object, and $w_{\text{goal}}, w_{\text{gripper}}, w_{\text{vel}}$ are weighting coefficients.
Similarly, for Upright tasks:
{\color{blue}
\begin{align}
    & J_{\text{Upright}} =  \nonumber \\
    & w_{\text{Upright}} \cdot \left( 1 - \hat{\mathbf{z}}_{\text{obj}} \cdot \hat{\mathbf{z}}_{\text{world}} \right) + w_{\text{gripper}} \cdot \| \mathbf{p}_{\text{gripper}} - \mathbf{p}_{\text{obj}} \|_2
\label{eq:upright_cost_function}
\end{align}}%
\revised{where $\hat{\mathbf{z}}_{\text{obj}}$ is the object body axis that points upward in the target upright pose, expressed in world frame, and $\hat{\mathbf{z}}_{\text{world}} = [0, 0, 1]^{\top}$. This axis-alignment cost vanishes exactly at the upright orientation and avoids the ambiguities of comparing quaternions directly.}

For all Sumo experiments, we use a Cross-Entropy Method (CEM) optimizer with a prediction horizon of $1.5$ seconds, $32$ rollouts, and update the action distribution with three elite samples at a frequency of $20$Hz. We sample over four spline control points along the prediction horizon and interpolate the remaining actions. Finally, we linearly schedule the noise variance ramping from $0.02$ to $0.6$ over the prediction horizon. \revised{Implementation details of the optimizer are provided in Sec.~\ref{sec: implementation details}. The cost weights are shared across all five objects without per-object tuning: $w_{\text{goal}}{=}60$, $w_{\text{gripper}}{=}4$, $w_{\text{vel}}{=}20$ for Move, and $w_{\text{Upright}}{=}100$, $w_{\text{gripper}}{=}0.5$ for Upright.}
All experiments are conducted synchronously on a desktop computer equipped with an Intel Core i7-12700K CPU and 64GB of RAM. Current robot and object states are provided by the ground-truth simulator.

\subsection{Hierarchical Structure Simplifies Loco-Manipulation}
We first evaluate whether hierarchical structure simplifies loco-manipulation. Fig. \ref{fig:hierarchical_structure} compares Sumo against both real-time MPC (E2E MPC) and end-to-end RL (E2E RL) on the Move tasks. This comparison isolates the benefit of planning in the command space of a pre-trained whole-body controller rather than optimizing directly over joint-level actions or learning a task-specific end-to-end policy.

For the E2E MPC baseline, we use the same CEM optimizer, which is used in Sumo and common for locomotion and manipulation~\cite{li2025_judo, howell2022mjpc, alvarez2025real}, whose action space is the joint-level commands of the robot. We design rewards to include terms for both locomotion and manipulation objectives:
\begin{align}
    & J_{\text{E2E MPC}} = J_{\text{Locomotion}} + J_{\text{Move}}
\label{eq:e2e_mpc_cost_function}
\end{align}
where $J_{\text{Locomotion}}$ is the locomotion-related cost from~\cite{howell2022mjpc} and $J_{\text{Move}}$ is defined in \eqref{eq:move_cost_function}. Additionally, we allow the E2E MPC to update at $50$Hz to match Sumo's low-level policy frequency.

For the E2E RL baseline, we train a state-based, goal-conditioned single-task policy that controls the $19$-DoF joints of Spot to move an object to a goal location. We train the policy using PPO in MJLab~\cite{mjlab2025} with $4096$ parallel environments for $5000$ iterations. We use $15$ reward terms, including terms that encourage natural walking through a phase-based gait and foot-height schedule, progress toward the manipulation goal, and shaping that encourages the gripper to approach the object and the robot to stay behind it. The policy also receives bonuses when it reaches the goal and keeps the object within $0.2$m, which we count as success. As with Sumo, we tune the reward until the policy reliably solves the box task, then apply the same setup to four additional objects without further task-specific reward engineering. In both baselines, we make a best-faith effort to tune hyperparameters to achieve the best performance.

We find that Sumo successfully solves all five Move tasks with at least $80\%$ success across $20$ evaluation trials, with some tasks at $100\%$ success. In contrast, E2E MPC struggles to consistently solve these tasks, achieving at most $50\%$ success. E2E RL is competitive on the box, chair, and cone, but degrades sharply on more complex geometries such as the tire and tire rack. We summarize these results in Fig. \ref{fig:hierarchical_structure}.

These experiments show that hierarchical structure simplifies loco-manipulation in two complementary ways. Compared to E2E MPC, planning in the task space of a pre-trained low-level policy reduces the effective search space and stabilizes the robot's contact-rich dynamics. Compared to E2E RL, Sumo achieves similar or better success with only $3$ reward terms and no task-specific retraining or tuning, whereas E2E RL requires $15$ reward terms and about $2$ hours of GPU compute per task. This comparison also motivates the next question: once a low-level whole-body controller is fixed, should the high-level policy be learned or optimized online at test time?

\begin{figure}
    \centering
    \input{figures/compute_time_comparison.tex}
    \caption{Maximum success rate vs compute time comparison between Sumo (blue) and hierarchical RL (red) baseline during hyperparameter Bayesian optimization. Solid lines represent the mean across 5 optimization runs, and the shaded region shows $\pm 1$ standard deviation. Individual success rates are shown as scatter points. Our method achieves similar asymptotic performance compared to hierarchical RL in an order of magnitude less wall-clock time.
    }
    \label{fig:compute_time_comparison}
\end{figure}
\input{figures/spot_tasks_freeze_frame.tex}
\subsection{Test-Time Search Enables Generalization}
We next ask whether the high-level decision-making module should be learned or optimized online. Our experiments show that keeping this layer online is what enables generalization in our setting. Here, generalization means reusing the same learned controller on new objects and new task objectives by changing only the planner's object model or cost at test time, without retraining. This distinction matters because RL policies are typically trained for a \emph{known} distribution of environments and objects, whereas open-world manipulation introduces geometries, masses, and contact conditions that are difficult to model comprehensively at training time.

Fig. \ref{fig:generalization} (a) shows that a hierarchical RL policy trained to move a $1.5$ kg box to a goal does not generalize reliably once object geometry and weight move beyond its training distribution, even after randomizing the box size, weight, and friction. While the hierarchical RL policy achieves a $100\%$ success rate on this training task, its performance degrades quickly on objects such as the tire and traffic cone. In contrast, our method can generalize to different objects \emph{without additional training} by simply replacing the object model. All tasks here use the same cost function and hyperparameters described in Eq. \eqref{eq:move_cost_function}.

\revised{The hierarchical RL policy commands the \emph{same} pretrained low-level policy as Sumo through the same command space, and observes the robot state, object pose and velocity, and goal. Neither high-level policy is conditioned on explicit object parameters such as geometry or inertia. Object information enters Sumo only through the planner's simulation model and has no analogue in the RL training setting.}

Fig. \ref{fig:generalization} (b) shows the same advantage for different task objectives. By changing the planner cost function at test time, similar in spirit to guidance for diffusion models~\cite{janner2022diffuser}, we replace the goal-directed objective with the orientation objective in Eq. \eqref{eq:upright_cost_function} and enable the robot to reliably upright different objects without additional training. In contrast, the same RL policy cannot be steered to upright different objects without additional training.
\vspace{-5pt}
\subsection{Sample-Based MPC Provides High Practical Efficiency}
Finally, we compare the practical iteration speed of sample-based MPC and RL during hyperparameter tuning. The goal of this experiment is not a hardware-normalized compute comparison, but rather a wall-clock comparison of the tuning loop practitioners would run in practice. In Fig. \ref{fig:compute_time_comparison}, we mimic this process by performing a Bayesian optimization over the cost weights in \eqref{eq:move_cost_function} for the Move Box task for both Hierarchical RL and Sumo to find the weights that maximize the success rate. We perform five independent optimization runs and track the maximum success rate over time across $50$ trials. Both methods achieve similar asymptotic performance, but Sumo reaches the same performance with an order-of-magnitude less compute time.
For Sumo, we measure the compute time as CPU hours on a desktop computer equipped with an Intel Core i7-12700K CPU and 64GB of RAM, and RL training time is measured as GPU hours on an NVIDIA RTX A6000 GPU.

\section{Autonomous Demonstrations with Sumo} \label{sec: experiments}
In this section, we move from controlled simulation analysis to full-task demonstrations on eight Spot tasks in the real world and four G1 tasks in simulation. Together, these case studies test whether the same hierarchical framework can handle diverse manipulation modes and object interactions beyond the simplified analysis tasks. Unlike the simplified cost functions used in Sec. \ref{sec: analysis}, these demonstrations use task-specific cost functions tailored to each task; full descriptions are provided in the \ifarxiv appendix\else\revised{supplementary} appendix\fi. \revised{The additional terms address what the idealized analysis setting does not exercise: safety terms penalize plans that risk falls, grasp-detection terms let the planner verify that the gripper actually holds the object, and approach/standoff terms shape how the robot engages the object, biasing the search away from local minima that are harmless in simulation but costly on hardware.}
\vspace{-5pt}
\subsection{Spot Quadruped Loco-Manipulation}
In the first case study, we evaluate Sumo on a diverse set of challenging loco-manipulation tasks on the Boston Dynamics Spot quadruped. These tasks stress three recurring difficulties: objects that are larger than the robot or heavier than the arm payload, contact conditions and geometries that create large sim-to-real gaps, and distinct manipulation modes such as uprighting, stacking, dragging, and pushing. The tasks and corresponding performance are summarized in Table \ref{tab:spot_hardware_results}.

For all hardware deployments, we use the same set of optimizer parameters as detailed in Sec. \ref{sec: analysis}. 
To estimate the robot and object states, we fuse the Spot robot joint encoder reading at $333$Hz with motion-capture (MoCap) measurements for torso and object positions and orientations at $120$Hz using a low-pass filter.
We compute the dynamics rollouts and CEM updates asynchronously on a desktop computer equipped with an AMD Threadripper Pro 5995WX CPU with $64$ cores, which sends joint-level commands to the Spot robot via WiFi. \revised{State estimation thus relies on external MoCap and planning runs off-board: these experiments demonstrate the control approach rather than full onboard autonomy (see Sec.~\ref{sec: conclusions}).}
\begin{table}
    \centering
    \caption{Real-world completion time and success rate performance of Spot loco-manipulation tasks across $10$ trials.}
    \begin{tabular}{l c c c}
        \toprule
        Task Name & Completion Time & Successes & Time Limit \\
        \midrule
        Tire Upright & $9.2 \pm 4.7$s & $10/10$ & $30$s \\
        Barrier Upright & $10.5 \pm 7.1$s & $9/10$ & $30$s \\
        Cone Upright & $10.2 \pm 7.9$s & $9/10$ & $30$s \\
        Chair Upright & $27.3 \pm 19.1$s & $8/10$ & $60$s \\
        Tire Stack & $16.5 \pm 8.4$s & $8/10$ & $30$s \\
        Barrier Drag & $20.2 \pm 6.7$s & $9/10$ & $30$s \\
        Tire Rack Drag & $19.1 \pm 6.2$s & $9/10$ & $30$s \\
        Rugged Box Push & $38.3 \pm 16.9$s & $10/10$ & $90$s \\
        \bottomrule
    \end{tabular}
    \label{tab:spot_hardware_results}
    \vspace{-20pt}
\end{table}

Below, we briefly describe each task to provide qualitative context for the quantitative results in Table \ref{tab:spot_hardware_results}.

\noindent\textbf{Tire Upright:}
In Fig. \ref{fig:spot_tasks_freeze_frame} (a), the robot is tasked to upright a tire of $15$kg, which is heavier than the maximum lifting capacity of $11$kg of the Spot arm.
Additionally, the rubber tire presents hard-to-model geometries and friction properties, causing a large sim-to-real gap. The robot solves the task using a combination of its arm, torso, and legs.

\noindent\textbf{Crowd Barrier Upright:}
In Fig. \ref{fig:spot_tasks_freeze_frame} (b), the robot is tasked to upright a crowd barrier of $16$kg from a lying position using its arm and gripper.

\noindent\textbf{Traffic Cone Upright:}
In Fig. \ref{fig:spot_tasks_freeze_frame} (c), the robot is tasked to upright a traffic cone of $3.5$kg using its arm.

\noindent\textbf{Chair Upright:}
In Fig. \ref{fig:spot_tasks_freeze_frame} (d), the robot is tasked to upright a chair of $16.5$kg using its arm and body.

\noindent\textbf{Tire Stack:}
In Fig. \ref{fig:spot_tasks_freeze_frame} (e), the robot is tasked to stack a $15$kg tire on top of another tire. In addition to large weight and deformation, the friction coefficient between the rubber tires is high, which current simulators struggle to simulate accurately, causing an even larger sim-to-real gap. The robot solves the task using a combination of its arm, torso, and legs.

\noindent\textbf{Crowd Barrier Drag:}
In Fig. \ref{fig:spot_tasks_freeze_frame} (f), the robot is tasked to drag a crowd barrier of $15$kg in upright orientation. The robot uses the gripper to grasp the barrier and drags it towards a goal position with its arm and body.

\noindent\textbf{Tire Rack Drag:}
In Fig. \ref{fig:spot_tasks_freeze_frame} (g), the robot is tasked to drag a tire rack of $10$kg. The robot uses the gripper to grasp the tire rack and drags it towards a goal position.

\noindent\textbf{Rugged Box Push:}
In Fig. \ref{fig:spot_tasks_freeze_frame} (h), the robot is tasked to push a $20$kg box to a goal using its arm and body.

\vspace{-5pt}
\subsection{G1 Humanoid Loco-Manipulation}
In the second case study, we show that Sumo can also be applied to humanoid robots. Through simulated experiments using a Unitree G1, our method uses a pre-trained locomotion policy that is widely available~\cite{mjlab2025}, and solves loco-manipulation tasks such as pushing a box and opening a door. Simulation performances are summarized in Table \ref{tab:g1_simulation_results}.
\input{figures/g1_tasks_freeze_frame.tex}

\noindent\textbf{Box Pushing}
In Fig. \ref{fig:g1_tasks_freeze_frame} (a), the robot is tasked to push a $10$kg box to a target position using its arms and body.

\noindent\textbf{Chair Pushing}
In Fig. \ref{fig:g1_tasks_freeze_frame} (b), the robot is tasked with pushing a $16.5$kg chair to a goal using its arms and body.

\noindent\textbf{Door Opening}
In Fig. \ref{fig:g1_tasks_freeze_frame} (c), the robot is asked to open a door using its arm and body.

\noindent\textbf{Table Pushing}
In Fig. \ref{fig:g1_tasks_freeze_frame} (d), the robot is tasked with pushing a $10$kg table to a goal using its arms and body.

\begin{table}
    \centering
    \caption{Performance of G1 humanoid loco-manipulation tasks in simulation.}
    \begin{tabular}{l c c c}
        \toprule
        Task Name& Completion Time (s) & Successes & Time Limit \\
        \midrule
        G1 Box Push & 11.83 $\pm$ 2.97 s & $9$/$10$ & $30$s \\
        G1 Door Open & 4.73 $\pm$ 0.98 s & $10$/$10$ & $30$s \\
        G1 Chair Push & 6.86 $\pm$ 0.288 s & $10$/$10$ & $30$s \\
        G1 Table Push & 4.86 $\pm$ 1.65 s & $8$/$10$ & $30$s \\
        \bottomrule
    \end{tabular}
    \vspace{-20pt}
    \label{tab:g1_simulation_results}
\end{table}

\vspace{-5pt}
\revised{\subsection{Open-Source Benchmark}\label{sec: benchmark}
We release the tasks in this paper as an open-source benchmark for whole-body loco-manipulation. The release contains simulation environments for the analysis tasks of Sec.~\ref{sec: analysis} and the demonstration tasks of this section on both robots; MJCF models of the robots and all manipulated objects, with masses and geometries matched to our physical objects; pretrained low-level policies integrated into the policy-in-the-loop C++ rollout backends of Sec.~\ref{sec: methods}; complete cost implementations with all weights, success criteria, and reset distributions; CEM, MPPI, and predictive sampling as interchangeable optimizers, with an interactive visualizer; and utilities to record and replay rollout trajectories. Tasks share a common interface, so new objects and objectives require only an object model and a cost function. \ifarxiv The code is publicly available at \url{https://github.com/rai-opensource/sumo}, and videos of all tasks are available on the project page at \url{https://sumo.rai-inst.com}.\else The code accompanies this submission as supplementary material; the public repository will be linked in the final version.\fi}

%% file: figures/hierarchy_comparison.tex
\begin{figure}[!t]
    \centering
    \begin{tikzpicture}
        \begin{axis}[
            ybar,
            bar width=0.3cm,
            width=1.0\linewidth,
            height=4cm,
            ylabel={Success Rate},
            symbolic x coords={Box, Chair, Cone, Tire, Tire Rack},
            xtick=data,
            ymin=-0.05, ymax=1.05,
            legend style={at={(0.5,-0.25)}, anchor=north, legend columns=-1},
            ymajorgrids=true,
            grid style=dashed,
            area legend,
        ]
        \addplot+[
            fill={rgb,255:red,244; green,176; blue,34}, 
            fill opacity=0.75,
            draw={rgb,255:red,244; green,176; blue,34}
        ]
        coordinates {(Box,0.85) (Chair,0.95) (Cone, 1.0) (Tire,1.0) (Tire Rack, 1.0)};
        \addplot+[
            fill={rgb,255:red,174; green,78; blue,255}, 
            fill opacity=0.75,
            draw={rgb,255:red,174; green,78; blue,255},
        ]
        coordinates {(Box,0.98) (Chair,1.0) (Cone, 1.0) (Tire,0.16) (Tire Rack,0.0)};
        \addplot+[
            fill={rgb,255:red,19; green,12; blue,44}, 
            fill opacity=0.65,
            draw={rgb,255:red,19; green,12; blue,44},
            draw opacity=0.70,
        ]
        coordinates {(Box,0.0) (Chair,0.3) (Cone, 0.1) (Tire,0.5) (Tire Rack,0.3)};
        \legend{Sumo (Ours), E2E RL, E2E MPC}
        \end{axis}
    \end{tikzpicture}
    \caption{Comparing Sumo (ours, yellow) to end-to-end RL (purple) and MPC (navy) on five loco-manipulation tasks that ask the robot to move an object to a goal. Sumo achieves high success across all objects. End-to-end RL is competitive on the box, chair, and cone but degrades sharply on the tire and tire rack, while end-to-end MPC struggles across the board.}
    \vspace{-15pt}
    \label{fig:hierarchical_structure}
\end{figure}

%% file: figures/compute_time_comparison.tex
\begin{tikzpicture}
\begin{axis}[
    width=\linewidth,
    height=4.5cm,
    xlabel={Cumulative Compute Hours in Log Scale},
    ylabel={Max Success Rate},
    xmode=log,
    xmin=0.0637, xmax=60.7,
    ymin=0, ymax=1.0,
    grid=major,
    legend pos=south east,
    legend style={font=\small, fill opacity={0.75}},
    xminorticks=false,
    set layers, 
]

\addplot[
    color={rgb,255:red,244; green,176; blue,34}, 
    fill opacity=0.75,
    mark=*,
    mark size=0.8pt,
    only marks,
    forget plot
]
coordinates {
    (0.0779, 0.000000)
    (0.1529, 0.000000)
    (0.1950, 1.000000)
    (0.2692, 0.100000)
    (0.3421, 0.000000)
    (0.4075, 0.500000)
    (0.4572, 0.700000)
    (0.5225, 0.500000)
    (0.6050, 0.000000)
    (0.6897, 0.000000)
    (0.7617, 0.400000)
    (0.8200, 0.600000)
    (0.8874, 0.700000)
    (0.9566, 0.700000)
    (1.0276, 0.400000)
    (1.0731, 0.900000)
    (1.1362, 0.600000)
    (1.1989, 0.700000)
    (1.2732, 0.300000)
    (1.3334, 0.900000)
    (1.3979, 0.600000)
    (1.4602, 0.800000)
    (1.5149, 0.700000)
    (1.5838, 0.500000)
    (1.6320, 0.900000)
    (1.6929, 0.400000)
    (1.7466, 0.600000)
    (1.8095, 0.500000)
    (1.8835, 0.000000)
    (1.9574, 0.000000)
    (2.0203, 0.300000)
    (2.0827, 0.500000)
    (2.1387, 0.600000)
    (2.2067, 0.700000)
    (2.2684, 0.600000)
    (2.3161, 0.700000)
    (2.3764, 0.700000)
    (2.4496, 0.100000)
    (2.5177, 0.500000)
    (2.5831, 0.400000)
    (2.6588, 0.000000)
    (2.7233, 0.600000)
    (2.7862, 0.600000)
    (2.8390, 0.700000)
    (2.8986, 0.500000)
    (2.9499, 0.800000)
    (3.0135, 0.500000)
    (3.0679, 0.600000)
    (3.1222, 0.800000)
    (3.1733, 0.900000)
};

\addplot[
    color={rgb,255:red,244; green,176; blue,34}, 
    fill opacity=0.75,
    mark=*,
    mark size=0.8pt,
    only marks,
    forget plot
]
coordinates {
    (0.0886, 0.000000)
    (0.1766, 0.000000)
    (0.2494, 0.100000)
    (0.3376, 0.100000)
    (0.4267, 0.000000)
    (0.4908, 0.600000)
    (0.5632, 0.400000)
    (0.6260, 0.700000)
    (0.7001, 0.600000)
    (0.7820, 0.000000)
    (0.8454, 0.400000)
    (0.9111, 0.600000)
    (0.9739, 0.700000)
    (1.0446, 0.500000)
    (1.0970, 0.700000)
    (1.1753, 0.000000)
    (1.2575, 0.000000)
    (1.3301, 0.400000)
    (1.4103, 0.000000)
    (1.4884, 0.000000)
    (1.5486, 0.600000)
    (1.6122, 0.400000)
    (1.6714, 0.700000)
    (1.7401, 0.500000)
    (1.8199, 0.200000)
    (1.8795, 0.700000)
    (1.9352, 0.800000)
    (1.9925, 0.700000)
    (2.0425, 0.900000)
    (2.0991, 0.800000)
    (2.1419, 0.900000)
    (2.2053, 0.700000)
    (2.2553, 0.800000)
    (2.3159, 0.600000)
    (2.3796, 0.800000)
    (2.4410, 0.600000)
    (2.5183, 0.000000)
    (2.5823, 0.500000)
    (2.6404, 0.500000)
    (2.6998, 0.700000)
    (2.7533, 0.900000)
    (2.8312, 0.000000)
    (2.8906, 0.700000)
    (2.9468, 0.600000)
    (3.0172, 0.500000)
    (3.0942, 0.000000)
    (3.1721, 0.000000)
    (3.2218, 0.900000)
    (3.2993, 0.000000)
    (3.3510, 0.700000)
};

\addplot[
    color={rgb,255:red,244; green,176; blue,34}, 
    fill opacity=0.75,
    mark=*,
    mark size=0.8pt,
    only marks,
    forget plot
]
coordinates {
    (0.0637, 0.400000)
    (0.1337, 0.000000)
    (0.2034, 0.000000)
    (0.2586, 0.700000)
    (0.3010, 0.900000)
    (0.3652, 0.300000)
    (0.4070, 0.900000)
    (0.4770, 0.000000)
    (0.5424, 0.300000)
    (0.6124, 0.000000)
    (0.6776, 0.200000)
    (0.7208, 0.800000)
    (0.7712, 0.500000)
    (0.8337, 0.500000)
    (0.8853, 0.700000)
    (0.9350, 0.700000)
    (1.0055, 0.000000)
    (1.0598, 0.500000)
    (1.1297, 0.000000)
    (1.1928, 0.300000)
    (1.2640, 0.000000)
    (1.3158, 0.700000)
    (1.3540, 0.800000)
    (1.4169, 0.700000)
    (1.4668, 0.700000)
    (1.5267, 0.800000)
    (1.5776, 0.600000)
    (1.6261, 0.700000)
    (1.6952, 0.000000)
    (1.7514, 0.500000)
    (1.8127, 0.400000)
    (1.8588, 0.700000)
    (1.9121, 0.800000)
    (1.9651, 0.700000)
    (2.0242, 0.700000)
    (2.1058, 0.000000)
    (2.1863, 0.000000)
    (2.2552, 0.500000)
    (2.3353, 0.000000)
    (2.4121, 0.300000)
    (2.4758, 0.600000)
    (2.5235, 0.900000)
    (2.6016, 0.200000)
    (2.6642, 0.700000)
    (2.7386, 0.500000)
    (2.7819, 0.900000)
    (2.8310, 0.800000)
    (2.8827, 0.900000)
    (2.9450, 0.700000)
    (3.0120, 0.700000)
};

\addplot[
    color={rgb,255:red,244; green,176; blue,34}, 
    fill opacity=0.75,
    mark=*,
    mark size=0.8pt,
    only marks,
    forget plot
]
coordinates {
    (0.0698, 0.100000)
    (0.1366, 0.300000)
    (0.1923, 0.500000)
    (0.2417, 0.800000)
    (0.3054, 0.600000)
    (0.3752, 0.000000)
    (0.4251, 0.700000)
    (0.4889, 0.600000)
    (0.5589, 0.000000)
    (0.6188, 0.400000)
    (0.6838, 0.300000)
    (0.7317, 0.700000)
    (0.7863, 0.500000)
    (0.8312, 0.600000)
    (0.8857, 0.500000)
    (0.9566, 0.000000)
    (1.0096, 0.500000)
    (1.0790, 0.000000)
    (1.1492, 0.000000)
    (1.1937, 0.700000)
    (1.2585, 0.300000)
    (1.2958, 0.900000)
    (1.3484, 0.600000)
    (1.3952, 0.700000)
    (1.4418, 0.900000)
    (1.4927, 0.800000)
    (1.5430, 0.800000)
    (1.6001, 0.700000)
    (1.6518, 0.800000)
    (1.7053, 0.500000)
    (1.7484, 0.700000)
    (1.8035, 0.400000)
    (1.8663, 0.400000)
    (1.9282, 0.400000)
    (1.9823, 0.600000)
    (2.0281, 0.800000)
    (2.0816, 0.500000)
    (2.1356, 0.600000)
    (2.1956, 0.500000)
    (2.2662, 0.000000)
    (2.3197, 0.600000)
    (2.3622, 0.800000)
    (2.4167, 0.500000)
    (2.4633, 0.700000)
    (2.5150, 0.800000)
    (2.5844, 0.200000)
    (2.6464, 0.400000)
    (2.6955, 0.800000)
    (2.7669, 0.000000)
    (2.8222, 0.700000)
};

\addplot[
    color={rgb,255:red,244; green,176; blue,34}, 
    fill opacity=0.75,
    mark=*,
    mark size=0.8pt,
    only marks,
    forget plot
]
coordinates {
    (0.0779, 0.000000)
    (0.1401, 0.600000)
    (0.2181, 0.000000)
    (0.2739, 0.700000)
    (0.3351, 0.600000)
    (0.4124, 0.000000)
    (0.4901, 0.000000)
    (0.5681, 0.000000)
    (0.6438, 0.000000)
    (0.7075, 0.600000)
    (0.7815, 0.000000)
    (0.8413, 0.600000)
    (0.9061, 0.400000)
    (0.9549, 0.700000)
    (0.9998, 0.700000)
    (1.0602, 0.600000)
    (1.1135, 0.800000)
    (1.1752, 0.500000)
    (1.2343, 0.500000)
    (1.2884, 0.500000)
    (1.3510, 0.500000)
    (1.4035, 0.800000)
    (1.4610, 0.600000)
    (1.5117, 0.500000)
    (1.5719, 0.500000)
    (1.6262, 0.600000)
    (1.6972, 0.000000)
    (1.7586, 0.500000)
    (1.8079, 0.700000)
    (1.8779, 0.000000)
    (1.9472, 0.000000)
    (2.0027, 0.600000)
    (2.0608, 0.400000)
    (2.1175, 0.600000)
    (2.1735, 0.700000)
    (2.2235, 0.700000)
    (2.2747, 0.500000)
    (2.3450, 0.000000)
    (2.4157, 0.000000)
    (2.4777, 0.400000)
    (2.5330, 0.700000)
    (2.5906, 0.400000)
    (2.6367, 0.700000)
    (2.6767, 0.900000)
    (2.7475, 0.000000)
    (2.8121, 0.300000)
    (2.8704, 0.400000)
    (2.9300, 0.400000)
    (2.9763, 0.800000)
    (3.0391, 0.300000)
};

\addplot[
    color={rgb,255:red,174; green,78; blue,255}, 
    fill opacity=0.75,
    mark=*,
    mark size=0.8pt,
    only marks,
    forget plot
]
coordinates {
    (1.2829, 0.052673)
    (2.5647, 0.684881)
    (3.8428, 0.054217)
    (4.9835, 0.000000)
    (6.1124, 0.000000)
    (7.2510, 0.000000)
    (8.3707, 0.504808)
    (9.4899, 0.000000)
    (10.8251, 0.871983)
    (10.8251, 0.000000)
    (11.9817, 0.259259)
    (13.2643, 0.926418)
    (13.2643, 0.000000)
    (14.3585, 0.000000)
    (15.7082, 0.926373)
    (16.8380, 0.000000)
    (17.9108, 0.000000)
    (19.1126, 0.000000)
    (19.1126, 0.000000)
    (20.2948, 0.723806)
    (21.6536, 0.818659)
    (22.8064, 0.250000)
    (24.0455, 0.000000)
    (25.3197, 0.848559)
    (26.4818, 0.000000)
    (27.7607, 0.784158)
    (28.9056, 0.000000)
    (30.2311, 0.864807)
    (31.4635, 0.638364)
    (32.7494, 0.566020)
    (33.8600, 0.708736)
    (35.1195, 0.823843)
    (36.4125, 0.820100)
    (37.5323, 0.000000)
    (38.7537, 0.772780)
    (39.8844, 0.167572)
    (41.2044, 0.827705)
    (42.5451, 0.857932)
    (43.8838, 0.769378)
    (44.7272, 0.828866)
    (45.8305, 0.000000)
    (46.9216, 0.576923)
};

\addplot[
    color={rgb,255:red,174; green,78; blue,255}, 
    fill opacity=0.75,
    mark=*,
    mark size=0.8pt,
    only marks,
    forget plot
]
coordinates {
    (1.0544, 0.000000)
    (1.0544, 0.000000)
    (2.2790, 0.146586)
    (3.4932, 0.000000)
    (3.4932, 0.000000)
    (4.8934, 0.812342)
    (6.0027, 0.000000)
    (6.0027, 0.000000)
    (7.2538, 0.399863)
    (8.3472, 0.000000)
    (9.6568, 0.736229)
    (10.9183, 0.191508)
    (12.0056, 0.000000)
    (13.0621, 0.000000)
    (14.4176, 0.724005)
    (15.9573, 0.602468)
    (17.3225, 0.650297)
    (18.7595, 0.340128)
    (19.9368, 0.000000)
    (21.3220, 0.844384)
    (22.6217, 0.373206)
    (23.9708, 0.923657)
    (25.3265, 0.576996)
    (26.4812, 0.000000)
    (27.8808, 0.825562)
    (29.2403, 0.902264)
    (30.6161, 0.910706)
    (31.4761, 0.904561)
    (32.0037, 0.876565)
    (32.5172, 0.048851)
    (32.6888, 0.006026)
};

\addplot[
    color={rgb,255:red,174; green,78; blue,255}, 
    fill opacity=0.75,
    mark=*,
    mark size=0.8pt,
    only marks,
    forget plot
]
coordinates {
    (1.2506, 0.581650)
    (1.2506, 0.000000)
    (1.2506, 0.000000)
    (2.2835, 0.000000)
    (3.4156, 0.064167)
    (4.7291, 0.516988)
    (5.9278, 0.097518)
    (7.1561, 0.225948)
    (8.3699, 0.103825)
    (9.5187, 0.187179)
    (10.5741, 0.108696)
    (11.7619, 0.211847)
    (12.8716, 0.000000)
    (14.2092, 0.747662)
    (15.4459, 0.487099)
    (16.6366, 0.247500)
    (17.7277, 0.000000)
    (19.0436, 0.544194)
    (20.3583, 0.681626)
    (21.6262, 0.488960)
    (22.9652, 0.748239)
    (24.2857, 0.860136)
    (25.3437, 0.000000)
    (26.6058, 0.694426)
    (27.6854, 0.000000)
    (28.9324, 0.809847)
    (30.0404, 0.316520)
    (31.3515, 0.746584)
    (32.5907, 0.670489)
    (33.9116, 0.773226)
    (35.1664, 0.245088)
    (36.5259, 0.691618)
    (37.8991, 0.864454)
    (39.2545, 0.768782)
    (40.4919, 0.618661)
    (41.8200, 0.490231)
    (43.1808, 0.804168)
    (44.4596, 0.498205)
    (45.8233, 0.715629)
    (47.1401, 0.611659)
    (48.5128, 0.731203)
    (49.8598, 0.758262)
    (51.1852, 0.784004)
    (52.4860, 0.792809)
    (53.8439, 0.911770)
    (55.1763, 0.505130)
    (56.4292, 0.736533)
    (57.7536, 0.694925)
    (59.0040, 0.858275)
};

\addplot[
    color={rgb,255:red,174; green,78; blue,255}, 
    fill opacity=0.75,
    mark=*,
    mark size=0.8pt,
    only marks,
    forget plot
]
coordinates {
    (1.1606, 0.000000)
    (1.1606, 0.000000)
    (2.4413, 0.000000)
    (2.4413, 0.000000)
    (3.5687, 0.000000)
    (4.6320, 0.000000)
    (5.7850, 0.243421)
    (6.7897, 0.000000)
    (8.0389, 0.000000)
    (9.2842, 0.340560)
    (10.4479, 0.000000)
    (11.7560, 0.398222)
    (12.9039, 0.032967)
    (14.2482, 0.265670)
    (15.4444, 0.234731)
    (16.3832, 0.774278)
    (17.2898, 0.679559)
    (17.8775, 0.000000)
    (19.4412, 0.858983)
    (20.7493, 0.515213)
    (20.7493, 0.000000)
    (22.1072, 0.018182)
    (23.3888, 0.403756)
    (24.6976, 0.487935)
    (25.9324, 0.162393)
    (27.3796, 0.228448)
    (28.9418, 0.186563)
    (30.2538, 0.322980)
    (31.6541, 0.849251)
    (33.0365, 0.000000)
    (34.6450, 0.909415)
    (35.8360, 0.000000)
    (37.2273, 0.802022)
    (38.8895, 0.873150)
    (40.4982, 0.791502)
    (41.8832, 0.836991)
    (43.2003, 0.382304)
    (44.8559, 0.583966)
    (46.5455, 0.859406)
    (47.9125, 0.865096)
    (49.2958, 0.606957)
    (50.9163, 0.796409)
    (52.2308, 0.754559)
    (53.9740, 0.891498)
    (55.3619, 0.916645)
    (55.3619, 0.000000)
    (56.7419, 0.848860)
    (58.2723, 0.473731)
    (59.4352, 0.697297)
    (60.7178, 0.805069)
};

\addplot[
    color={rgb,255:red,174; green,78; blue,255}, 
    fill opacity=0.75,
    mark=*,
    mark size=0.8pt,
    only marks,
    forget plot
]
coordinates {
    (1.0675, 0.083333)
    (1.0675, 0.000000)
    (2.4160, 0.690036)
    (2.4160, 0.000000)
    (3.6674, 0.000000)
    (4.9796, 0.639452)
    (6.1507, 0.133929)
    (7.3131, 0.514861)
    (8.4607, 0.043907)
    (9.6932, 0.510067)
    (10.9829, 0.678789)
    (12.3055, 0.581452)
    (13.5032, 0.094203)
    (14.7962, 0.161812)
    (16.1333, 0.865833)
    (17.3937, 0.763928)
    (18.7219, 0.668237)
    (20.1097, 0.865872)
    (21.4574, 0.919476)
    (22.8148, 0.893082)
    (24.2063, 0.859550)
    (25.6034, 0.895962)
    (26.8770, 0.741228)
    (28.0589, 0.500305)
    (29.3231, 0.562235)
    (30.6154, 0.492817)
    (31.9117, 0.701583)
    (33.1491, 0.472600)
    (34.5127, 0.782952)
    (35.8093, 0.789239)
    (37.1718, 0.955739)
    (38.5469, 0.956389)
    (39.8470, 0.849765)
    (41.2117, 0.758847)
    (42.6161, 0.805640)
    (44.0020, 0.847313)
    (45.2798, 0.612086)
    (46.6988, 0.960670)
    (48.0340, 0.798815)
    (49.4346, 0.867433)
    (50.8273, 0.921017)
    (52.1411, 0.886875)
    (53.2321, 0.000000)
    (54.5598, 0.879292)
    (55.7725, 0.000000)
    (57.1890, 0.909622)
    (58.3712, 0.728312)
    (59.4772, 0.803196)
    (60.3999, 0.804597)
};

\addplot[
    name path=sampling_upper,
    draw=none,
    forget plot
] coordinates {
    (0.0168, 0.000000)
    (0.0337, 0.000000)
    (0.0505, 0.000000)
    (0.0674, 0.240000)
    (0.0842, 0.254919)
    (0.1010, 0.254919)
    (0.1179, 0.254919)
    (0.1347, 0.254919)
    (0.1516, 0.493238)
    (0.1684, 0.493238)
    (0.1852, 0.493238)
    (0.2021, 0.822490)
    (0.2189, 0.822490)
    (0.2357, 0.822490)
    (0.2526, 0.892410)
    (0.2694, 0.940666)
    (0.2863, 0.960666)
    (0.3031, 1.000000)
    (0.3199, 1.000000)
    (0.3368, 1.000000)
    (0.3536, 1.000000)
    (0.3705, 1.000000)
    (0.3873, 1.000000)
    (0.4041, 1.000000)
    (0.4210, 1.000000)
    (0.4378, 1.000000)
    (0.4547, 1.000000)
    (0.4715, 1.000000)
    (0.4883, 1.000000)
    (0.5052, 0.941421)
    (0.5220, 0.941421)
    (0.5389, 0.941421)
    (0.5557, 0.941421)
    (0.5725, 0.941421)
    (0.5894, 0.941421)
    (0.6062, 0.941421)
    (0.6231, 0.941421)
    (0.6399, 0.936619)
    (0.6567, 0.936619)
    (0.6736, 0.936619)
    (0.6904, 0.936619)
    (0.7072, 0.936619)
    (0.7241, 0.936619)
    (0.7409, 0.936619)
    (0.7578, 0.936619)
    (0.7746, 0.936619)
    (0.7914, 0.936619)
    (0.8083, 0.936619)
    (0.8251, 0.936619)
    (0.8420, 0.936619)
    (0.8588, 0.936619)
    (0.8756, 0.936619)
    (0.8925, 0.936619)
    (0.9093, 0.936619)
    (0.9262, 0.936619)
    (0.9430, 0.936619)
    (0.9598, 0.936619)
    (0.9767, 0.936619)
    (0.9935, 0.936619)
    (1.0104, 0.936619)
    (1.0272, 0.936619)
    (1.0440, 0.936619)
    (1.0609, 0.936619)
    (1.0777, 0.936619)
    (1.0946, 0.936619)
    (1.1114, 0.936619)
    (1.1282, 0.941980)
    (1.1451, 0.941980)
    (1.1619, 0.941980)
    (1.1787, 0.941980)
    (1.1956, 0.941980)
    (1.2124, 0.941980)
    (1.2293, 0.941980)
    (1.2461, 0.941980)
    (1.2629, 0.941980)
    (1.2798, 0.941980)
    (1.2966, 0.961980)
    (1.3135, 0.961980)
    (1.3303, 0.961980)
    (1.3471, 0.961980)
    (1.3640, 0.961980)
    (1.3808, 0.961980)
    (1.3977, 0.961980)
    (1.4145, 0.961980)
    (1.4313, 0.961980)
    (1.4482, 0.961980)
    (1.4650, 0.961980)
    (1.4819, 0.961980)
    (1.4987, 0.961980)
    (1.5155, 0.961980)
    (1.5324, 0.961980)
    (1.5492, 0.961980)
    (1.5660, 0.961980)
    (1.5829, 0.961980)
    (1.5997, 0.961980)
    (1.6166, 0.961980)
    (1.6334, 0.961980)
    (1.6502, 0.961980)
    (1.6671, 0.961980)
    (1.6839, 0.961980)
    (1.7008, 0.961980)
    (1.7176, 0.961980)
    (1.7344, 0.961980)
    (1.7513, 0.961980)
    (1.7681, 0.961980)
    (1.7850, 0.961980)
    (1.8018, 0.961980)
    (1.8186, 0.961980)
    (1.8355, 0.961980)
    (1.8523, 0.961980)
    (1.8692, 0.961980)
    (1.8860, 0.961980)
    (1.9028, 0.961980)
    (1.9197, 0.961980)
    (1.9365, 0.954833)
    (1.9534, 0.954833)
    (1.9702, 0.954833)
    (1.9870, 0.954833)
    (2.0039, 0.954833)
    (2.0207, 0.954833)
    (2.0375, 0.954833)
    (2.0544, 0.963246)
    (2.0712, 0.963246)
    (2.0881, 0.963246)
    (2.1049, 0.963246)
    (2.1217, 0.963246)
    (2.1386, 0.963246)
    (2.1554, 0.963246)
    (2.1723, 0.963246)
    (2.1891, 0.963246)
    (2.2059, 0.963246)
    (2.2228, 0.963246)
    (2.2396, 0.963246)
    (2.2565, 0.963246)
    (2.2733, 0.963246)
    (2.2901, 0.963246)
    (2.3070, 0.963246)
    (2.3238, 0.963246)
    (2.3407, 0.963246)
    (2.3575, 0.963246)
    (2.3743, 0.963246)
    (2.3912, 0.963246)
    (2.4080, 0.963246)
    (2.4248, 0.963246)
    (2.4417, 0.963246)
    (2.4585, 0.963246)
    (2.4754, 0.963246)
    (2.4922, 0.963246)
    (2.5090, 0.963246)
    (2.5259, 0.963246)
    (2.5427, 0.963246)
    (2.5596, 0.963246)
    (2.5764, 0.963246)
    (2.5932, 0.963246)
    (2.6101, 0.963246)
    (2.6269, 0.963246)
    (2.6438, 0.963246)
    (2.6606, 0.963246)
    (2.6774, 0.960000)
    (2.6943, 0.960000)
    (2.7111, 0.960000)
    (2.7280, 0.960000)
    (2.7448, 0.960000)
    (2.7616, 0.960000)
    (2.7785, 0.960000)
    (2.7953, 0.960000)
    (2.8122, 0.960000)
    (2.8290, 0.960000)
    (2.8458, 0.960000)
    (2.8627, 0.960000)
    (2.8795, 0.960000)
    (2.8963, 0.960000)
    (2.9132, 0.960000)
    (2.9300, 0.960000)
    (2.9469, 0.960000)
    (2.9637, 0.960000)
    (2.9805, 0.960000)
    (2.9974, 0.960000)
    (3.0142, 0.960000)
    (3.0311, 0.960000)
    (3.0479, 0.960000)
    (3.0647, 0.960000)
    (3.0816, 0.960000)
    (3.0984, 0.960000)
    (3.1153, 0.960000)
    (3.1321, 0.960000)
    (3.1489, 0.960000)
    (3.1658, 0.960000)
    (3.1826, 0.960000)
    (3.1995, 0.960000)
    (3.2163, 0.960000)
    (3.2331, 0.960000)
    (3.2500, 0.960000)
    (3.2668, 0.960000)
    (3.2837, 0.960000)
    (3.3005, 0.960000)
    (3.3173, 0.960000)
    (3.3342, 0.960000)
    (3.3510, 0.960000)
};

\addplot[
    name path=sampling_lower,
    draw=none,
    forget plot
] coordinates {
    (0.0168, 0.000000)
    (0.0337, 0.000000)
    (0.0505, 0.000000)
    (0.0674, 0.000000)
    (0.0842, 0.000000)
    (0.1010, 0.000000)
    (0.1179, 0.000000)
    (0.1347, 0.000000)
    (0.1516, 0.026762)
    (0.1684, 0.026762)
    (0.1852, 0.026762)
    (0.2021, 0.177510)
    (0.2189, 0.177510)
    (0.2357, 0.177510)
    (0.2526, 0.267590)
    (0.2694, 0.339334)
    (0.2863, 0.359334)
    (0.3031, 0.383772)
    (0.3199, 0.383772)
    (0.3368, 0.383772)
    (0.3536, 0.383772)
    (0.3705, 0.383772)
    (0.3873, 0.383772)
    (0.4041, 0.383772)
    (0.4210, 0.383772)
    (0.4378, 0.383772)
    (0.4547, 0.383772)
    (0.4715, 0.383772)
    (0.4883, 0.383772)
    (0.5052, 0.658579)
    (0.5220, 0.658579)
    (0.5389, 0.658579)
    (0.5557, 0.658579)
    (0.5725, 0.658579)
    (0.5894, 0.658579)
    (0.6062, 0.658579)
    (0.6231, 0.658579)
    (0.6399, 0.703381)
    (0.6567, 0.703381)
    (0.6736, 0.703381)
    (0.6904, 0.703381)
    (0.7072, 0.703381)
    (0.7241, 0.703381)
    (0.7409, 0.703381)
    (0.7578, 0.703381)
    (0.7746, 0.703381)
    (0.7914, 0.703381)
    (0.8083, 0.703381)
    (0.8251, 0.703381)
    (0.8420, 0.703381)
    (0.8588, 0.703381)
    (0.8756, 0.703381)
    (0.8925, 0.703381)
    (0.9093, 0.703381)
    (0.9262, 0.703381)
    (0.9430, 0.703381)
    (0.9598, 0.703381)
    (0.9767, 0.703381)
    (0.9935, 0.703381)
    (1.0104, 0.703381)
    (1.0272, 0.703381)
    (1.0440, 0.703381)
    (1.0609, 0.703381)
    (1.0777, 0.703381)
    (1.0946, 0.703381)
    (1.1114, 0.703381)
    (1.1282, 0.738020)
    (1.1451, 0.738020)
    (1.1619, 0.738020)
    (1.1787, 0.738020)
    (1.1956, 0.738020)
    (1.2124, 0.738020)
    (1.2293, 0.738020)
    (1.2461, 0.738020)
    (1.2629, 0.738020)
    (1.2798, 0.738020)
    (1.2966, 0.758020)
    (1.3135, 0.758020)
    (1.3303, 0.758020)
    (1.3471, 0.758020)
    (1.3640, 0.758020)
    (1.3808, 0.758020)
    (1.3977, 0.758020)
    (1.4145, 0.758020)
    (1.4313, 0.758020)
    (1.4482, 0.758020)
    (1.4650, 0.758020)
    (1.4819, 0.758020)
    (1.4987, 0.758020)
    (1.5155, 0.758020)
    (1.5324, 0.758020)
    (1.5492, 0.758020)
    (1.5660, 0.758020)
    (1.5829, 0.758020)
    (1.5997, 0.758020)
    (1.6166, 0.758020)
    (1.6334, 0.758020)
    (1.6502, 0.758020)
    (1.6671, 0.758020)
    (1.6839, 0.758020)
    (1.7008, 0.758020)
    (1.7176, 0.758020)
    (1.7344, 0.758020)
    (1.7513, 0.758020)
    (1.7681, 0.758020)
    (1.7850, 0.758020)
    (1.8018, 0.758020)
    (1.8186, 0.758020)
    (1.8355, 0.758020)
    (1.8523, 0.758020)
    (1.8692, 0.758020)
    (1.8860, 0.758020)
    (1.9028, 0.758020)
    (1.9197, 0.758020)
    (1.9365, 0.805167)
    (1.9534, 0.805167)
    (1.9702, 0.805167)
    (1.9870, 0.805167)
    (2.0039, 0.805167)
    (2.0207, 0.805167)
    (2.0375, 0.805167)
    (2.0544, 0.836754)
    (2.0712, 0.836754)
    (2.0881, 0.836754)
    (2.1049, 0.836754)
    (2.1217, 0.836754)
    (2.1386, 0.836754)
    (2.1554, 0.836754)
    (2.1723, 0.836754)
    (2.1891, 0.836754)
    (2.2059, 0.836754)
    (2.2228, 0.836754)
    (2.2396, 0.836754)
    (2.2565, 0.836754)
    (2.2733, 0.836754)
    (2.2901, 0.836754)
    (2.3070, 0.836754)
    (2.3238, 0.836754)
    (2.3407, 0.836754)
    (2.3575, 0.836754)
    (2.3743, 0.836754)
    (2.3912, 0.836754)
    (2.4080, 0.836754)
    (2.4248, 0.836754)
    (2.4417, 0.836754)
    (2.4585, 0.836754)
    (2.4754, 0.836754)
    (2.4922, 0.836754)
    (2.5090, 0.836754)
    (2.5259, 0.836754)
    (2.5427, 0.836754)
    (2.5596, 0.836754)
    (2.5764, 0.836754)
    (2.5932, 0.836754)
    (2.6101, 0.836754)
    (2.6269, 0.836754)
    (2.6438, 0.836754)
    (2.6606, 0.836754)
    (2.6774, 0.880000)
    (2.6943, 0.880000)
    (2.7111, 0.880000)
    (2.7280, 0.880000)
    (2.7448, 0.880000)
    (2.7616, 0.880000)
    (2.7785, 0.880000)
    (2.7953, 0.880000)
    (2.8122, 0.880000)
    (2.8290, 0.880000)
    (2.8458, 0.880000)
    (2.8627, 0.880000)
    (2.8795, 0.880000)
    (2.8963, 0.880000)
    (2.9132, 0.880000)
    (2.9300, 0.880000)
    (2.9469, 0.880000)
    (2.9637, 0.880000)
    (2.9805, 0.880000)
    (2.9974, 0.880000)
    (3.0142, 0.880000)
    (3.0311, 0.880000)
    (3.0479, 0.880000)
    (3.0647, 0.880000)
    (3.0816, 0.880000)
    (3.0984, 0.880000)
    (3.1153, 0.880000)
    (3.1321, 0.880000)
    (3.1489, 0.880000)
    (3.1658, 0.880000)
    (3.1826, 0.880000)
    (3.1995, 0.880000)
    (3.2163, 0.880000)
    (3.2331, 0.880000)
    (3.2500, 0.880000)
    (3.2668, 0.880000)
    (3.2837, 0.880000)
    (3.3005, 0.880000)
    (3.3173, 0.880000)
    (3.3342, 0.880000)
    (3.3510, 0.880000)
};

\addplot[
    fill={rgb,255:red,251; green,226; blue,78}, 
    fill opacity=0.4,
    forget plot
] fill between[of=sampling_upper and sampling_lower];

\addplot[
    color={rgb,255:red,244; green,176; blue,34}, 
    draw opacity=0.75,
    line width=3pt,
    mark=none,
] coordinates {
    (0.0168, 0.000000)
    (0.0337, 0.000000)
    (0.0505, 0.000000)
    (0.0674, 0.080000)
    (0.0842, 0.100000)
    (0.1010, 0.100000)
    (0.1179, 0.100000)
    (0.1347, 0.100000)
    (0.1516, 0.260000)
    (0.1684, 0.260000)
    (0.1852, 0.260000)
    (0.2021, 0.500000)
    (0.2189, 0.500000)
    (0.2357, 0.500000)
    (0.2526, 0.580000)
    (0.2694, 0.640000)
    (0.2863, 0.660000)
    (0.3031, 0.700000)
    (0.3199, 0.700000)
    (0.3368, 0.700000)
    (0.3536, 0.700000)
    (0.3705, 0.700000)
    (0.3873, 0.700000)
    (0.4041, 0.700000)
    (0.4210, 0.700000)
    (0.4378, 0.700000)
    (0.4547, 0.700000)
    (0.4715, 0.700000)
    (0.4883, 0.700000)
    (0.5052, 0.800000)
    (0.5220, 0.800000)
    (0.5389, 0.800000)
    (0.5557, 0.800000)
    (0.5725, 0.800000)
    (0.5894, 0.800000)
    (0.6062, 0.800000)
    (0.6231, 0.800000)
    (0.6399, 0.820000)
    (0.6567, 0.820000)
    (0.6736, 0.820000)
    (0.6904, 0.820000)
    (0.7072, 0.820000)
    (0.7241, 0.820000)
    (0.7409, 0.820000)
    (0.7578, 0.820000)
    (0.7746, 0.820000)
    (0.7914, 0.820000)
    (0.8083, 0.820000)
    (0.8251, 0.820000)
    (0.8420, 0.820000)
    (0.8588, 0.820000)
    (0.8756, 0.820000)
    (0.8925, 0.820000)
    (0.9093, 0.820000)
    (0.9262, 0.820000)
    (0.9430, 0.820000)
    (0.9598, 0.820000)
    (0.9767, 0.820000)
    (0.9935, 0.820000)
    (1.0104, 0.820000)
    (1.0272, 0.820000)
    (1.0440, 0.820000)
    (1.0609, 0.820000)
    (1.0777, 0.820000)
    (1.0946, 0.820000)
    (1.1114, 0.820000)
    (1.1282, 0.840000)
    (1.1451, 0.840000)
    (1.1619, 0.840000)
    (1.1787, 0.840000)
    (1.1956, 0.840000)
    (1.2124, 0.840000)
    (1.2293, 0.840000)
    (1.2461, 0.840000)
    (1.2629, 0.840000)
    (1.2798, 0.840000)
    (1.2966, 0.860000)
    (1.3135, 0.860000)
    (1.3303, 0.860000)
    (1.3471, 0.860000)
    (1.3640, 0.860000)
    (1.3808, 0.860000)
    (1.3977, 0.860000)
    (1.4145, 0.860000)
    (1.4313, 0.860000)
    (1.4482, 0.860000)
    (1.4650, 0.860000)
    (1.4819, 0.860000)
    (1.4987, 0.860000)
    (1.5155, 0.860000)
    (1.5324, 0.860000)
    (1.5492, 0.860000)
    (1.5660, 0.860000)
    (1.5829, 0.860000)
    (1.5997, 0.860000)
    (1.6166, 0.860000)
    (1.6334, 0.860000)
    (1.6502, 0.860000)
    (1.6671, 0.860000)
    (1.6839, 0.860000)
    (1.7008, 0.860000)
    (1.7176, 0.860000)
    (1.7344, 0.860000)
    (1.7513, 0.860000)
    (1.7681, 0.860000)
    (1.7850, 0.860000)
    (1.8018, 0.860000)
    (1.8186, 0.860000)
    (1.8355, 0.860000)
    (1.8523, 0.860000)
    (1.8692, 0.860000)
    (1.8860, 0.860000)
    (1.9028, 0.860000)
    (1.9197, 0.860000)
    (1.9365, 0.880000)
    (1.9534, 0.880000)
    (1.9702, 0.880000)
    (1.9870, 0.880000)
    (2.0039, 0.880000)
    (2.0207, 0.880000)
    (2.0375, 0.880000)
    (2.0544, 0.900000)
    (2.0712, 0.900000)
    (2.0881, 0.900000)
    (2.1049, 0.900000)
    (2.1217, 0.900000)
    (2.1386, 0.900000)
    (2.1554, 0.900000)
    (2.1723, 0.900000)
    (2.1891, 0.900000)
    (2.2059, 0.900000)
    (2.2228, 0.900000)
    (2.2396, 0.900000)
    (2.2565, 0.900000)
    (2.2733, 0.900000)
    (2.2901, 0.900000)
    (2.3070, 0.900000)
    (2.3238, 0.900000)
    (2.3407, 0.900000)
    (2.3575, 0.900000)
    (2.3743, 0.900000)
    (2.3912, 0.900000)
    (2.4080, 0.900000)
    (2.4248, 0.900000)
    (2.4417, 0.900000)
    (2.4585, 0.900000)
    (2.4754, 0.900000)
    (2.4922, 0.900000)
    (2.5090, 0.900000)
    (2.5259, 0.900000)
    (2.5427, 0.900000)
    (2.5596, 0.900000)
    (2.5764, 0.900000)
    (2.5932, 0.900000)
    (2.6101, 0.900000)
    (2.6269, 0.900000)
    (2.6438, 0.900000)
    (2.6606, 0.900000)
    (2.6774, 0.920000)
    (2.6943, 0.920000)
    (2.7111, 0.920000)
    (2.7280, 0.920000)
    (2.7448, 0.920000)
    (2.7616, 0.920000)
    (2.7785, 0.920000)
    (2.7953, 0.920000)
    (2.8122, 0.920000)
    (2.8290, 0.920000)
    (2.8458, 0.920000)
    (2.8627, 0.920000)
    (2.8795, 0.920000)
    (2.8963, 0.920000)
    (2.9132, 0.920000)
    (2.9300, 0.920000)
    (2.9469, 0.920000)
    (2.9637, 0.920000)
    (2.9805, 0.920000)
    (2.9974, 0.920000)
    (3.0142, 0.920000)
    (3.0311, 0.920000)
    (3.0479, 0.920000)
    (3.0647, 0.920000)
    (3.0816, 0.920000)
    (3.0984, 0.920000)
    (3.1153, 0.920000)
    (3.1321, 0.920000)
    (3.1489, 0.920000)
    (3.1658, 0.920000)
    (3.1826, 0.920000)
    (3.1995, 0.920000)
    (3.2163, 0.920000)
    (3.2331, 0.920000)
    (3.2500, 0.920000)
    (3.2668, 0.920000)
    (3.2837, 0.920000)
    (3.3005, 0.920000)
    (3.3173, 0.920000)
    (3.3342, 0.920000)
    (3.3510, 0.920000)
};
\addlegendentry{Sumo (ours)}

\addplot[
    name path=rl_upper,
    draw=none,
    forget plot
] coordinates {
    (0.3051, 0.000000)
    (0.6102, 0.000000)
    (0.9153, 0.000000)
    (1.2205, 0.050000)
    (1.5256, 0.364904)
    (1.8307, 0.364904)
    (2.1358, 0.364904)
    (2.4409, 0.579142)
    (2.7460, 0.710584)
    (3.0511, 0.710584)
    (3.3563, 0.710584)
    (3.6614, 0.710584)
    (3.9665, 0.710584)
    (4.2716, 0.710584)
    (4.5767, 0.710584)
    (4.8818, 0.710584)
    (5.1869, 0.840158)
    (5.4921, 0.840158)
    (5.7972, 0.796298)
    (6.1023, 0.796298)
    (6.4074, 0.796298)
    (6.7125, 0.796298)
    (7.0176, 0.796298)
    (7.3227, 0.796298)
    (7.6279, 0.796298)
    (7.9330, 0.796298)
    (8.2381, 0.796298)
    (8.5432, 0.796298)
    (8.8483, 0.796298)
    (9.1534, 0.796298)
    (9.4585, 0.780418)
    (9.7637, 0.780418)
    (10.0688, 0.780418)
    (10.3739, 0.780418)
    (10.6790, 0.780418)
    (10.9841, 0.847583)
    (11.2892, 0.847583)
    (11.5943, 0.847583)
    (11.8995, 0.840036)
    (12.2046, 0.840036)
    (12.5097, 0.840036)
    (12.8148, 0.840036)
    (13.1199, 0.840036)
    (13.4250, 0.864705)
    (13.7301, 0.864705)
    (14.0353, 0.864705)
    (14.3404, 0.891691)
    (14.6455, 0.891691)
    (14.9506, 0.891691)
    (15.2557, 0.891691)
    (15.5608, 0.891691)
    (15.8659, 0.891691)
    (16.1711, 0.935665)
    (16.4762, 0.889598)
    (16.7813, 0.889598)
    (17.0864, 0.889598)
    (17.3915, 0.889598)
    (17.6966, 0.889598)
    (18.0017, 0.889598)
    (18.3069, 0.889598)
    (18.6120, 0.889598)
    (18.9171, 0.889598)
    (19.2222, 0.889598)
    (19.5273, 0.901849)
    (19.8324, 0.901849)
    (20.1376, 0.901860)
    (20.4427, 0.901860)
    (20.7478, 0.901860)
    (21.0529, 0.901860)
    (21.3580, 0.906401)
    (21.6631, 0.923900)
    (21.9682, 0.923900)
    (22.2734, 0.923900)
    (22.5785, 0.923900)
    (22.8836, 0.923900)
    (23.1887, 0.923816)
    (23.4938, 0.923816)
    (23.7989, 0.923816)
    (24.1040, 0.943639)
    (24.4092, 0.928984)
    (24.7143, 0.928984)
    (25.0194, 0.928984)
    (25.3245, 0.928984)
    (25.6296, 0.928984)
    (25.9347, 0.928984)
    (26.2398, 0.928984)
    (26.5450, 0.928984)
    (26.8501, 0.928984)
    (27.1552, 0.928984)
    (27.4603, 0.928984)
    (27.7654, 0.928984)
    (28.0705, 0.928984)
    (28.3756, 0.928984)
    (28.6808, 0.928984)
    (28.9859, 0.928984)
    (29.2910, 0.928984)
    (29.5961, 0.928984)
    (29.9012, 0.928984)
    (30.2063, 0.928984)
    (30.5114, 0.928984)
    (30.8166, 0.928984)
    (31.1217, 0.928984)
    (31.4268, 0.928984)
    (31.7319, 0.928984)
    (32.0370, 0.928984)
    (32.3421, 0.928984)
    (32.6472, 0.928984)
    (32.9524, 0.928984)
    (33.2575, 0.928984)
    (33.5626, 0.928984)
    (33.8677, 0.928984)
    (34.1728, 0.928984)
    (34.4779, 0.928984)
    (34.7830, 0.932351)
    (35.0882, 0.932351)
    (35.3933, 0.932351)
    (35.6984, 0.932351)
    (36.0035, 0.932351)
    (36.3086, 0.932351)
    (36.6137, 0.932351)
    (36.9188, 0.932351)
    (37.2240, 0.946398)
    (37.5291, 0.946398)
    (37.8342, 0.946398)
    (38.1393, 0.945759)
    (38.4444, 0.945759)
    (38.7495, 0.946063)
    (39.0546, 0.946063)
    (39.3598, 0.946063)
    (39.6649, 0.946063)
    (39.9700, 0.946063)
    (40.2751, 0.946063)
    (40.5802, 0.946063)
    (40.8853, 0.946063)
    (41.1904, 0.946063)
    (41.4956, 0.946063)
    (41.8007, 0.946063)
    (42.1058, 0.946063)
    (42.4109, 0.946063)
    (42.7160, 0.946063)
    (43.0211, 0.946063)
    (43.3262, 0.946063)
    (43.6314, 0.946063)
    (43.9365, 0.946063)
    (44.2416, 0.946063)
    (44.5467, 0.946063)
    (44.8518, 0.946063)
    (45.1569, 0.946063)
    (45.4620, 0.946063)
    (45.7672, 0.946063)
    (46.0723, 0.946063)
    (46.3774, 0.946063)
    (46.6825, 0.946063)
    (46.9876, 0.948095)
    (47.2927, 0.948095)
    (47.5978, 0.948095)
    (47.9030, 0.948095)
    (48.2081, 0.948095)
    (48.5132, 0.948095)
    (48.8183, 0.948095)
    (49.1234, 0.948095)
    (49.4285, 0.948095)
    (49.7336, 0.948095)
    (50.0388, 0.948095)
    (50.3439, 0.948095)
    (50.6490, 0.948095)
    (50.9541, 0.948095)
    (51.2592, 0.948095)
    (51.5643, 0.948095)
    (51.8694, 0.948095)
    (52.1746, 0.948095)
    (52.4797, 0.948095)
    (52.7848, 0.948095)
    (53.0899, 0.948095)
    (53.3950, 0.948095)
    (53.7001, 0.948095)
    (54.0052, 0.944741)
    (54.3104, 0.944741)
    (54.6155, 0.944741)
    (54.9206, 0.944741)
    (55.2257, 0.944741)
    (55.5308, 0.945042)
    (55.8359, 0.945042)
    (56.1410, 0.945042)
    (56.4462, 0.945042)
    (56.7513, 0.945042)
    (57.0564, 0.945042)
    (57.3615, 0.945042)
    (57.6666, 0.945042)
    (57.9717, 0.945042)
    (58.2769, 0.945042)
    (58.5820, 0.945042)
    (58.8871, 0.945042)
    (59.1922, 0.945042)
    (59.4973, 0.945042)
    (59.8024, 0.945042)
    (60.1075, 0.945042)
    (60.4127, 0.945042)
    (60.7178, 0.945042)
};

\addplot[
    name path=rl_lower,
    draw=none,
    forget plot
] coordinates {
    (0.3051, 0.000000)
    (0.6102, 0.000000)
    (0.9153, 0.000000)
    (1.2205, 0.000000)
    (1.5256, 0.000000)
    (1.8307, 0.000000)
    (2.1358, 0.000000)
    (2.4409, 0.009236)
    (2.7460, 0.130677)
    (3.0511, 0.130677)
    (3.3563, 0.130677)
    (3.6614, 0.130677)
    (3.9665, 0.130677)
    (4.2716, 0.130677)
    (4.5767, 0.130677)
    (4.8818, 0.130677)
    (5.1869, 0.267405)
    (5.4921, 0.267405)
    (5.7972, 0.408633)
    (6.1023, 0.408633)
    (6.4074, 0.408633)
    (6.7125, 0.408633)
    (7.0176, 0.408633)
    (7.3227, 0.408633)
    (7.6279, 0.408633)
    (7.9330, 0.408633)
    (8.2381, 0.408633)
    (8.5432, 0.408633)
    (8.8483, 0.408633)
    (9.1534, 0.408633)
    (9.4585, 0.463370)
    (9.7637, 0.463370)
    (10.0688, 0.463370)
    (10.3739, 0.463370)
    (10.6790, 0.463370)
    (10.9841, 0.471046)
    (11.2892, 0.471046)
    (11.5943, 0.471046)
    (11.8995, 0.501657)
    (12.2046, 0.501657)
    (12.5097, 0.501657)
    (12.8148, 0.501657)
    (13.1199, 0.501657)
    (13.4250, 0.498762)
    (13.7301, 0.498762)
    (14.0353, 0.498762)
    (14.3404, 0.538181)
    (14.6455, 0.538181)
    (14.9506, 0.538181)
    (15.2557, 0.538181)
    (15.5608, 0.538181)
    (15.8659, 0.538181)
    (16.1711, 0.564526)
    (16.4762, 0.761016)
    (16.7813, 0.761016)
    (17.0864, 0.761016)
    (17.3915, 0.761016)
    (17.6966, 0.761016)
    (18.0017, 0.761016)
    (18.3069, 0.761016)
    (18.6120, 0.761016)
    (18.9171, 0.761016)
    (19.2222, 0.761016)
    (19.5273, 0.782646)
    (19.8324, 0.782646)
    (20.1376, 0.782651)
    (20.4427, 0.782651)
    (20.7478, 0.782651)
    (21.0529, 0.782651)
    (21.3580, 0.790926)
    (21.6631, 0.794869)
    (21.9682, 0.794869)
    (22.2734, 0.794869)
    (22.5785, 0.794869)
    (22.8836, 0.794869)
    (23.1887, 0.795184)
    (23.4938, 0.795184)
    (23.7989, 0.795184)
    (24.1040, 0.807071)
    (24.4092, 0.866484)
    (24.7143, 0.866484)
    (25.0194, 0.866484)
    (25.3245, 0.866484)
    (25.6296, 0.866484)
    (25.9347, 0.866484)
    (26.2398, 0.866484)
    (26.5450, 0.866484)
    (26.8501, 0.866484)
    (27.1552, 0.866484)
    (27.4603, 0.866484)
    (27.7654, 0.866484)
    (28.0705, 0.866484)
    (28.3756, 0.866484)
    (28.6808, 0.866484)
    (28.9859, 0.866484)
    (29.2910, 0.866484)
    (29.5961, 0.866484)
    (29.9012, 0.866484)
    (30.2063, 0.866484)
    (30.5114, 0.866484)
    (30.8166, 0.866484)
    (31.1217, 0.866484)
    (31.4268, 0.866484)
    (31.7319, 0.866484)
    (32.0370, 0.866484)
    (32.3421, 0.866484)
    (32.6472, 0.866484)
    (32.9524, 0.866484)
    (33.2575, 0.866484)
    (33.5626, 0.866484)
    (33.8677, 0.866484)
    (34.1728, 0.866484)
    (34.4779, 0.866484)
    (34.7830, 0.883289)
    (35.0882, 0.883289)
    (35.3933, 0.883289)
    (35.6984, 0.883289)
    (36.0035, 0.883289)
    (36.3086, 0.883289)
    (36.6137, 0.883289)
    (36.9188, 0.883289)
    (37.2240, 0.883748)
    (37.5291, 0.883748)
    (37.8342, 0.883748)
    (38.1393, 0.886115)
    (38.4444, 0.886115)
    (38.7495, 0.886071)
    (39.0546, 0.886071)
    (39.3598, 0.886071)
    (39.6649, 0.886071)
    (39.9700, 0.886071)
    (40.2751, 0.886071)
    (40.5802, 0.886071)
    (40.8853, 0.886071)
    (41.1904, 0.886071)
    (41.4956, 0.886071)
    (41.8007, 0.886071)
    (42.1058, 0.886071)
    (42.4109, 0.886071)
    (42.7160, 0.886071)
    (43.0211, 0.886071)
    (43.3262, 0.886071)
    (43.6314, 0.886071)
    (43.9365, 0.886071)
    (44.2416, 0.886071)
    (44.5467, 0.886071)
    (44.8518, 0.886071)
    (45.1569, 0.886071)
    (45.4620, 0.886071)
    (45.7672, 0.886071)
    (46.0723, 0.886071)
    (46.3774, 0.886071)
    (46.6825, 0.886071)
    (46.9876, 0.885750)
    (47.2927, 0.885750)
    (47.5978, 0.885750)
    (47.9030, 0.885750)
    (48.2081, 0.885750)
    (48.5132, 0.885750)
    (48.8183, 0.885750)
    (49.1234, 0.885750)
    (49.4285, 0.885750)
    (49.7336, 0.885750)
    (50.0388, 0.885750)
    (50.3439, 0.885750)
    (50.6490, 0.885750)
    (50.9541, 0.885750)
    (51.2592, 0.885750)
    (51.5643, 0.885750)
    (51.8694, 0.885750)
    (52.1746, 0.885750)
    (52.4797, 0.885750)
    (52.7848, 0.885750)
    (53.0899, 0.885750)
    (53.3950, 0.885750)
    (53.7001, 0.885750)
    (54.0052, 0.908031)
    (54.3104, 0.908031)
    (54.6155, 0.908031)
    (54.9206, 0.908031)
    (55.2257, 0.908031)
    (55.5308, 0.910622)
    (55.8359, 0.910622)
    (56.1410, 0.910622)
    (56.4462, 0.910622)
    (56.7513, 0.910622)
    (57.0564, 0.910622)
    (57.3615, 0.910622)
    (57.6666, 0.910622)
    (57.9717, 0.910622)
    (58.2769, 0.910622)
    (58.5820, 0.910622)
    (58.8871, 0.910622)
    (59.1922, 0.910622)
    (59.4973, 0.910622)
    (59.8024, 0.910622)
    (60.1075, 0.910622)
    (60.4127, 0.910622)
    (60.7178, 0.910622)
};

\addplot[
    fill={rgb,255:red,174; green,78; blue,255}, 
    fill opacity=0.3,
    forget plot
]
fill between[of=rl_upper and rl_lower];

\addplot[
    color={rgb,255:red,174; green,78; blue,255}, 
    draw opacity=0.75,
    line width=3pt,
    mark=none,
]
coordinates {
    (0.3051, 0.000000)
    (0.6102, 0.000000)
    (0.9153, 0.000000)
    (1.2205, 0.016667)
    (1.5256, 0.143531)
    (1.8307, 0.143531)
    (2.1358, 0.143531)
    (2.4409, 0.294189)
    (2.7460, 0.420630)
    (3.0511, 0.420630)
    (3.3563, 0.420630)
    (3.6614, 0.420630)
    (3.9665, 0.420630)
    (4.2716, 0.420630)
    (4.5767, 0.420630)
    (4.8818, 0.420630)
    (5.1869, 0.553782)
    (5.4921, 0.553782)
    (5.7972, 0.602466)
    (6.1023, 0.602466)
    (6.4074, 0.602466)
    (6.7125, 0.602466)
    (7.0176, 0.602466)
    (7.3227, 0.602466)
    (7.6279, 0.602466)
    (7.9330, 0.602466)
    (8.2381, 0.602466)
    (8.5432, 0.602466)
    (8.8483, 0.602466)
    (9.1534, 0.602466)
    (9.4585, 0.621894)
    (9.7637, 0.621894)
    (10.0688, 0.621894)
    (10.3739, 0.621894)
    (10.6790, 0.621894)
    (10.9841, 0.659314)
    (11.2892, 0.659314)
    (11.5943, 0.659314)
    (11.8995, 0.670847)
    (12.2046, 0.670847)
    (12.5097, 0.670847)
    (12.8148, 0.670847)
    (13.1199, 0.670847)
    (13.4250, 0.681734)
    (13.7301, 0.681734)
    (14.0353, 0.681734)
    (14.3404, 0.714936)
    (14.6455, 0.714936)
    (14.9506, 0.714936)
    (15.2557, 0.714936)
    (15.5608, 0.714936)
    (15.8659, 0.714936)
    (16.1711, 0.750096)
    (16.4762, 0.825307)
    (16.7813, 0.825307)
    (17.0864, 0.825307)
    (17.3915, 0.825307)
    (17.6966, 0.825307)
    (18.0017, 0.825307)
    (18.3069, 0.825307)
    (18.6120, 0.825307)
    (18.9171, 0.825307)
    (19.2222, 0.825307)
    (19.5273, 0.842248)
    (19.8324, 0.842248)
    (20.1376, 0.842256)
    (20.4427, 0.842256)
    (20.7478, 0.842256)
    (21.0529, 0.842256)
    (21.3580, 0.848664)
    (21.6631, 0.859385)
    (21.9682, 0.859385)
    (22.2734, 0.859385)
    (22.5785, 0.859385)
    (22.8836, 0.859385)
    (23.1887, 0.859500)
    (23.4938, 0.859500)
    (23.7989, 0.859500)
    (24.1040, 0.875355)
    (24.4092, 0.897734)
    (24.7143, 0.897734)
    (25.0194, 0.897734)
    (25.3245, 0.897734)
    (25.6296, 0.897734)
    (25.9347, 0.897734)
    (26.2398, 0.897734)
    (26.5450, 0.897734)
    (26.8501, 0.897734)
    (27.1552, 0.897734)
    (27.4603, 0.897734)
    (27.7654, 0.897734)
    (28.0705, 0.897734)
    (28.3756, 0.897734)
    (28.6808, 0.897734)
    (28.9859, 0.897734)
    (29.2910, 0.897734)
    (29.5961, 0.897734)
    (29.9012, 0.897734)
    (30.2063, 0.897734)
    (30.5114, 0.897734)
    (30.8166, 0.897734)
    (31.1217, 0.897734)
    (31.4268, 0.897734)
    (31.7319, 0.897734)
    (32.0370, 0.897734)
    (32.3421, 0.897734)
    (32.6472, 0.897734)
    (32.9524, 0.897734)
    (33.2575, 0.897734)
    (33.5626, 0.897734)
    (33.8677, 0.897734)
    (34.1728, 0.897734)
    (34.4779, 0.897734)
    (34.7830, 0.907820)
    (35.0882, 0.907820)
    (35.3933, 0.907820)
    (35.6984, 0.907820)
    (36.0035, 0.907820)
    (36.3086, 0.907820)
    (36.6137, 0.907820)
    (36.9188, 0.907820)
    (37.2240, 0.915073)
    (37.5291, 0.915073)
    (37.8342, 0.915073)
    (38.1393, 0.915937)
    (38.4444, 0.915937)
    (38.7495, 0.916067)
    (39.0546, 0.916067)
    (39.3598, 0.916067)
    (39.6649, 0.916067)
    (39.9700, 0.916067)
    (40.2751, 0.916067)
    (40.5802, 0.916067)
    (40.8853, 0.916067)
    (41.1904, 0.916067)
    (41.4956, 0.916067)
    (41.8007, 0.916067)
    (42.1058, 0.916067)
    (42.4109, 0.916067)
    (42.7160, 0.916067)
    (43.0211, 0.916067)
    (43.3262, 0.916067)
    (43.6314, 0.916067)
    (43.9365, 0.916067)
    (44.2416, 0.916067)
    (44.5467, 0.916067)
    (44.8518, 0.916067)
    (45.1569, 0.916067)
    (45.4620, 0.916067)
    (45.7672, 0.916067)
    (46.0723, 0.916067)
    (46.3774, 0.916067)
    (46.6825, 0.916067)
    (46.9876, 0.916923)
    (47.2927, 0.916923)
    (47.5978, 0.916923)
    (47.9030, 0.916923)
    (48.2081, 0.916923)
    (48.5132, 0.916923)
    (48.8183, 0.916923)
    (49.1234, 0.916923)
    (49.4285, 0.916923)
    (49.7336, 0.916923)
    (50.0388, 0.916923)
    (50.3439, 0.916923)
    (50.6490, 0.916923)
    (50.9541, 0.916923)
    (51.2592, 0.916923)
    (51.5643, 0.916923)
    (51.8694, 0.916923)
    (52.1746, 0.916923)
    (52.4797, 0.916923)
    (52.7848, 0.916923)
    (53.0899, 0.916923)
    (53.3950, 0.916923)
    (53.7001, 0.916923)
    (54.0052, 0.926386)
    (54.3104, 0.926386)
    (54.6155, 0.926386)
    (54.9206, 0.926386)
    (55.2257, 0.926386)
    (55.5308, 0.927832)
    (55.8359, 0.927832)
    (56.1410, 0.927832)
    (56.4462, 0.927832)
    (56.7513, 0.927832)
    (57.0564, 0.927832)
    (57.3615, 0.927832)
    (57.6666, 0.927832)
    (57.9717, 0.927832)
    (58.2769, 0.927832)
    (58.5820, 0.927832)
    (58.8871, 0.927832)
    (59.1922, 0.927832)
    (59.4973, 0.927832)
    (59.8024, 0.927832)
    (60.1075, 0.927832)
    (60.4127, 0.927832)
    (60.7178, 0.927832)
};
\addlegendentry{Hierarchical RL}

\end{axis}
\end{tikzpicture}

%% file: figures/spot_tasks_freeze_frame.tex
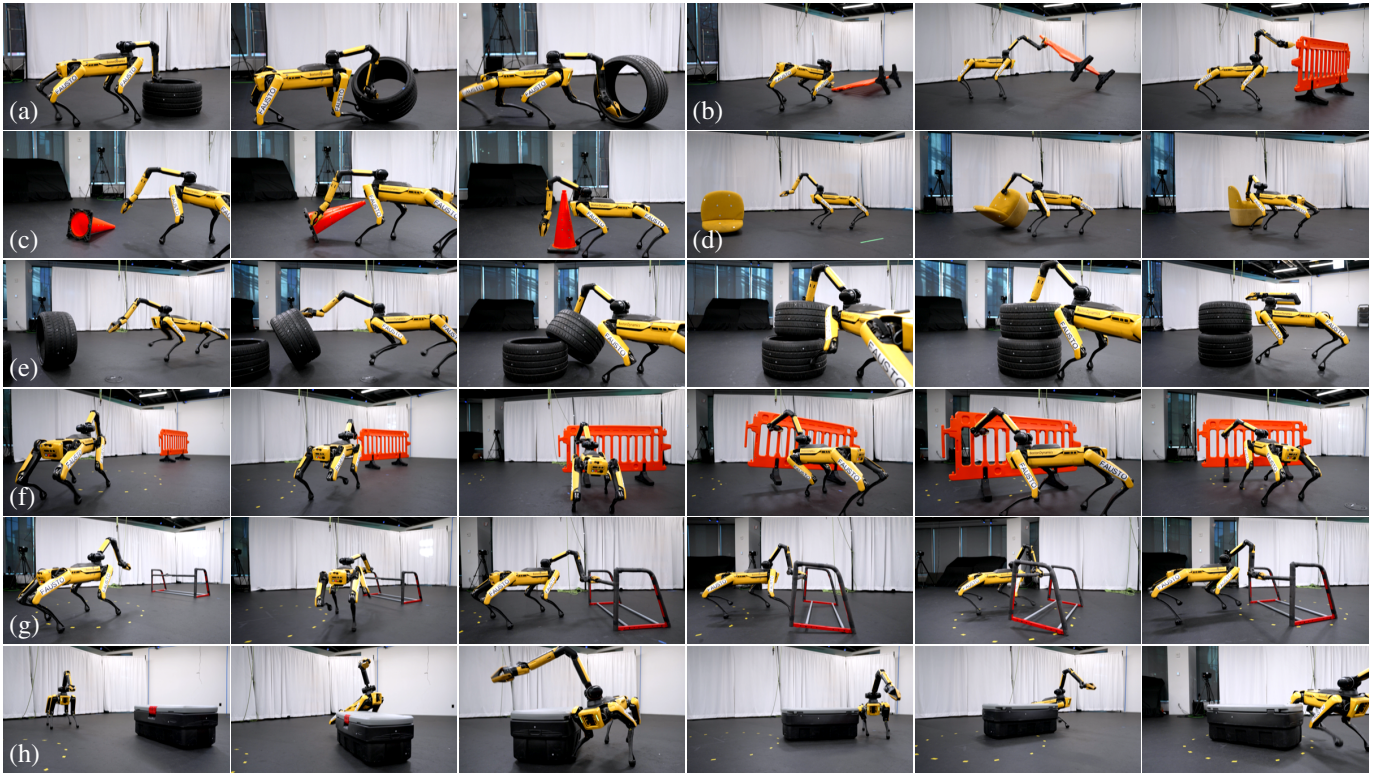
\begin{figure*}[t]
\centering
\setlength{\tabcolsep}{0pt}
\begin{tabular}{@{}c@{\hspace{2pt}}c@{}}
    \multicolumn{2}{c}{\renewcommand{\arraystretch}{0}\begin{tabular}{@{}c@{\hspace{1pt}}c@{\hspace{1pt}}c@{\hspace{1pt}}c@{\hspace{1pt}}c@{\hspace{1pt}}c@{}}
        \begin{tikzpicture}
            \node[anchor=south west,inner sep=0] (img) {\includegraphics[width=\dimexpr(\textwidth-5pt)/6\relax]{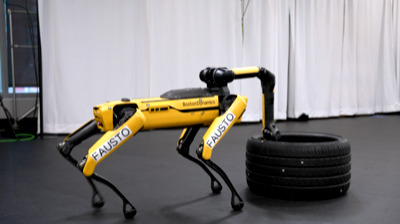}};
            \node[anchor=south west,inner sep=2pt] at (img.south west) {\color{white}(a)};
        \end{tikzpicture} &
        \begin{tikzpicture}
            \node[anchor=south west,inner sep=0] (img) {\includegraphics[width=\dimexpr(\textwidth-5pt)/6\relax]{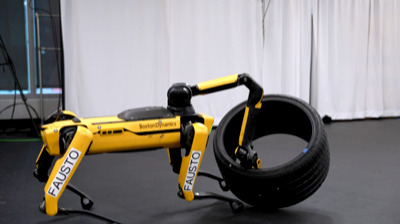}};
        \end{tikzpicture} &
        \begin{tikzpicture}
            \node[anchor=south west,inner sep=0] (img) {\includegraphics[width=\dimexpr(\textwidth-5pt)/6\relax]{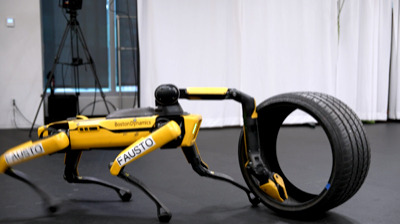}};
        \end{tikzpicture} &
        \begin{tikzpicture}
            \node[anchor=south west,inner sep=0] (img) {\includegraphics[width=\dimexpr(\textwidth-5pt)/6\relax]{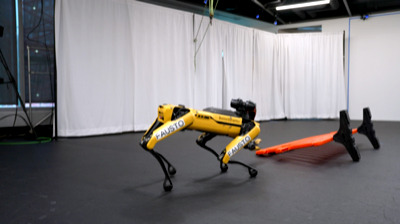}};
            \node[anchor=south west,inner sep=2pt] at (img.south west) {\color{white}(b)};
        \end{tikzpicture} &
        \begin{tikzpicture}
            \node[anchor=south west,inner sep=0] (img) {\includegraphics[width=\dimexpr(\textwidth-5pt)/6\relax]{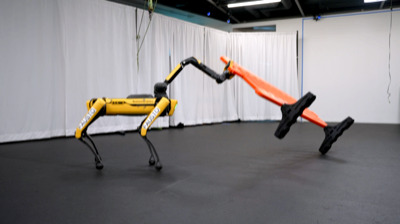}};
        \end{tikzpicture} &
        \begin{tikzpicture}
            \node[anchor=south west,inner sep=0] (img) {\includegraphics[width=\dimexpr(\textwidth-5pt)/6\relax]{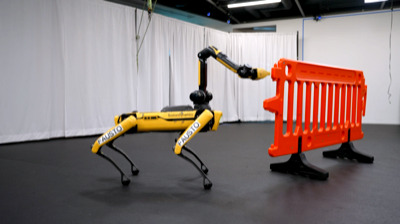}};
        \end{tikzpicture}
    \end{tabular}} \\
    \noalign{\vskip 1pt}

    \multicolumn{2}{c}{\renewcommand{\arraystretch}{0}\begin{tabular}{@{}c@{\hspace{1pt}}c@{\hspace{1pt}}c@{\hspace{1pt}}c@{\hspace{1pt}}c@{\hspace{1pt}}c@{}}
        \begin{tikzpicture}
            \node[anchor=south west,inner sep=0] (img) {\includegraphics[width=\dimexpr(\textwidth-5pt)/6\relax]{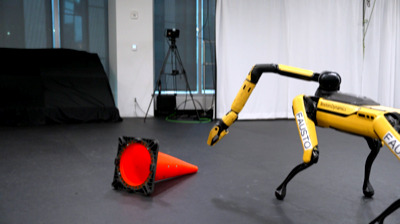}};
            \node[anchor=south west,inner sep=2pt] at (img.south west) {\color{white}(c)};
        \end{tikzpicture} &
        \begin{tikzpicture}
            \node[anchor=south west,inner sep=0] (img) {\includegraphics[width=\dimexpr(\textwidth-5pt)/6\relax]{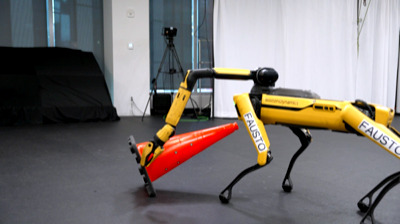}};
        \end{tikzpicture} &
        \begin{tikzpicture}
            \node[anchor=south west,inner sep=0] (img) {\includegraphics[width=\dimexpr(\textwidth-5pt)/6\relax]{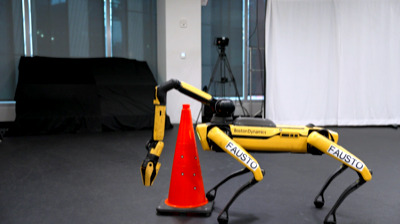}};
        \end{tikzpicture} &
        \begin{tikzpicture}
            \node[anchor=south west,inner sep=0] (img) {\includegraphics[width=\dimexpr(\textwidth-5pt)/6\relax]{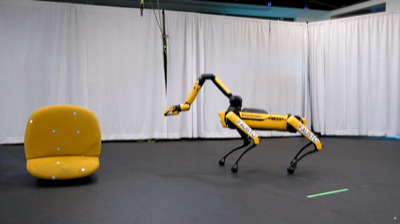}};
            \node[anchor=south west,inner sep=2pt] at (img.south west) {\color{white}(d)};
        \end{tikzpicture} &
        \begin{tikzpicture}
            \node[anchor=south west,inner sep=0] (img) {\includegraphics[width=\dimexpr(\textwidth-5pt)/6\relax]{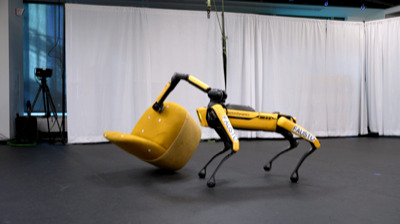}};
        \end{tikzpicture} &
        \begin{tikzpicture}
            \node[anchor=south west,inner sep=0] (img) {\includegraphics[width=\dimexpr(\textwidth-5pt)/6\relax]{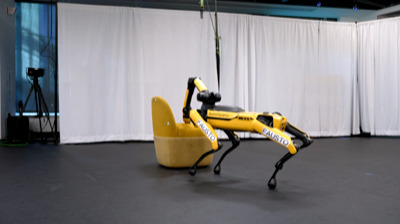}};
        \end{tikzpicture}
    \end{tabular}} \\
    \noalign{\vskip 1pt}

    \multicolumn{2}{c}{\renewcommand{\arraystretch}{0}\begin{tabular}{@{}c@{\hspace{1pt}}c@{\hspace{1pt}}c@{\hspace{1pt}}c@{\hspace{1pt}}c@{\hspace{1pt}}c@{}}
        \begin{tikzpicture}
            \node[anchor=south west,inner sep=0] (img) {\includegraphics[width=\dimexpr(\textwidth-5pt)/6\relax]{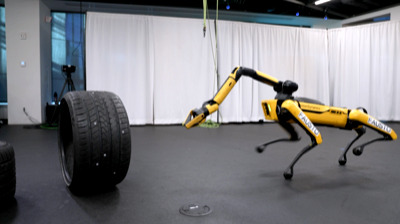}};
            \node[anchor=south west,inner sep=2pt] at (img.south west) {\color{white}(e)};
        \end{tikzpicture} &
        \begin{tikzpicture}
            \node[anchor=south west,inner sep=0] (img) {\includegraphics[width=\dimexpr(\textwidth-5pt)/6\relax]{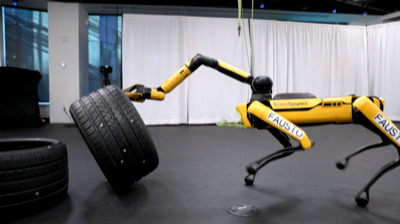}};
        \end{tikzpicture} &
        \begin{tikzpicture}
            \node[anchor=south west,inner sep=0] (img) {\includegraphics[width=\dimexpr(\textwidth-5pt)/6\relax]{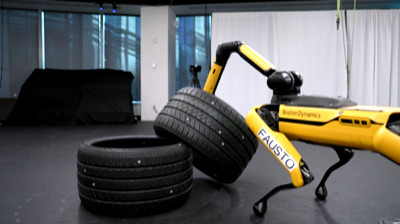}};
        \end{tikzpicture} &
        \begin{tikzpicture}
            \node[anchor=south west,inner sep=0] (img) {\includegraphics[width=\dimexpr(\textwidth-5pt)/6\relax]{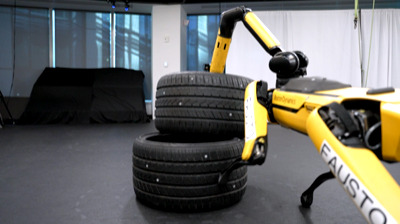}};
        \end{tikzpicture} &
        \begin{tikzpicture}
            \node[anchor=south west,inner sep=0] (img) {\includegraphics[width=\dimexpr(\textwidth-5pt)/6\relax]{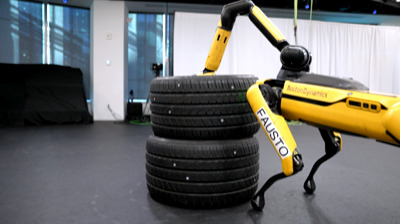}};
        \end{tikzpicture} &
        \begin{tikzpicture}
            \node[anchor=south west,inner sep=0] (img) {\includegraphics[width=\dimexpr(\textwidth-5pt)/6\relax]{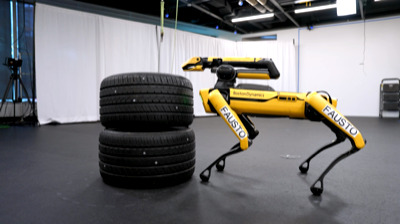}};
        \end{tikzpicture}
    \end{tabular}} \\
    \noalign{\vskip 1pt}
    \multicolumn{2}{c}{\renewcommand{\arraystretch}{0}\begin{tabular}{@{}c@{\hspace{1pt}}c@{\hspace{1pt}}c@{\hspace{1pt}}c@{\hspace{1pt}}c@{\hspace{1pt}}c@{}}
        \begin{tikzpicture}
            \node[anchor=south west,inner sep=0] (img) {\includegraphics[width=\dimexpr(\textwidth-5pt)/6\relax]{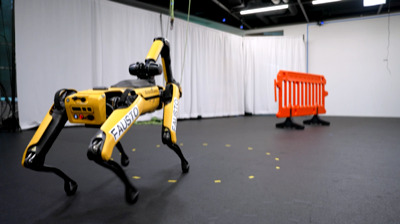}};
            \node[anchor=south west,inner sep=2pt] at (img.south west) {\color{white}(f)};
        \end{tikzpicture} &
        \begin{tikzpicture}
            \node[anchor=south west,inner sep=0] (img) {\includegraphics[width=\dimexpr(\textwidth-5pt)/6\relax]{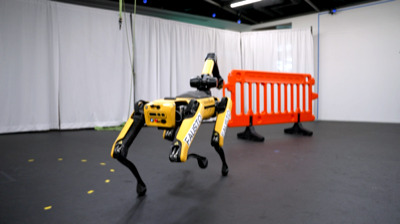}};
        \end{tikzpicture} &
        \begin{tikzpicture}
            \node[anchor=south west,inner sep=0] (img) {\includegraphics[width=\dimexpr(\textwidth-5pt)/6\relax]{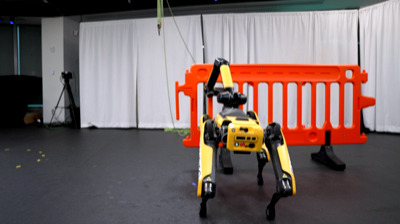}};
        \end{tikzpicture} &
        \begin{tikzpicture}
            \node[anchor=south west,inner sep=0] (img) {\includegraphics[width=\dimexpr(\textwidth-5pt)/6\relax]{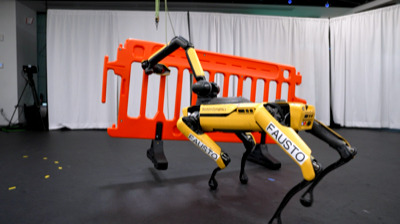}};
        \end{tikzpicture} &
        \begin{tikzpicture}
            \node[anchor=south west,inner sep=0] (img) {\includegraphics[width=\dimexpr(\textwidth-5pt)/6\relax]{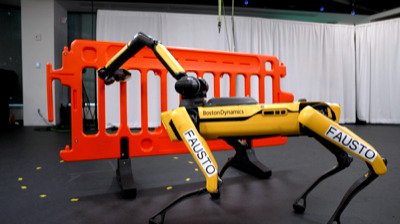}};
        \end{tikzpicture} &
        \begin{tikzpicture}
            \node[anchor=south west,inner sep=0] (img) {\includegraphics[width=\dimexpr(\textwidth-5pt)/6\relax]{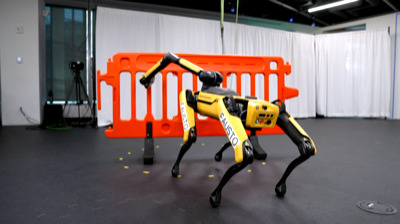}};
        \end{tikzpicture}
    \end{tabular}} \\
    \noalign{\vskip 1pt}
    \multicolumn{2}{c}{\renewcommand{\arraystretch}{0}\begin{tabular}{@{}c@{\hspace{1pt}}c@{\hspace{1pt}}c@{\hspace{1pt}}c@{\hspace{1pt}}c@{\hspace{1pt}}c@{}}
        \begin{tikzpicture}
            \node[anchor=south west,inner sep=0] (img) {\includegraphics[width=\dimexpr(\textwidth-5pt)/6\relax]{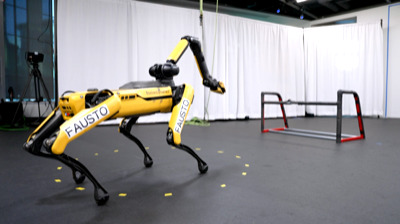}};
            \node[anchor=south west,inner sep=2pt] at (img.south west) {\color{white}(g)};
        \end{tikzpicture} &
        \begin{tikzpicture}
            \node[anchor=south west,inner sep=0] (img) {\includegraphics[width=\dimexpr(\textwidth-5pt)/6\relax]{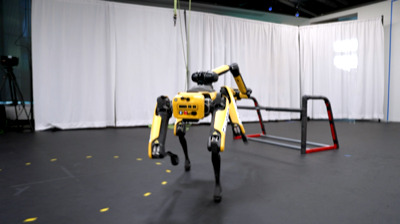}};
        \end{tikzpicture} &
        \begin{tikzpicture}
            \node[anchor=south west,inner sep=0] (img) {\includegraphics[width=\dimexpr(\textwidth-5pt)/6\relax]{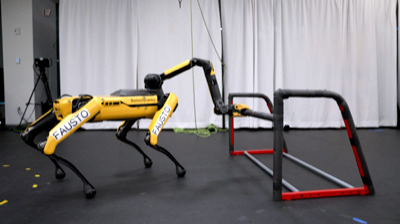}};
        \end{tikzpicture} &
        \begin{tikzpicture}
            \node[anchor=south west,inner sep=0] (img) {\includegraphics[width=\dimexpr(\textwidth-5pt)/6\relax]{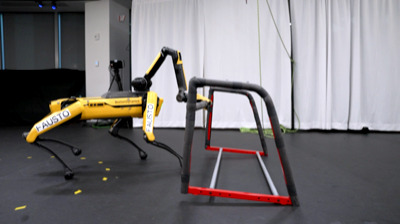}};
        \end{tikzpicture} &
        \begin{tikzpicture}
            \node[anchor=south west,inner sep=0] (img) {\includegraphics[width=\dimexpr(\textwidth-5pt)/6\relax]{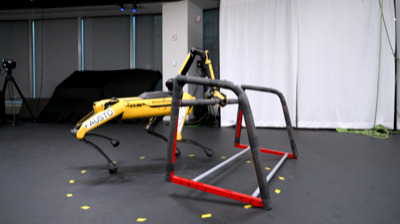}};
        \end{tikzpicture} &
        \begin{tikzpicture}
            \node[anchor=south west,inner sep=0] (img) {\includegraphics[width=\dimexpr(\textwidth-5pt)/6\relax]{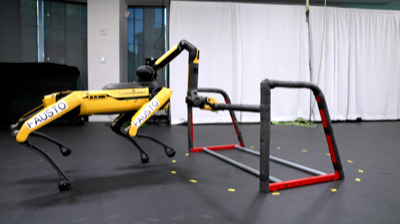}};
        \end{tikzpicture}
    \end{tabular}} \\
    \noalign{\vskip 1pt}
    \multicolumn{2}{c}{\renewcommand{\arraystretch}{0}\begin{tabular}{@{}c@{\hspace{1pt}}c@{\hspace{1pt}}c@{\hspace{1pt}}c@{\hspace{1pt}}c@{\hspace{1pt}}c@{}}
        \begin{tikzpicture}
            \node[anchor=south west,inner sep=0] (img) {\includegraphics[width=\dimexpr(\textwidth-5pt)/6\relax]{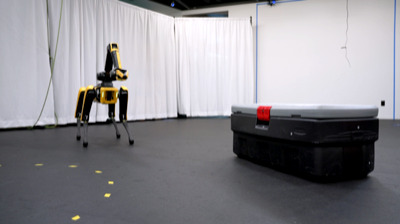}};
            \node[anchor=south west,inner sep=2pt] at (img.south west) {\color{white}(h)};
        \end{tikzpicture} &
        \begin{tikzpicture}
            \node[anchor=south west,inner sep=0] (img) {\includegraphics[width=\dimexpr(\textwidth-5pt)/6\relax]{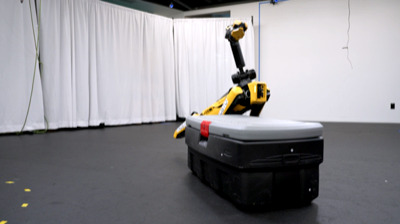}};
        \end{tikzpicture} &
        \begin{tikzpicture}
            \node[anchor=south west,inner sep=0] (img) {\includegraphics[width=\dimexpr(\textwidth-5pt)/6\relax]{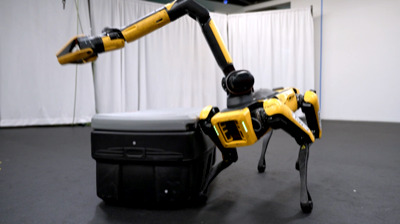}};
        \end{tikzpicture} &
        \begin{tikzpicture}
            \node[anchor=south west,inner sep=0] (img) {\includegraphics[width=\dimexpr(\textwidth-5pt)/6\relax]{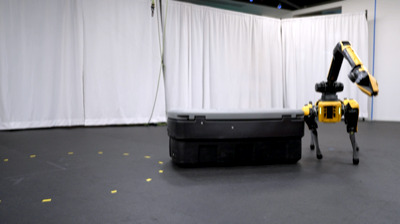}};
        \end{tikzpicture} &
        \begin{tikzpicture}
            \node[anchor=south west,inner sep=0] (img) {\includegraphics[width=\dimexpr(\textwidth-5pt)/6\relax]{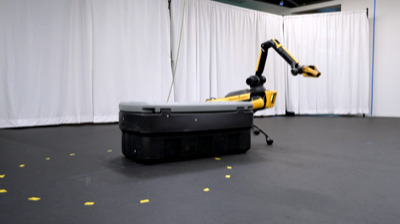}};
        \end{tikzpicture} &
        \begin{tikzpicture}
            \node[anchor=south west,inner sep=0] (img) {\includegraphics[width=\dimexpr(\textwidth-5pt)/6\relax]{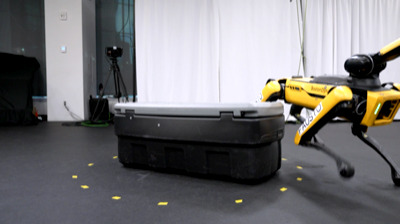}};
        \end{tikzpicture}
    \end{tabular}}
\end{tabular}
\caption{Freeze-frame sequences showing Spot robot task progressions: (a) uprighting a tire, (b) uprighting a crowd control barrier, (c) uprighting a traffic cone, (d) uprighting a chair, (e) stacking a two tires, (f) dragging a crowd control barrier, (g) dragging a tire rack, and (h) pushing a heavy box.}
\vspace{-15pt}
\label{fig:spot_tasks_freeze_frame}
\end{figure*}

%% file: figures/g1_tasks_freeze_frame.tex
\begin{figure}
\centering
\setlength{\tabcolsep}{0.5pt}
\begin{tabular}{@{}c@{}}
    \renewcommand{\arraystretch}{0}\begin{tabular}{@{}c@{\hspace{1pt}}c@{\hspace{1pt}}c@{}}
        \begin{tikzpicture}
            \node[anchor=south west,inner sep=0] (img) {\includegraphics[width=\dimexpr(\columnwidth-5pt)/3\relax]{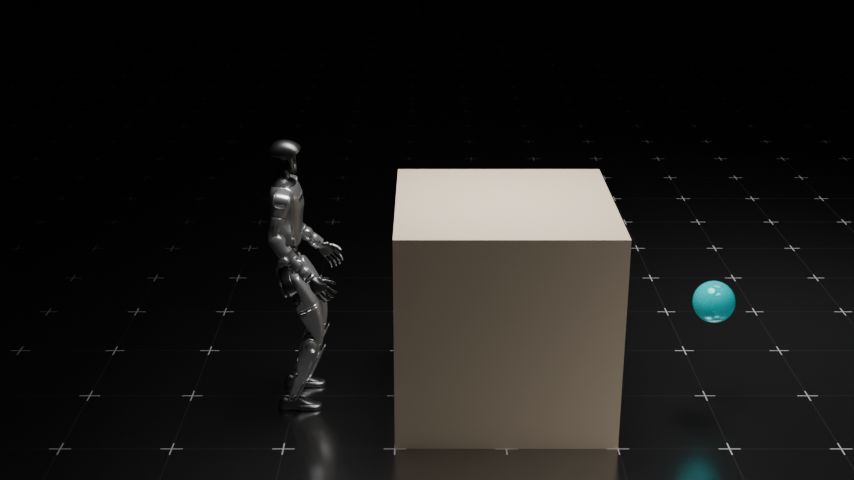}};
            \node[anchor=south west,inner sep=2pt] at (0,0) {\color{white}(a)};
        \end{tikzpicture} &
        \begin{tikzpicture}
            \node[anchor=south west,inner sep=0] (img) {\includegraphics[width=\dimexpr(\columnwidth-5pt)/3\relax]{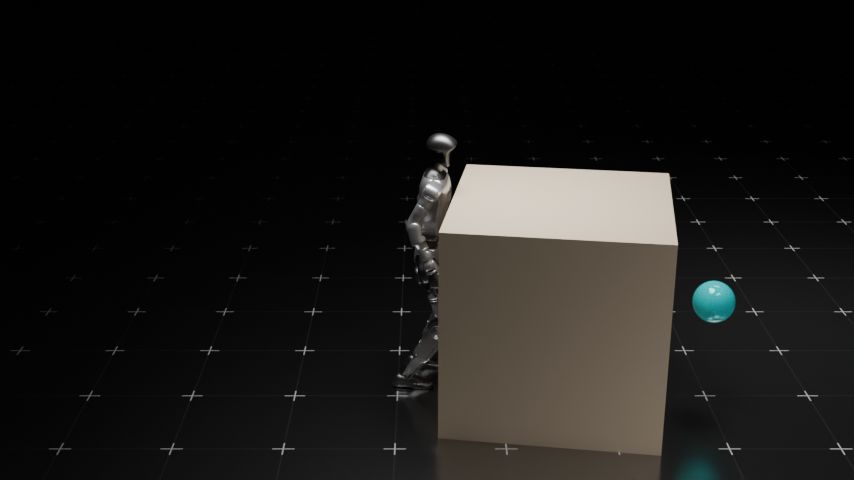}};
        \end{tikzpicture} &
        \begin{tikzpicture}
            \node[anchor=south west,inner sep=0] (img) {\includegraphics[width=\dimexpr(\columnwidth-5pt)/3\relax]{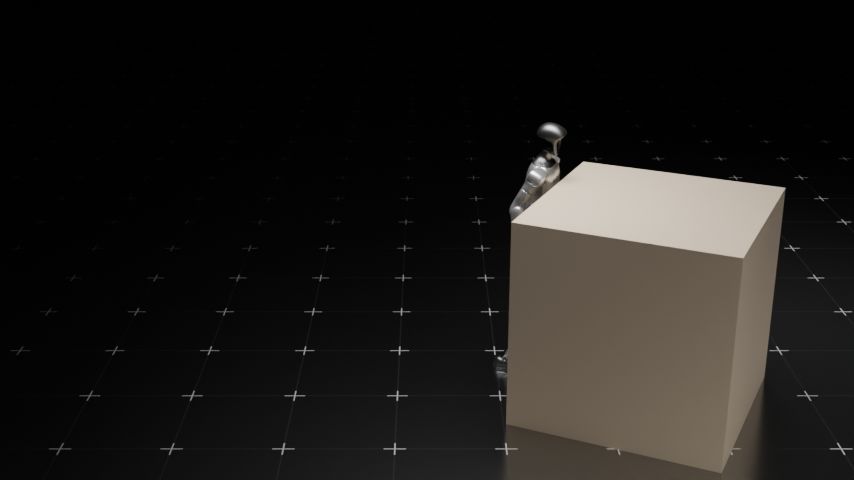}};
        \end{tikzpicture}
    \end{tabular} \\
    \noalign{\vskip 1pt}
    \renewcommand{\arraystretch}{0}\begin{tabular}{@{}c@{\hspace{1pt}}c@{\hspace{1pt}}c@{}}
        \begin{tikzpicture}
            \node[anchor=south west,inner sep=0] (img) {\includegraphics[width=\dimexpr(\columnwidth-5pt)/3\relax]{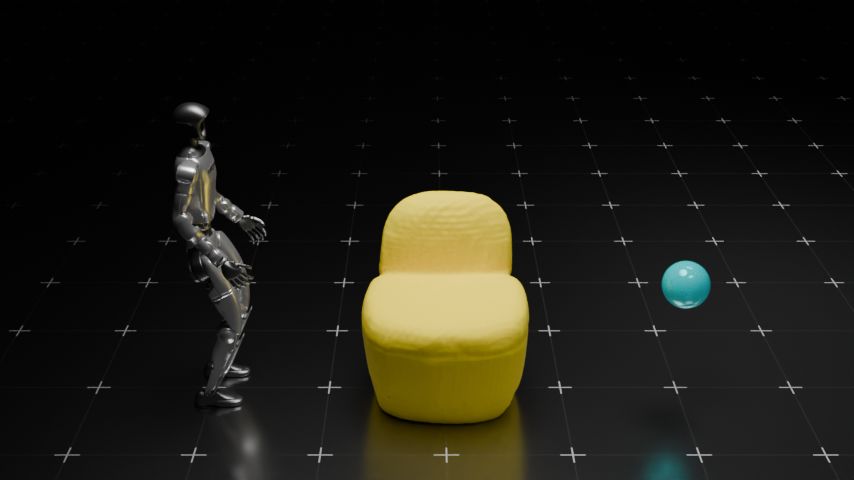}};
            \node[anchor=south west,inner sep=2pt] at (0,0) {\color{white}(b)};
        \end{tikzpicture} &
        \begin{tikzpicture}
            \node[anchor=south west,inner sep=0] (img) {\includegraphics[width=\dimexpr(\columnwidth-5pt)/3\relax]{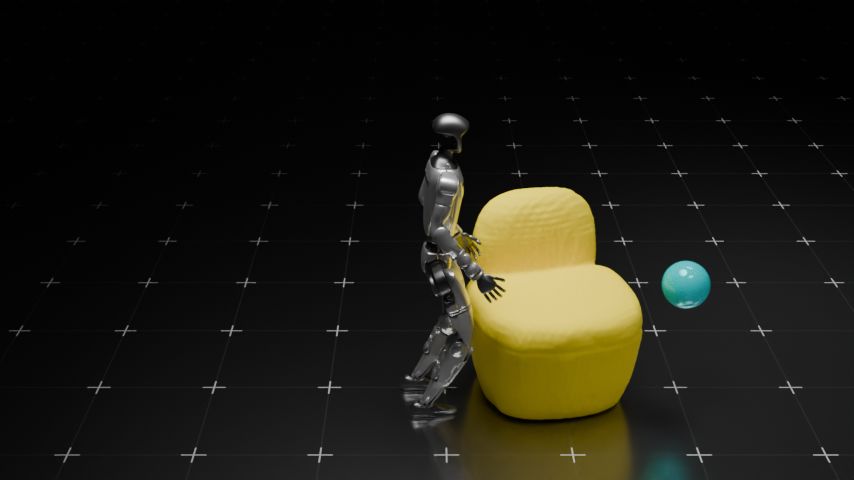}};
        \end{tikzpicture} &
        \begin{tikzpicture}
            \node[anchor=south west,inner sep=0] (img) {\includegraphics[width=\dimexpr(\columnwidth-5pt)/3\relax]{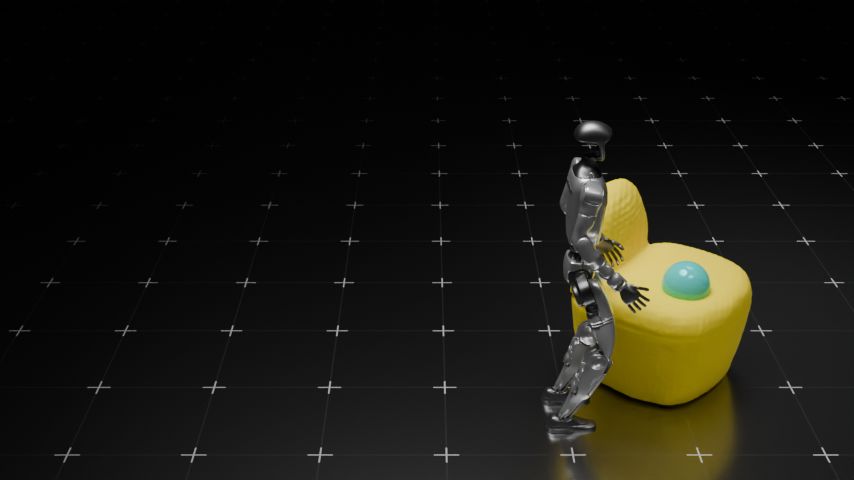}};
        \end{tikzpicture}
    \end{tabular} \\
    \noalign{\vskip 1pt}
    \renewcommand{\arraystretch}{0}\begin{tabular}{@{}c@{\hspace{1pt}}c@{\hspace{1pt}}c@{}}
        \begin{tikzpicture}
            \node[anchor=south west,inner sep=0] (img) {\includegraphics[width=\dimexpr(\columnwidth-5pt)/3\relax]{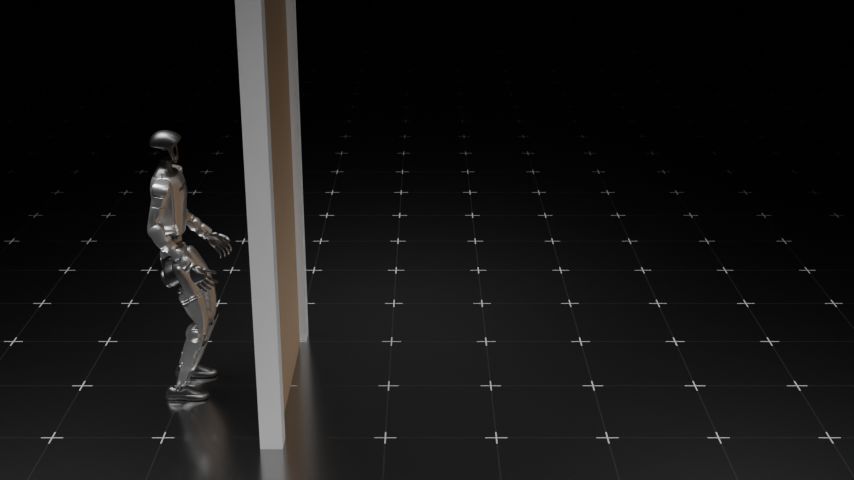}};
            \node[anchor=south west,inner sep=2pt] at (0,0) {\color{white}(c)};
        \end{tikzpicture} &
        \begin{tikzpicture}
            \node[anchor=south west,inner sep=0] (img) {\includegraphics[width=\dimexpr(\columnwidth-5pt)/3\relax]{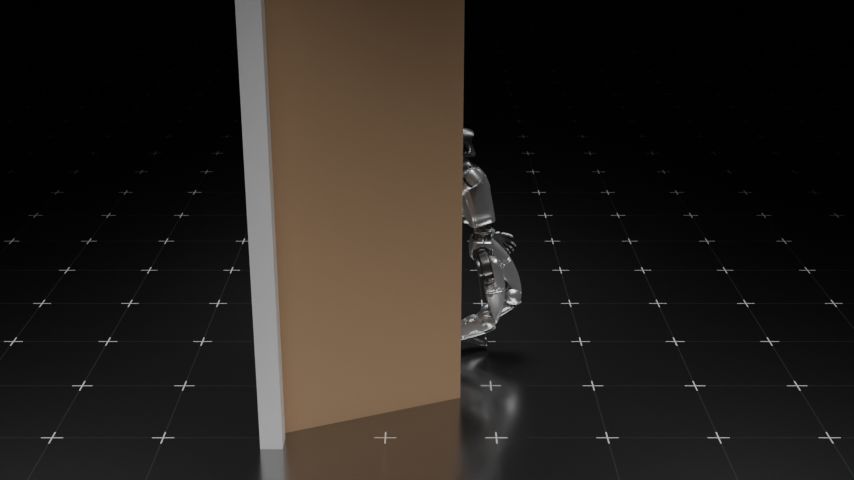}};
        \end{tikzpicture} &
        \begin{tikzpicture}
            \node[anchor=south west,inner sep=0] (img) {\includegraphics[width=\dimexpr(\columnwidth-5pt)/3\relax]{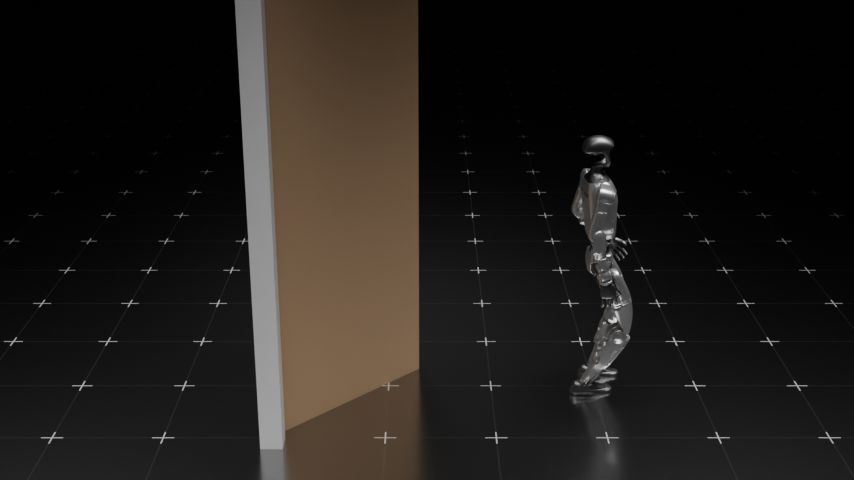}};
        \end{tikzpicture}
    \end{tabular} \\
    \noalign{\vskip 1pt}
    \renewcommand{\arraystretch}{0}\begin{tabular}{@{}c@{\hspace{1pt}}c@{\hspace{1pt}}c@{}}
        \begin{tikzpicture}
            \node[anchor=south west,inner sep=0] (img) {\includegraphics[width=\dimexpr(\columnwidth-5pt)/3\relax]{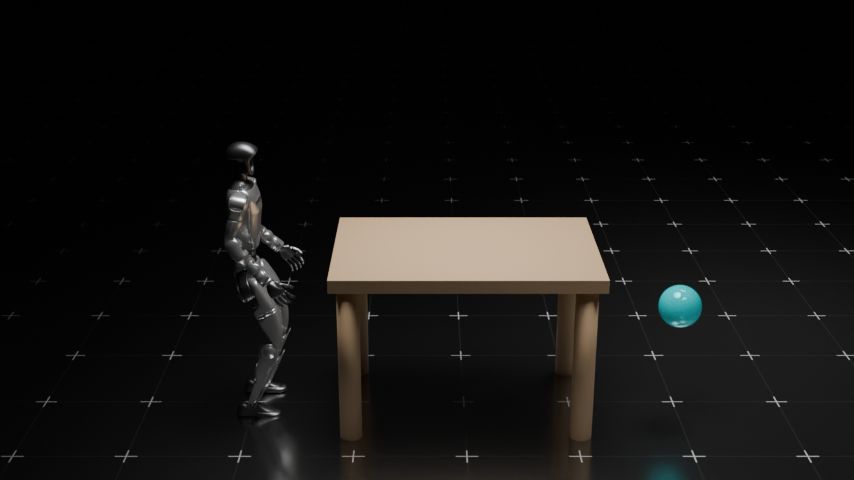}};
            \node[anchor=south west,inner sep=2pt] at (0,0) {\color{white}(d)};
        \end{tikzpicture} &
        \begin{tikzpicture}
            \node[anchor=south west,inner sep=0] (img) {\includegraphics[width=\dimexpr(\columnwidth-5pt)/3\relax]{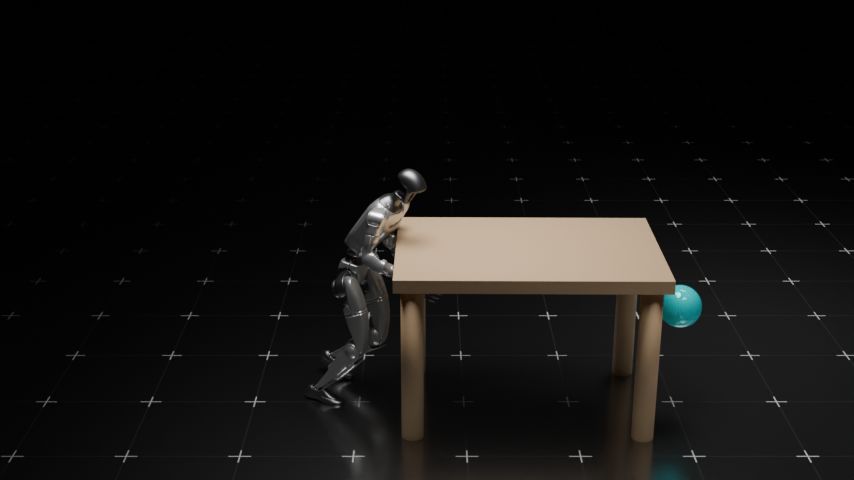}};
        \end{tikzpicture} &
        \begin{tikzpicture}
            \node[anchor=south west,inner sep=0] (img) {\includegraphics[width=\dimexpr(\columnwidth-5pt)/3\relax]{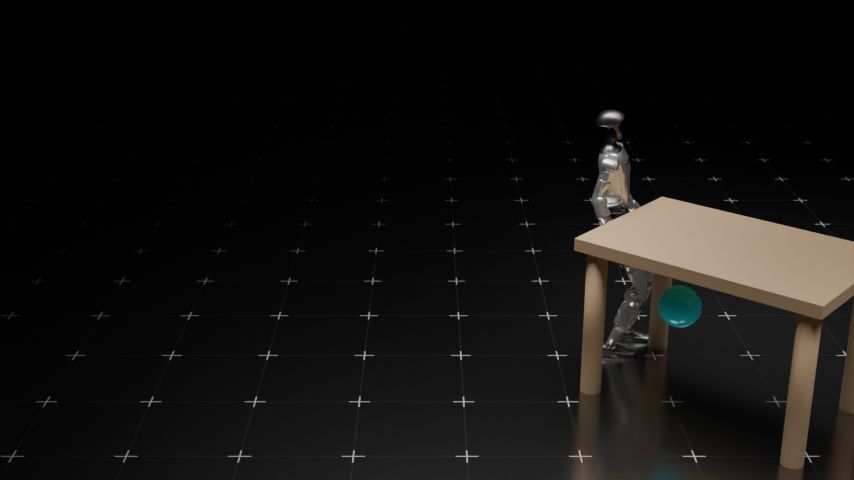}};
        \end{tikzpicture}
    \end{tabular}
\end{tabular}
\caption{Freeze-frame sequences showing G1 humanoid task progression: (a) pushing a box, (b) pushing a chair, (c) opening a door, and (d) pushing a table to a goal position (blue sphere).}
\label{fig:g1_tasks_freeze_frame}
\end{figure}
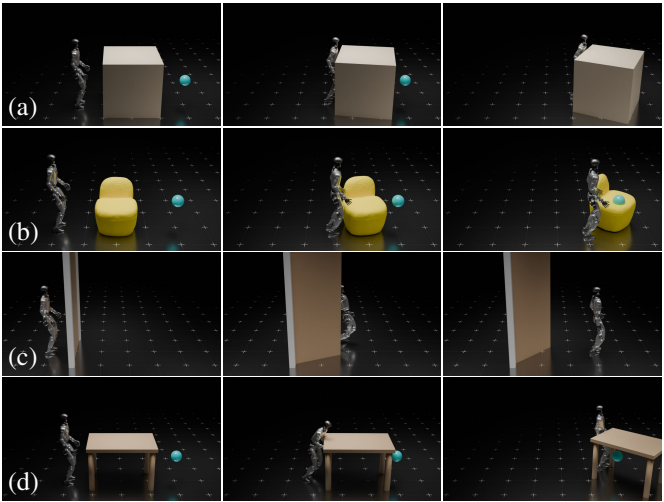

%% file: sections/conclusions.tex
\section{Conclusions and Future Work}\label{sec: conclusions}
This paper explores the combination of RL pre-training and test-time sample-based MPC for whole-body loco-manipulation. We find this approach, which we call Sumo, excels at challenging tasks involving large and heavy objects that cannot be solved via pick-and-place-style manipulation.
We also find that the decomposition of difficult loco-manipulation tasks into a two-stage optimization problem makes the sub-problems much more tractable. Additionally, we show that using sample-based MPC as a high-level policy enables test-time task flexibility and generalization that is difficult for RL.

Sumo still has several shortcomings and avenues for future work: First, \revised{our current system depends on off-board estimation and planning.} \revised{Recent powerful compact-form-factor computers, such as the M-series Mac mini, make deploying Sumo-style algorithms computationally tractable. However, we leave fully on-board deployment on the Spot and G1 robots for future work.} Second, as an algorithm relying entirely on sim-to-real transfer, the ``sim-to-real gap'' remains a significant challenge. While modeling the robot and object dynamics perfectly is impossible and often not necessary, building a ``good enough'' model for control remains an art that requires some trial and error. In fact, we believe real-time MPC as the high-level policy significantly accelerates this tuning loop compared to RL training. Nevertheless, Sumo can undoubtedly benefit from system-identification and model-learning methods that update object parameters on the fly from real-world interactions. Finally, Sumo, in its current form, does not leverage any human priors. Future work should explore incorporating pre-trained foundation models that can automatically guide robot behaviors without manually engineered rewards\revised{---for example, foundation and agentic models that supply goals, object models, or cost functions, which Sumo then executes through dynamic whole-body planning and control}.

%% file: sections/acknowledgement.tex
\section*{Acknowledgments}

\ifarxiv
We thank Emmanuel Panov for supporting hardware deployment and Jonathan Foster for building 3D object meshes used in the experiments. We also thank Swaminathan Gurumurthy for providing thoughtful feedback and discussions. Finally, we thank Yuval Tassa and Taylor Howell for answering MuJoCo-related questions. This work was done in part during an internship at the RAI Institute.

\fi
Coding agents like Claude Code and OpenAI Codex were used to assist in generating parts of the code for experiments included in the paper. AI systems were also used to proofread and correct grammatical errors in the manuscript.

%% file: sections/appendix.tex
\appendix
\section{Supplementary Material}
\input{sections/appendix_content}

%% file: sections/appendix_content.tex
\subsection{Task Rewards}
Here, we provide a detailed description of the task rewards from the demonstration section as deployed on the Spot robot on hardware and G1 robot in simulation. Please refer to the code for an exact implementation.

\revised{All task costs below share a common template: a task-objective term (orientation alignment or goal reaching), end-effector proximity terms, body-positioning terms, control regularization, and safety terms. Object-specific grasp and approach locations are fixed sites expressed in the object frame (e.g., the left and right grasp points on the crowd barrier); they move rigidly with the estimated object pose, and no online contact-point optimization is performed. Similarly, the desired robot positions $\mathbf{p}^{\text{des}}$ are computed analytically from the current object state; we give the box-push torso target below as a representative example. Control-regularization terms act on the high-level planner action $\mathbf{a}$, not on joint-level controls.}

\revised{\noindent\textbf{Safety cost.} All tasks share the safety cost
\begin{align}
    J_{\text{safety}} = w_{\text{fall}} \cdot \mathds{1}\left[ \min_{t} \, h_{\text{torso}}^{t} < h_{\text{min}} \right],
\end{align}
which penalizes any rollout in which the robot's torso (pelvis) height drops below a threshold at any point along the prediction horizon, with $w_{\text{fall}} = 2500$ and $h_{\text{min}} = 0.35$\,m for Spot and $0.6$\,m for the G1. Individual tasks add analogous indicator penalties for task-specific failures, e.g., the object tipping over during dragging or the robot leaving the hardware workspace.}

\revised{\noindent\textbf{Grasp cost.} Tasks that require grasping (barrier upright, barrier drag, and tire rack drag) detect grasps from gripper-closure resistance. Let $e = q_{\text{grip}} - c_{\text{grip}}$ denote the difference between the measured and commanded gripper joint positions, which is positive when an object physically blocks the gripper from closing. A rollout state is labeled as grasping when $e > 0.15$\,rad and the gripper is not fully closed, yielding
\begin{align}
    J_{\text{grasp}} = \; & - w_{\text{quality}} \cdot \mathds{1}\left[\text{grasping}\right] \cdot \min\left( e / 0.5, \, 1 \right) \nonumber \\
    & + w_{\text{false}} \cdot \mathds{1}\left[\text{gripper fully closed in free space}\right],
\end{align}
with $w_{\text{quality}} = 100$ and $w_{\text{false}} = 1000$: the first term rewards firm grasps in proportion to the sensed closure resistance, and the second penalizes closing the gripper on nothing, preventing the planner from exploiting false grasps during rollouts.}

\subsubsection{Spot Tasks}\mbox{}\\
\noindent\textbf{Tire Upright:}
The tire upright task uses a cost function with proximity terms to guide the robot's end-effectors toward the object, an orientation term to upright the tire, and regularization terms:
\begin{align}
    J_{\text{Tire Upright}} = \;
    & w_{\text{orient}} \cdot \exp\left( |{\mathbf{y}_{\text{tire}}}^z| / \sigma \right) \nonumber \\
    & + w_{\text{gripper}} \cdot \| \mathbf{p}_{\text{gripper}} - \mathbf{p}_{\text{gripper}}^{\text{des}} \|_2 \nonumber \\
    & + w_{\text{foot}} \cdot \min\left( \| \mathbf{p}_{\text{fr}} - \mathbf{p}_{\text{fr}}^{\text{des}} \|_2, \| \mathbf{p}_{\text{fl}} - \mathbf{p}_{\text{fl}}^{\text{des}} \|_2 \right) \nonumber \\
    & + w_{\text{torso}} \cdot \| \mathbf{p}_{\text{torso}} - \mathbf{p}_{\text{torso}}^{\text{des}} \|_2 \nonumber \\
    & + w_{\text{ctrl}} \cdot \| \mathbf{a} \|_2 + J_{\text{safety}}
\end{align}
where ${\mathbf{y}_{\text{tire}}}^z$ is the $z$-component of the tire's $y$-axis (zero when upright), $\sigma$ is a smoothing parameter, $\mathbf{p}_{\text{gripper}}$, $\mathbf{p}_{\text{fr}}$, $\mathbf{p}_{\text{fl}}$, and $\mathbf{p}_{\text{torso}}$ are the positions of the gripper, front-right foot, front-left foot, and torso, respectively. The desired positions $\mathbf{p}^{\text{des}}$ are computed dynamically based on the tire position, encouraging the robot to position itself around the tire for manipulation. \revised{Each desired position is a fixed offset expressed in the tire frame---a grasp point on the tire for the gripper, contact points for the feet, and a standoff pose for the torso---so that all targets move rigidly with the estimated tire pose.} $J_{\text{safety}}$ includes penalties for the robot falling.

\noindent\textbf{Crowd Barrier Upright:}
The crowd barrier upright task uses orientation alignment, grasp-aware proximity terms, and grasp quality feedback:
\begin{align}
    J_{\text{Barrier}} = \;
    & w_{\text{orient}} \cdot \left(1 - \exp\left( \alpha \cdot (\mathbf{z}_{\text{barrier}} \cdot \mathbf{z}_{\text{world}} - 1) \right)\right) \nonumber \\
    & + w_{\text{grasp}} \cdot \min\left( \| \mathbf{p}_{\text{grip}} - \mathbf{p}_{\text{grasp}}^{L} \|, \| \mathbf{p}_{\text{grip}} - \mathbf{p}_{\text{grasp}}^{R} \| \right) \nonumber \\
    & + w_{\text{grip}} \cdot \left( 1 - |\mathbf{x}_{\text{grip}} \cdot \mathbf{x}_{\text{barrier}}| + 1 - |\mathbf{y}_{\text{grip}} \cdot \mathbf{z}_{\text{barrier}}| \right) \nonumber \\
    & + w_{\text{approach}} \cdot \min\left( \| \mathbf{p}_{\text{torso}} - \mathbf{p}_{\text{appr}}^{L} \|, \| \mathbf{p}_{\text{torso}} - \mathbf{p}_{\text{appr}}^{R} \| \right) \nonumber \\
    & + w_{\text{vel}} \cdot \| \mathbf{v}_{\text{obj}} \|_2^2 + w_{\text{ctrl}} \cdot \| \mathbf{a} \|_2 + J_{\text{grasp}} + J_{\text{safety}}
\end{align}
where $\mathbf{z}_{\text{barrier}}$ is the barrier's $z$-axis (aligned with $\mathbf{z}_{\text{world}}$ when upright), $\alpha$ is a sparsity parameter, and $\mathbf{p}_{\text{grasp}}^{L/R}$ are left/right grasp points on the barrier. The gripper orientation cost aligns the gripper's $x$-axis with the barrier's long axis and the gripper's $y$-axis with the barrier's $z$-axis. $J_{\text{grasp}}$ rewards successful grasps detected via gripper resistance (position error when closing) and penalizes closing the gripper in empty space\revised{, as formalized at the beginning of this appendix}. $J_{\text{safety}}$ penalizes robot falling. \revised{The weights are $w_{\text{orient}}{=}100$ (with $\alpha{=}3$), $w_{\text{grasp}}{=}500$, $w_{\text{grip}}{=}100$, $w_{\text{approach}}{=}10$, $w_{\text{vel}}{=}10$, and $w_{\text{ctrl}}{=}2$.}

\noindent\textbf{Traffic Cone Upright:}
The traffic cone upright task uses orientation alignment and proximity terms:
\begin{align}
    J_{\text{Cone}} = \;
    & w_{\text{orient}} \cdot \left(1 - \exp\left( \alpha \cdot (\mathbf{z}_{\text{cone}} \cdot \mathbf{z}_{\text{world}} - 1) \right)\right) \nonumber \\
    & + w_{\text{gripper}} \cdot \| \mathbf{p}_{\text{gripper}} - \mathbf{p}_{\text{cone}} \|_2 \nonumber \\
    & - w_{\text{torso}} \cdot \min\left( d_{\text{thresh}}, \| \mathbf{p}_{\text{torso}} - \mathbf{p}_{\text{cone}} \|_2 \right) \nonumber \\
    & + w_{\text{vel}} \cdot \| \mathbf{v}_{\text{obj}} \|_2^2 + w_{\text{ctrl}} \cdot \| \mathbf{a} \|_2 + J_{\text{safety}}
\end{align}
where $\mathbf{z}_{\text{cone}}$ is the cone's $z$-axis (aligned with $\mathbf{z}_{\text{world}}$ when upright), and $d_{\text{thresh}}$ is a threshold that caps the torso term. \revised{Note the sign of the torso term: because it enters the cost negatively, it \emph{rewards} torso distance from the cone up to the threshold $d_{\text{thresh}}$, encouraging the body to keep a standoff from the object while the gripper term draws the end-effector in. This division of labor---end-effector close, body clear---prevents the base from crowding or trampling the object while the arm manipulates it.} $J_{\text{safety}}$ includes penalties for robot falling.

\noindent\textbf{Chair Upright:}
The chair upright task uses a similar cost structure as the traffic cone:
\begin{align}
    J_{\text{Chair}} = \;
    & w_{\text{orient}} \cdot \left(1 - \exp\left( \alpha \cdot (\mathbf{z}_{\text{chair}} \cdot \mathbf{z}_{\text{world}} - 1) \right)\right) \nonumber \\
    & + w_{\text{gripper}} \cdot \| \mathbf{p}_{\text{gripper}} - \mathbf{p}_{\text{chair}} \|_2 \nonumber \\
    & - w_{\text{torso}} \cdot \min\left( d_{\text{thresh}}, \| \mathbf{p}_{\text{torso}} - \mathbf{p}_{\text{chair}} \|_2 \right) \nonumber \\
    & + w_{\text{vel}} \cdot \| \mathbf{v}_{\text{obj}} \|_2^2 + w_{\text{ctrl}} \cdot \| \mathbf{a} \|_2 + J_{\text{safety}}
\end{align}
where $\mathbf{z}_{\text{chair}}$ is the chair's $z$-axis (aligned with $\mathbf{z}_{\text{world}}$ when upright), $\mathbf{p}_{\text{gripper}}$ and $\mathbf{p}_{\text{chair}}$ are the gripper and chair positions, and $d_{\text{thresh}}$ is a threshold that caps the torso term. \revised{As in the cone task, the negative torso term rewards a body standoff of up to $d_{\text{thresh}}$ from the chair, keeping the base clear while the gripper term attracts the end-effector.} $\mathbf{v}_{\text{obj}}$ penalizes fast object movements, \revised{$\mathbf{a}$ is the planner action,} and $J_{\text{safety}}$ includes penalties for robot falling.

\noindent\textbf{Tire Stack:}
The tire stack task requires placing a standing tire on top of a flat tire, using stacking alignment and proximity terms:
\begin{align}
    J_{\text{Stack}} = \;
    & w_{xy} \cdot \| \mathbf{p}_{\text{top}}^{xy} - \mathbf{p}_{\text{bottom}}^{xy} \|_2 + w_{z} \cdot | p_{\text{top}}^z - p_{\text{desired}}^z | \nonumber \\
    & + w_{\text{orient}} \cdot (1 - \mathbf{y}_{\text{top}} \cdot \hat{\mathbf{u}}_{\text{stack}}) \nonumber \\
    & + w_{\text{bottom}} \cdot (\| \mathbf{v}_{\text{bottom}} \|_2 + \| \boldsymbol{\omega}_{\text{bottom}} \|_2) \nonumber \\
    & + w_{\text{gripper}} \cdot \| \mathbf{p}_{\text{gripper}} - \mathbf{p}_{\text{gripper}}^{\text{des}} \|_2 \nonumber \\
    & + w_{\text{torso}} \cdot \| \mathbf{p}_{\text{torso}} - \mathbf{p}_{\text{torso}}^{\text{des}} \|_2 + w_{\text{ctrl}} \cdot \| \mathbf{a} \|_2 + J_{\text{safety}}
\end{align}
where $\mathbf{p}_{\text{top}}^{xy}$ and $\mathbf{p}_{\text{bottom}}^{xy}$ are the XY positions of the top and bottom tires (top should be directly above bottom), $p_{\text{desired}}^z$ is the target height for proper stacking, $\mathbf{y}_{\text{top}}$ is the top tire's y-axis, and $\hat{\mathbf{u}}_{\text{stack}}$ is the unit vector from top to bottom tire. The bottom tire velocity term keeps the target tire stationary. Desired gripper and torso positions are computed dynamically based on the stacking geometry. \revised{The weights are $w_{xy}{=}100$, $w_{z}{=}200$, $w_{\text{orient}}{=}20$, $w_{\text{bottom}}{=}50$, $w_{\text{gripper}}{=}10$, $w_{\text{torso}}{=}10$, and $w_{\text{ctrl}}{=}1$; the orientation term is activated only once the top tire is within $2$\,m of the stacking target.}

\noindent\textbf{Crowd Barrier Drag:}
The crowd barrier drag task extends the barrier upright cost with a goal-reaching term:
\begin{align}
    J_{\text{Barrier Drag}} = \;
    & w_{\text{goal}} \cdot \| \mathbf{p}_{\text{barrier}} - \mathbf{p}_{\text{goal}} \|_2 \nonumber \\
    & + w_{\text{orient}} \cdot \left(1 - \exp\left( \alpha \cdot (\mathbf{z}_{\text{barrier}} \cdot \mathbf{z}_{\text{world}} - 1) \right)\right) \nonumber \\
    & + w_{\text{grasp}} \cdot \min\left( \| \mathbf{p}_{\text{grip}} - \mathbf{p}_{\text{grasp}}^{L} \|, \| \mathbf{p}_{\text{grip}} - \mathbf{p}_{\text{grasp}}^{R} \| \right) \nonumber \\
    & + w_{\text{grip}} \cdot \left( 1 - |\mathbf{x}_{\text{grip}} \cdot \mathbf{x}_{\text{barrier}}| + 1 - |\mathbf{y}_{\text{grip}} \cdot \mathbf{z}_{\text{barrier}}| \right) \nonumber \\
    & + w_{\text{vel}} \cdot \| \mathbf{v}_{\text{obj}} \|_2^2 + w_{\text{ctrl}} \cdot \| \mathbf{a} \|_2 + J_{\text{grasp}} + J_{\text{safety}}
\end{align}
where $\mathbf{p}_{\text{goal}}$ is the target position. \revised{The weights are $w_{\text{goal}}{=}500$, $w_{\text{orient}}{=}50$, $w_{\text{grasp}}{=}250$, $w_{\text{grip}}{=}100$, $w_{\text{vel}}{=}10$, and $w_{\text{ctrl}}{=}2$; $J_{\text{safety}}$ additionally penalizes the barrier tipping over during the drag.}

\noindent\textbf{Tire Rack Drag:}
The tire rack drag task uses goal-reaching, orientation alignment, and grasp terms:
\begin{align}
    J_{\text{Rack}} = \;
    & w_{\text{goal}} \cdot \| \mathbf{p}_{\text{rack}} - \mathbf{p}_{\text{goal}} \|_2 \nonumber \\
    & + w_{\text{orient}} \cdot \bigl( (1 - \mathbf{x}_{\text{rack}} \cdot \mathbf{x}_{\text{world}}) \nonumber \\
    & \quad + (1 - \mathbf{y}_{\text{rack}} \cdot \mathbf{y}_{\text{world}}) \nonumber \\
    & \quad + (1 - \mathbf{z}_{\text{rack}} \cdot \mathbf{z}_{\text{world}}) \bigr) \nonumber \\
    & + w_{\text{grasp}} \cdot \| \mathbf{p}_{\text{gripper}} - \mathbf{p}_{\text{grasp}} \|_2  \nonumber \\
    & + w_{\text{grip}} \cdot (1 - \mathbf{z}_{\text{gripper}} \cdot \mathbf{z}_{\text{rack}}) \nonumber \\
    & + w_{\text{approach}} \cdot \min_{i} \| \mathbf{p}_{\text{torso}} - \mathbf{p}_{\text{appr}}^{i} \|_2 + J_{\text{grasp}} + J_{\text{safety}}
\end{align}
where the orientation cost keeps the rack aligned with world axes, $\mathbf{p}_{\text{appr}}^{i}$ are multiple approach sites (left, mid, right), and the gripper orientation cost aligns the gripper's $z$-axis with the rack's $z$-axis. $J_{\text{safety}}$ includes penalties for robot falling and rack tipping over. \revised{The weights are $w_{\text{goal}}{=}500$, $w_{\text{orient}}{=}50$, $w_{\text{grasp}}{=}250$, $w_{\text{grip}}{=}250$, and $w_{\text{approach}}{=}100$.}

\noindent\textbf{Rugged Box Push:}
The rugged box push task uses goal-reaching with dynamic torso positioning for pushing:
\begin{align}
    J_{\text{Box Push}} = \;
    & w_{\text{goal}} \cdot \| \mathbf{p}_{\text{box}} - \mathbf{p}_{\text{goal}} \|_2 \nonumber \\
    & + w_{\text{orient}} \cdot \bigl( |1 - \mathbf{x}_{\text{box}} \cdot \mathbf{x}_{\text{world}}| \nonumber \\
    & \quad + |1 - \mathbf{y}_{\text{box}} \cdot \mathbf{y}_{\text{world}}| + |1 - \mathbf{z}_{\text{box}} \cdot \mathbf{z}_{\text{world}}| \bigr) \nonumber \\
    & + w_{\text{torso}} \cdot \| \mathbf{p}_{\text{torso}} - \mathbf{p}_{\text{torso}}^{\text{des}} \|_2 \nonumber \\
    &+ w_{\text{gripper}} \cdot \| \mathbf{p}_{\text{gripper}} - \mathbf{p}_{\text{box}} \|_2 \nonumber \\
    & + w_{\text{ctrl}} \cdot \| \mathbf{a} \|_2 + J_{\text{safety}}
\end{align}
where the desired torso position $\mathbf{p}_{\text{torso}}^{\text{des}}$ is computed dynamically as the position opposite the goal direction from the box, encouraging the robot to position behind the box for pushing. \revised{Specifically, in the horizontal plane, $\mathbf{p}_{\text{torso}}^{\text{des}} = \mathbf{p}_{\text{box}} - d \cdot \left( \mathbf{p}_{\text{goal}} - \mathbf{p}_{\text{box}} \right) / \| \mathbf{p}_{\text{goal}} - \mathbf{p}_{\text{box}} \|$ with standoff $d = 0.6$\,m and a fixed desired torso height of $0.4$\,m, which places the robot behind the box relative to the goal. The weights are $w_{\text{goal}}{=}500$, $w_{\text{orient}}{=}50$, $w_{\text{torso}}{=}30$, $w_{\text{gripper}}{=}0.1$, and $w_{\text{ctrl}}{=}2$.}

\subsubsection{G1 Tasks}\mbox{}\\
\noindent\textbf{Box Pushing:}
The G1 box pushing task uses goal-reaching, orientation, and bimanual proximity terms:
\begin{align}
    J_{\text{G1 Box}} = \;
    & w_{\text{goal}} \cdot \| \mathbf{p}_{\text{box}} - \mathbf{p}_{\text{goal}} \|_2 + w_{\text{orient}} \cdot |1 - \mathbf{y}_{\text{box}} \cdot \mathbf{z}_{\text{world}}| \nonumber \\
    & + w_{\text{hand}} \cdot \min\left( \| \mathbf{p}_{\text{left}} - \mathbf{p}_{\text{box}} \|, \| \mathbf{p}_{\text{right}} - \mathbf{p}_{\text{box}} \| \right) \nonumber \\
    & - w_{\text{pelvis}} \cdot \| \mathbf{p}_{\text{pelvis}} - \mathbf{p}_{\text{box}} \|_2 - w_{\text{facing}} \cdot \mathbf{x}_{\text{robot}} \cdot \mathbf{x}_{\text{world}} \nonumber \\
    & + w_{\text{ctrl}} \cdot (\| \mathbf{v}_{\text{base}} \|_2 + \| \mathbf{q}_{\text{arm}} - \mathbf{q}_{\text{arm}}^{\text{default}} \|_2) + J_{\text{safety}}
\end{align}
where $\mathbf{y}_{\text{box}}$ is the box's y-axis (should point up), $\mathbf{p}_{\text{left/right}}$ are left/right palm positions, and the robot orientation term encourages facing forward. The control cost penalizes base velocity and arm deviation from default poses. \revised{The negative pelvis term (with a small weight) rewards keeping the torso from crowding the box, so that the hands, rather than the body, make contact; the same term appears in the remaining G1 tasks. The weights are $w_{\text{goal}}{=}50$, $w_{\text{orient}}{=}15$, $w_{\text{hand}}{=}10$, $w_{\text{pelvis}}{=}0.1$, $w_{\text{facing}}{=}50$, and $w_{\text{ctrl}}{=}3$.}

\noindent\textbf{Chair Pushing:}
The G1 chair pushing task is similar to box pushing with XY-only goal distance:
\begin{align}
    J_{\text{G1 Chair}} = \;
    & w_{\text{goal}} \cdot \| \mathbf{p}_{\text{chair}}^{xy} - \mathbf{p}_{\text{goal}}^{xy} \|_2 + w_{\text{orient}} \cdot |1 - \mathbf{z}_{\text{chair}} \cdot \mathbf{z}_{\text{world}}| \nonumber \\
    & + w_{\text{hand}} \cdot \min\left( \| \mathbf{p}_{\text{left}} - \mathbf{p}_{\text{chair}} \|, \| \mathbf{p}_{\text{right}} - \mathbf{p}_{\text{chair}} \| \right) \nonumber \\
    & - w_{\text{pelvis}} \cdot \| \mathbf{p}_{\text{pelvis}} - \mathbf{p}_{\text{chair}} \|_2 + w_{\text{vel}} \cdot \| \mathbf{v}_{\text{obj}} \|_2^2 \nonumber \\
    & + w_{\text{ctrl}} \cdot (\| \mathbf{v}_{\text{base}} \|_2 + \| \mathbf{q}_{\text{arm}} - \mathbf{q}_{\text{arm}}^{\text{default}} \|_2) + J_{\text{safety}}
\end{align}
where the goal distance uses only the XY components, and $\mathbf{z}_{\text{chair}}$ is the chair's z-axis (should point up). An additional object velocity term penalizes fast chair movements. \revised{The weights are $w_{\text{goal}}{=}30$, $w_{\text{orient}}{=}15$, $w_{\text{hand}}{=}10$, $w_{\text{pelvis}}{=}0.1$, $w_{\text{vel}}{=}10$, and $w_{\text{ctrl}}{=}3$.}

\noindent\textbf{Door Opening:}
The G1 door opening task uses goal-reaching (to the other side of the door) and hand-to-handle proximity:
\begin{align}
    J_{\text{G1 Door}} = \;
    & w_{\text{goal}} \cdot \| \mathbf{p}_{\text{pelvis}} - \mathbf{p}_{\text{goal}} \|_2 + w_{\text{hand}} \cdot \| \mathbf{p}_{\text{right}} - \mathbf{p}_{\text{handle}} \|_2 \nonumber \\
    & - w_{\text{pelvis}} \cdot \| \mathbf{p}_{\text{pelvis}} - \mathbf{p}_{\text{door}} \|_2 - w_{\text{facing}} \cdot \mathbf{x}_{\text{robot}} \cdot \mathbf{x}_{\text{world}} \nonumber \\
    & + w_{\text{ctrl}} \cdot (\| \mathbf{v}_{\text{base}} \|_2 + \| \mathbf{q}_{\text{arm}} - \mathbf{q}_{\text{arm}}^{\text{default}} \|_2) + J_{\text{safety}}
\end{align}
where $\mathbf{p}_{\text{goal}}$ is on the other side of the door (requiring the robot to open it), $\mathbf{p}_{\text{handle}}$ is the door handle position, and $\mathbf{p}_{\text{door}}$ is the door center. \revised{The weights are $w_{\text{goal}}{=}30$, $w_{\text{hand}}{=}5$, $w_{\text{facing}}{=}50$, and $w_{\text{ctrl}}{=}3$.}

\noindent\textbf{Table Pushing:}
The G1 table pushing task uses XY goal distance, table orientation, and bimanual proximity terms:
\begin{align}
    J_{\text{G1 Table}} = \;
    & w_{\text{goal}} \cdot \| \mathbf{p}_{\text{table}}^{xy} - \mathbf{p}_{\text{goal}}^{xy} \|_2 + w_{\text{orient}} \cdot |1 - \mathbf{y}_{\text{table}} \cdot \mathbf{z}_{\text{world}}| \nonumber \\
    & + w_{\text{hand}} \cdot \min\left( \| \mathbf{p}_{\text{left}} - \mathbf{p}_{\text{table}} \|, \| \mathbf{p}_{\text{right}} - \mathbf{p}_{\text{table}} \| \right) \nonumber \\
    & - w_{\text{pelvis}} \cdot \| \mathbf{p}_{\text{pelvis}} - \mathbf{p}_{\text{table}} \|_2 + w_{\text{vel}} \cdot \| \mathbf{v}_{\text{obj}} \|_2^2 \nonumber \\
    & + w_{\text{ctrl}} \cdot (\| \mathbf{v}_{\text{base}} \|_2 + \| \mathbf{q}_{\text{arm}} - \mathbf{q}_{\text{arm}}^{\text{default}} \|_2) + J_{\text{safety}}
\end{align}
where the goal distance uses only the XY components, and $\mathbf{y}_{\text{table}}$ is the table's y-axis (should point up when upright). \revised{The weights are $w_{\text{goal}}{=}30$, $w_{\text{orient}}{=}15$, $w_{\text{hand}}{=}10$, $w_{\text{pelvis}}{=}0.1$, and $w_{\text{ctrl}}{=}3$.}